\documentclass[lettersize,journal]{IEEEtran}

\usepackage[hyphens]{url}
\usepackage{etoolbox}
\usepackage[format=plain,labelformat=simple,labelsep=period,font=small,compatibility=false]{caption}
\usepackage[font=footnotesize,skip=3pt,subrefformat=parens]{subcaption}
\usepackage{times}    
\usepackage{xspace}
\usepackage[dvipsnames]{xcolor}
\usepackage{graphicx}
\usepackage{amsmath}
\usepackage{amssymb}
\usepackage{booktabs}
\usepackage[numbers,sort&compress]{natbib}
\usepackage{xcolor}         

\definecolor{mydarkred}{rgb}{0.6,0,0}
\definecolor{mydarkgreen}{rgb}{0,0.6,0}
\usepackage{amsfonts}       
\usepackage{nicefrac}       
\usepackage{microtype}      
\usepackage{amsthm}
\usepackage{bm,bbm}
\usepackage[ruled,vlined]{algorithm2e}

\newtheorem{theorem}{Theorem}
\newtheorem{lemma}{Lemma}

\newtheorem{assumption}{Assumption}

\usepackage{threeparttable}
\usepackage{pifont}
\usepackage{upgreek}
\usepackage{colortbl}
\usepackage{threeparttable}
\usepackage{multirow}

\usepackage{amsmath, amssymb, amsthm, mathrsfs, bm}

\begin{document}

\title{Missing Pattern Tree based Decision Grouping \\ and Ensemble for Enhancing Pair Utilization \\ in Deep Incomplete Multi-View Clustering}

\author{
Jie Xu,
Wenyuan Yang,
Yazhou Ren,~\IEEEmembership{Senior Member,~IEEE},
Lifang He,~\IEEEmembership{Senior Member,~IEEE},\\
Philip S. Yu,~\IEEEmembership{Fellow,~IEEE},
Xiaofeng Zhu,~\IEEEmembership{Senior Member,~IEEE}
\IEEEcompsocitemizethanks{
\IEEEcompsocthanksitem Jie Xu, Wenyuan Yang, Yazhou Ren are with the School of Computer Science and Engineering, University of Electronic Science and Technology of China, Chengdu 611731, China.
\IEEEcompsocthanksitem Lifang He is with the Department of Computer Science and Engineering, Lehigh University, PA 18015, USA
\IEEEcompsocthanksitem Philip S. Yu is with the Department of Computer Science, University of Illinois Chicago, Chicago, IL 60607 USA.
\IEEEcompsocthanksitem Xiaofeng Zhu is with the School of Computer Science and Technology, Hainan University, Haikou 570228, China.
}
}




\maketitle

\begin{abstract}
Real-world multi-view data often exhibit highly inconsistent missing patterns, posing significant challenges for incomplete multi-view clustering (IMVC). Although existing IMVC methods have made progress from both imputation-based and imputation-free routes, they largely overlook the issue of pair underutilization. Specifically, inconsistent missing patterns prevent incomplete but available multi-view pairs from being fully exploited, thereby limiting the model performance. To address this limitation, we propose a novel missing-pattern tree based IMVC framework. Specifically, to fully leverage available multi-view pairs, we first introduce a missing-pattern tree model to group data into multiple decision sets according to their missing patterns, and then perform multi-view clustering within each set. Furthermore, a multi-view decision ensemble module is proposed to aggregate clustering results across all decision sets. This module infers uncertainty-based weights to suppress unreliable clustering decisions and produce robust outputs. Finally, we develop an ensemble-to-individual knowledge distillation module module, which transfers ensemble knowledge to view-specific clustering models. This design enables mutual enhancement between ensemble and individual modules by optimizing cross-view consistency and inter-cluster discrimination losses. Extensive theoretical analysis supports our key designs, and empirical experiments on multiple benchmark datasets demonstrate that our method effectively mitigates the pair underutilization issue and achieve superior IMVC performance.
\end{abstract}


\section{Introduction}
\IEEEPARstart{M}{ulti}-view clustering~\cite{chaudhuri2009multi,kumar2011co} (MVC) is an important research topic that deserves continuous investigation, as it can leverage complementary information across different views and has been adopted in widespread unsupervised data analysis applications like multi-view/modal/graph/omics~\cite{lin2025multi,9782584,vazquez2020multigraph}.
In practice, however, the assumption that each sample is observed in all views often fails.
Missing views commonly arise in real-world environments due to sensor failures, unavailability, data privacy, etc, which leads to the attention-getting research topic of incomplete multi-view clustering, namely IMVC.

Recently, many efforts have been made to design IMVC methods for performing MVC tasks on incomplete multi-view data~\cite{wen2023highly,wan2024fast,Jiang_2025_ICCV}.
In the literature, IMVC approaches can be built on traditional machine learning methods such as matrix factorization~\cite{wen2018incomplete}, graph learning~\cite{wang2022highly}, and subspace learning~\cite{wang2020icmsc}.
While these traditional IMVC methods can infer clustering results from incomplete data representation matrices, they often rely on linear assumptions and handcrafted similarity measures, making them unsuitable for complex and large-scale data analysis tasks~\cite{wen2022survey}.

Recently, deep IMVC methods leverage the powerful capabilities of neural networks for jointly learning embedded features and clustering, has attracted increasing attention~\cite{wen2020dimc,xue2021clustering,zhang2026structure}.
The critical challenge for deep IMVC is that incomplete multi-view data exhibit \emph{inconsistent missing patterns}, e.g., different samples might miss different views, making it infeasible to apply deep models with fixed input structures. To tackle this, existing deep IMVC methods can be classified into imputation-based IMVC and imputation-free IMVC.

\begin{figure*}[!t]
    \centering
    \includegraphics[width=1\linewidth]{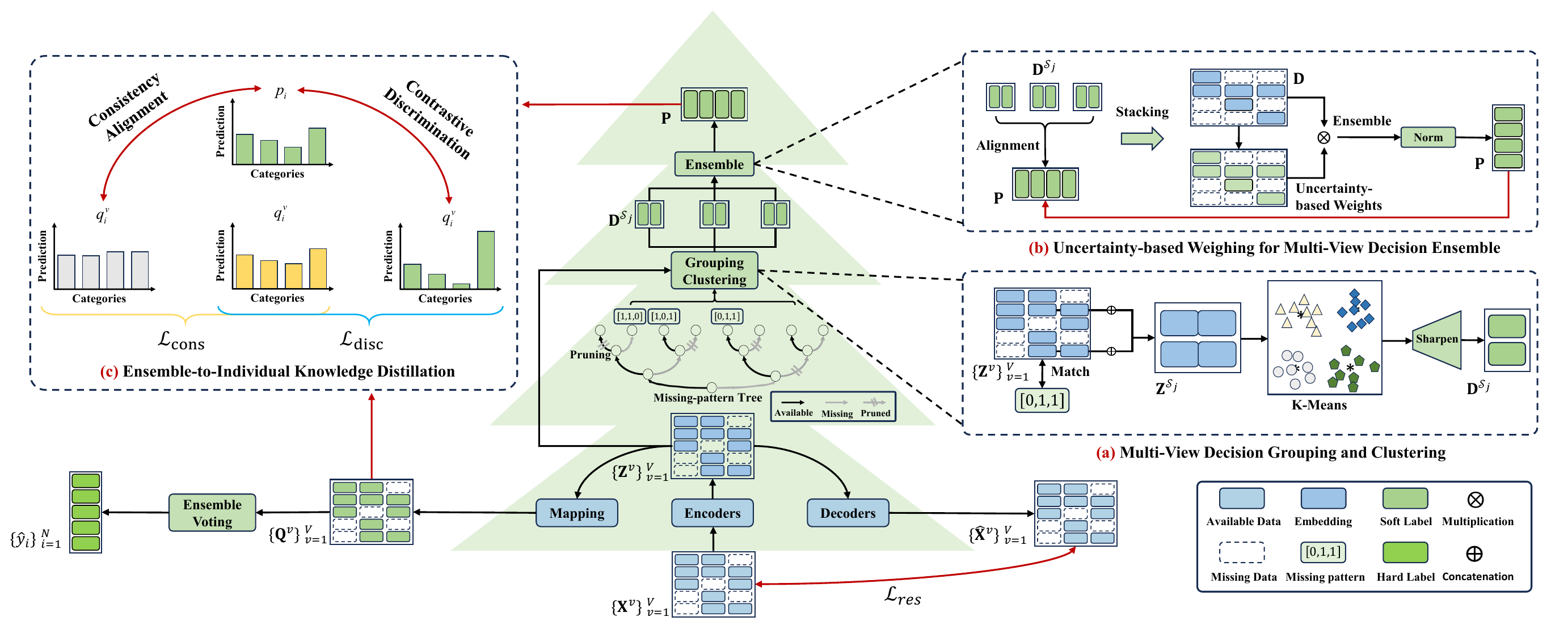}
    \caption{\textbf{Framework Overview of TreeEIC.} First, view-specific models generate sample embeddings via autoencoders. (a) The embeddings are grouped into multiple decision subsets according to missing patterns. Samples in each subset share the consistent missing pattern, allowing to obtain their clustering decisions in one feature space. (b) Clustering decisions from all subsets are aligned and weighted based on uncertainty to produce the ensemble robust clustering knowledge. (c) Knowledge distillation is then applied to transfer the ensemble results to individual view-specific models, responsible for cross-view consistency via $\mathcal{L}_{cons}$ and inter-cluster discrimination via $\mathcal{L}_{disc}$.}
    \label{fig:framework}
\end{figure*}

Despite the significant progress, deep IMVC still faces the following challenges.
First, imputation-based IMVC methods commonly exploit graph structures or cluster prototypes to recover missing views~\cite{yan2024deep,wen2020adaptive,zhang2026structure} and then conduct complete MVC. 
However, the imputation process in unsupervised scenarios can easily introduce noise and bias, as the imputed data might be inaccurate and form incorrect pairs with existing data, thereby degrading model performance~\cite{Xu_2022_AAAI,sun2025roll}.
To address this limitation, imputation-free deep IMVC methods~\cite{Xu_2022_AAAI,xu2023adaptive} have been proposed to avoid explicit missing view recovery.
This strategy usually trains the model on the complete part of multi-view data and then use it to infer the latent structure for the incomplete part of multi-view data.
Nevertheless, these methods fail to fully exploit the useful pairing information in the incomplete part, resulting in the \emph{pair underutilization issue}.
In particular, when the multi-view data is highly incomplete, the model struggles to learn from limited paired data and generalize to the majority of incomplete data.
{Although several imputation-free IMVC approaches~\cite{xu2023adaptive,dai2025imputation} attempt to project complete and incomplete multi-view data into a shared feature space for effective clustering, the original pair relationships are unchanged and these methods still ignore to address the pair underutilization issue limiting their performance.
}

To address the pair underutilization issue, we propose a novel deep IMVC method, termed \textbf{TreeEIC}, as illustrated in Figure~\ref{fig:framework}.
Our key insight is to perform clustering within subsets of samples that share consistent missing patterns, and then ensemble all patterns to produce robust results, inspired by ensemble learning~\cite{dietterich2002ensemble}.
First, we construct a \emph{missing-pattern tree} that hierarchically organizes all samples into multiple overlapping decision sets, 
where each set contains samples with consistent missing patterns.
By clustering within these sets, our method can fully utilize the originally available multi-view pairs to capture their complementary information.
Second, to mitigate the negative effect of unreliable and low-quality decision sets, we estimate the uncertainty of each decision and assign uncertainty-based weights in a \emph{multi-view decision ensemble} (MDE) process, resulting in a robust clustering decision for all views.
Then, we design an \emph{ensemble-to-individual knowledge distillation} (E2I) module, where the ensemble decision serves as a teacher and individual view decisions act as multiple students.
{This enables the robust clustering decision across all views to promote each view-specific clustering model, and in turn, each view-specific model further refines the ensemble clustering decisions, enabling iterative optimization.}
Finally, we provide theoretical analysis to validate the missing-pattern tree based grouping and uncertainty-weighted ensemble designs.
Extensive experiments on public benchmarks demonstrate that our method achieves state-of-the-art IMVC performance and exhibits superior robustness under highly inconsistent missing patterns.
In summary, our contributions include:
\begin{itemize}
    \item To the best of our knowledge, we are the first to address the pair underutilization issue in deep IMVC.
    We propose a novel missing pattern tree (MPT) based, imputation-free IMVC framework where incomplete multi-view data is organized by MPT with different missing patterns, enabling the effective exploitation of informative multi-view pairs to mitigate the pair underutilization issue.
    \item We establish a theoretically grounded multi-view decision grouping and ensemble framework that captures discriminative clustering information within each decision set and then aggregates clustering results across multiple decision sets by assigning uncertainty-based weights. This mechanism effectively suppresses unreliable decisions and produces a robust ensemble clustering outcome.
    \item We integrate ensemble clustering and incomplete multi-view learning into an end-to-end deep clustering framework. In an iterative ensemble-to-individual distillation scheme, our method transfers ensemble knowledge to individual views by imposing the inter-cluster discrimination and cross-view consistency losses.
\end{itemize}

\section{Related Work}\label{Method}

\subsection{Multi-View Clustering}
Existing MVC methods can be divided into multiple types as follows.
Graph-based MVC~\cite{nie2017self,peng2019comic} usually constructs multiple graph structures to explore the adjacency relationships between samples, and then utilizes spectral clustering.
Subspace-based MVC~\cite{8502831} expects to leverage consistent or diverse subspace representations to extract the useful information across multiple views.
Matrix factorization-based MVC~\cite{NIE2020107207} typically uses non-negative matrix factorization techniques to find the representative factors in multi-view data.
Kernel-based MVC~\cite{8611131} focuses on constructing multiple kernel functions on multi-view data to explore the possible non-linear relationship for promoting clustering tasks.
Tensor-based MVC~\cite{11370248,gu2024edison} stacks the feature representation matrices of multiple views into a tensor and then apply tensor decomposition to discover the clustering structures across views.
Deep learning-based MVC~\cite{Xu_2022_AAAI,10595464}, a.k.a, deep MVC, benefits from the powerful representation learning capabilities of deep neural networks and recently has achieved important progresses for the community.

Incomplete multi-view clustering (IMVC) or partial multi-view clustering research can trace back to \cite{li2014partial,xu2015multi}, which points out that we need to specially study IMVC due to the inherent limitations imposed on MVC methods in incomplete multi-view scenarios.
Basically, IMVC~\cite{liu2020efficient,wen2020cdimc,wang2021generative,lin2021completer,yan2024deep} can share the same methodologies as MVC by adding a data recovery process for the missing data during multi-view representation learning.
However, the inaccurate data recovery or noise imputation for the missing data is very likely to happen in unsupervised settings, especially for the data with high missing rates, which is harmful for the clustering performance instead.
To avoid this, pioneer work proposes imputation-free IMVC methods~\cite{Xu_2022_AAAI,xu2023adaptive} without requiring the unstable data reconstruction for missing data.
Nevertheless, these methods are unable to flexibly handle different missing patterns, resulting in the inability to effectively utilize the useful pair information contained in the incomplete multi-view data.
To address this urgent issue, we propose the missing pattern based solution with methodology novelty.

\subsection{Deep Incomplete Multi-View Clustering}
By introducing deep learning architectures, such as autoencoders~\cite{wang2018partial,wen2020dimc}, generation models~\cite{wang2021generative,zhang2026structure}, and attention based neural networks~\cite{yan2023gcfagg}, deep MVC can preserve the view-specific characteristics while capturing complex cross-view complementary information, demonstrating favourable clustering performance and scalability.
Following the success of deep MVC, researchers extend the similar pipelines to deep incomplete multi-view clustering (deep IMVC) tasks~\cite{tang2022deep,xu2023adaptive,Jiang_2025_ICCV}.
Existing approaches usually leverage view-specific autoencoders to learn features for incomplete multi-view data and then design various strategies to explore clustering-oriented multi-view information.
For example, 
\cite{wang2018partial,wang2021generative} combined autoencoders and generative adversarial networks to achieve the deep clustering of incomplete multi-view data as well as the data recovery of missing data.
\cite{wen2020dimc,wen2020cdimc} leveraged the embeddings pre-trained from encoder-decoder and incorporated a graph fusion strategy to achieve the consensus representation for incomplete views.
\cite{lin2021completer} introduced a dual network in deep IMVC pipeline to achieve the imputation of missing data, while maximizing the mutual information between two views to explore their common information.
\cite{Xu_2022_AAAI} proposed an imputation-free deep IMVC method which employs multiple autoencoders to extract multi-view embeddings and seeks their linear separability in a high-dimensional feature space.

Inspired by deep IMVC, our approach also adopts view-specific autoencoders to learn clustering-friendly embeddings for different views, while proposing the novel missing pattern tree based methodology for addressing the pair underutilization issue that has been ignored by previous work.

\subsection{Self-Supervised Multi-View Clustering}
Self-supervised learning (SSL) creates supervision signals for training models from data itself instead of using labels, which is the essential component in MVC domains.
The common SSL approaches for MVC include dimensionality reduction~\cite{Hinton2006Reducing}, contrastive learning~\cite{oord2018representation}, and pseudo-labelling~\cite{xu2022self}.
Most MVC methods adopt the dimensionality reduction models (such as subspace learning~\cite{zhang2017latent} and autoencoder~\cite{8387808,zhang2026structure}) to extract the data representations, which embed the multi-view data with inconsistent formats and modalities into the comparable feature space with the same dimensions, facilitating the subsequent multi-view information interaction and clustering.
Contrastive learning can achieve representation discrimination through narrowing the distance between positive sample pairs while widening that between negative sample pairs.
By treating different views of the same sample as positive samples pairs, contrastive learning is highly compatible with the goals of MVC that aims at exploring the association among views for clustering discrimination, and thus it has been employed in many recent deep MVC and IMVC methods~\cite{tang2022deepi,9852291}.
For pseudo-labelling, deep MVC and IMVC methods usually leverage this strategy to achieve iterative optimization and end-to-end clustering.
They iteratively employ clustering methods like K-Means~\cite{macqueen1967some} on the learned representations to obtain preliminary clustering, and then the clustering assignments construct pseudo labels to train an end-to-end clustering head by transferring clustering tasks into classification tasks~\cite{Xu_2022_AAAI,xu2022self}.


These SSL approaches are well-established and thus we also employ them as the deep learning backbone.
Unlike previous methods, our approach performs pseudo-label supervision and contrastive learning between the individual-view predictions and the ensemble clustering results, enhancing both cross-view consistency and inter-cluster discrimination.
Moreover, we establish the novel missing pattern tree model to achieve the higher pair utilization for IMVC tasks.

\section{Methodology}\label{MMMMMMMM}
\subsection{Preliminary}
\begin{figure}[!ht]
\centering
\includegraphics[width=0.7\linewidth]{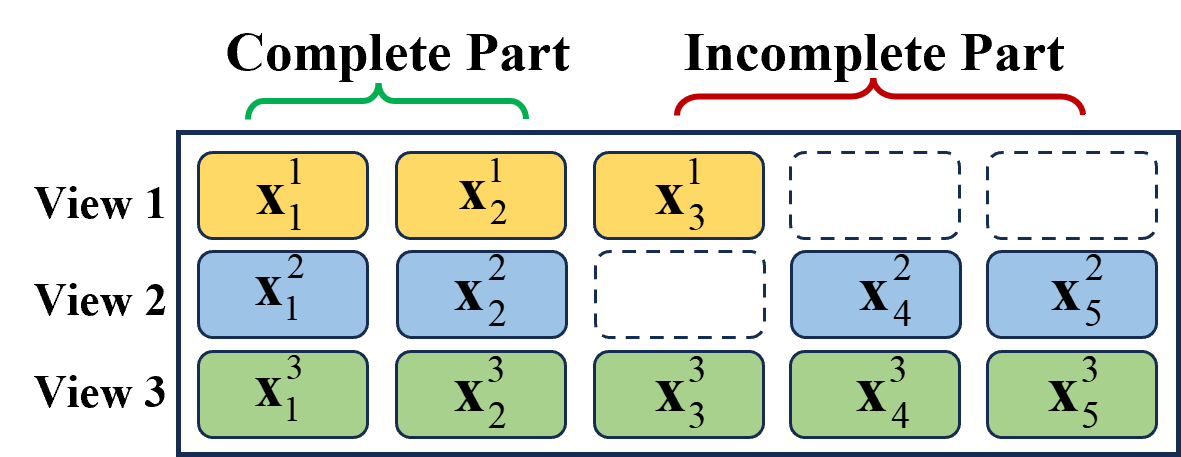}
\caption{Illustration of pair underutilization issue.}
\label{PUUI}
\end{figure}

\noindent\textbf{Problem Statement.}
Letting $N_v$ and $V$ denote the numbers of samples and views, $\mathcal{X} = \left\{ \mathbf{X}^v \in \mathbb{R}^{N_v \times D_v} \right\}_{v=1}^{V}$ represents the incomplete multi-view dataset, 
where $\mathbf{X}^v = \left\{ \mathbf{x}_i^v \in \mathbb{R}^{D_v} \right\}_{i=1}^{N_v}$ denotes the incomplete data of the $v$-th view.
The missing-view information is represented by a binary mask $\mathbf{A}=\left\{\mathbf{a}^v \in \{0,1\}^{N}\right\}^V_{v=1}$, where $a^v_i = 1$ if $\mathbf x_i^{v}$ is available, else $a^v_i = 0$. IMVC goal is to assign the $N$ samples to $K$ clusters.

\noindent\textbf{Pair Underutilization Issue.}
As shown in Fig.~\ref{PUUI}, we give a 3-view example to clarify the useful pair and the corresponding pair underutilization issue.
Useful pair refers to any two existed views for a sample.
For example, if three views of the $i$-th sample ($\mathbf x_i^{v_1}, \mathbf x_i^{v_2}, \mathbf x_i^{v_3}$) are not missing, they can constitutes three useful pairs, i.e., $<\mathbf x_i^{v_1}, \mathbf x_i^{v_2}>$, $<\mathbf x_i^{v_1}, \mathbf x_i^{v_3}>$, $<\mathbf x_i^{v_1}, \mathbf x_i^{v_3}>$.
Previous deep IMVC methods mainly explore multi-view complementarity on the complete part (the samples where all views exist)~\cite{lin2021completer,xu2023adaptive}, but overlook the fact that the incomplete part also contains useful pairs (e.g., $<\mathbf x_3^{1}, \mathbf x_3^{3}>$, $<\mathbf x_5^{2}, \mathbf x_5^{3}>$), which constitutes the pair underutilization issue.
To address this, our method introduces the missing pattern tree mechanism (detailed in Sec.~\ref{MPT}) enabling us sufficient utilization of the useful pairs in the incomplete part.

\noindent\textbf{View-Specific Deep Clustering Model.}
To achieve imputation-free and end-to-end clustering, 
we follow previous method~\cite{Xu_2022_AAAI} and employ $V$ view-specific clustering models to transform each view’s input 
${\mathbf{X}}^v$ into soft cluster assignments ${\mathbf{Q}}^v$.
Each model consists of an encoder $\mathcal{E}^v$, a decoder $\mathcal{D}^v$ and a clustering module $\mathcal{M}^v$. 
The encoder maps the input into an embedding space, 
$\mathbf{Z}^v = \mathcal{E}^v({\mathbf{X}}^v)$, 
while the decoder reconstructs the input from these embeddings. 
The reconstruction loss is optimized for training the autoencoder parameters:
\begin{equation}\label{eq:res}
\begin{aligned}
\mathcal{L}_{\mathrm{rec}} 
= \sum\nolimits_{v=1}^{V}\sum\nolimits_{i=1}^{N_v} 
\left\| {\mathbf{x}}^v_i - \mathcal{D}^v(\mathcal{E}^v({\mathbf{x}}^v_i)) \right\|_2^2.
\end{aligned}
\end{equation}
To obtain cluster assignments from each view’s latent representations, 
people usually introduce a parameterized mapping 
$\mathcal{M}^v(\mathbf{Z}^v; \mathbf{U}^v)$ that transforms 
the embedding $\mathbf{Z}^v \in \mathbb{R}^{N_v \times d_v}$ 
into a soft cluster assignment matrix, i.e., $\mathcal{M}^v(\mathbf{Z}^v; \mathbf{U}^v):\mathbf{Z}^v \in \mathbb{R}^{N_v \times d_v} \rightarrow \mathbf{Q}^v \in \mathbb{R}^{N_v \times K}$, where $\mathbf{U}^v = 
\left[
\mathbf{u}_1^v; \mathbf{u}_2^v; \dots; \mathbf{u}_K^v
\right] 
\in \mathbb{R}^{K \times d_v}$ denotes the learnable cluster centroids in the $v$-th view.
For the $i$-th sample and $j$-th cluster, the soft assignment is defined as:
\begin{equation}
q_{i,j}^v
=
\frac{
\left( 1 + \left\| \mathbf{z}_i^v - \mathbf{u}_j^v \right\|_2^2 \right)^{-1}
}{
\sum_{j=1}^{K} 
\left( 1 + \left\| \mathbf{z}_i^v - \mathbf{u}_{j}^v \right\|_2^2 \right)^{-1}
}\in \mathbf{Q}^v,
\label{eq:soft_assign}
\end{equation}
which represents the probability that the latent representation $\mathbf{z}_i^v$ is assigned to the $j$-th cluster form the $v$-th view.

Finally, the overall clustering result for the $i$-th sample is given by
\begin{equation}
\hat{y}_i^{} =  
\arg\max_{k}
\sum\nolimits_{v}
q^v_{i,k}~,~~s.t.~a_{i}^{v} = 1.
\label{eq:hard_label}
\end{equation}
\begin{figure}[!t]
    \centering
    \includegraphics[width=1\linewidth]{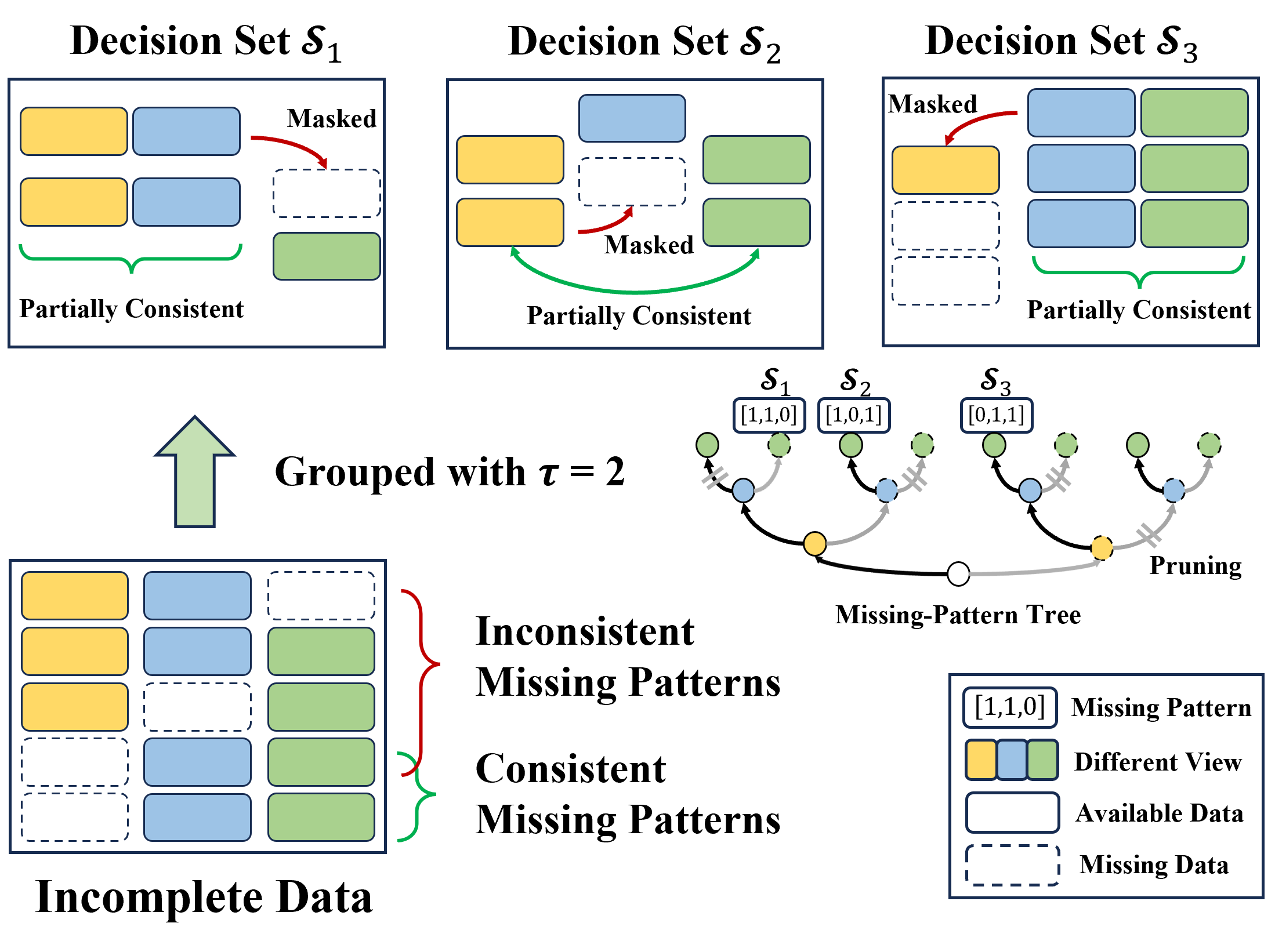}
    \caption{\textbf{Illustration of the missing patterns in MPT.}
    Numerous incomplete multi-view data exhibiting inconsistent missing patterns can be grouped into multiple decision sets. The missing pattern of samples in each decision set exhibits consistency and thus the pair relationship in these samples can be exploited to improve IMVC.}
    \label{fig:missing_pattern}
\end{figure}
\subsection{Missing Pattern Tree for Multi-View Decision}\label{MPT}
Heterogeneous and irregular data missing cause the underutilization issue of multi-view pairs in IMVC.
To address this, we propose the missing-pattern tree model for multi-view decision grouping and clustering as shown in Figure~\ref{fig:framework}(a).

\noindent\textbf{Missing Pattern Tree based Decision Grouping.}
Firstly, we encode each sample's missingness as a \emph{missing pattern} which formally represents the availability of each view for the sample.
Samples sharing the same missing pattern are then grouped into a decision set using a missing-pattern tree (MPT), thereby enabling effective exploitation of available cross-view information.

To be specific, the missing pattern of the $i$-th sample is defined as $\mathbf{m}_i = [a_i^1, a_i^2, \dots, a_i^V]$,
where $a_i^v$ is the $i$-th element of the $v$-th view binary mask $\mathbf{a}^v \in \mathbf{A}$.
{
To systematically represent all possible missing-view patterns, 
we construct a special binary tree, called the missing-pattern tree $\mathcal{T}$, of depth $V$. In the $\mathcal{T}$, the node at depth $0$ represents the root, and nodes at depths $1$ to $V$ are assigned values $0$ or $1$ to indicate whether the corresponding views are present.
Thus, the $j$-th possible missing pattern can be represented by the vector $\mathbf{m}_j \in \{0,1\}^V$, formed by the values of the nodes along the $j$-th path from the root to the corresponding leaf in $\mathcal{T}$, excluding the root node itself.
Since not all missing patterns are informative, which will be demonstrated in Section~\ref{sec:analsis}, 
we adopt a pruning strategy based on a threshold parameter $\tau \in \mathbb{Z}$, which dynamically adjusts according to the missing rate in the incomplete multi-view datasets. Specifically, we prune any branch that meets the following condition:
\begin{equation}
\sum_{v=1}^{d} m^v_j > \tau 
\;\text{or}\;
V - d < \tau - \!\sum_{v=1}^{d} m^v_j~,
\label{eq:prune}
\end{equation}
where $m_j^v\in \mathbf{m}_j$ and $d$ denotes the depth of the current node.
The threshold $\tau$ is adaptively determined by}
\begin{equation}\label{tau}
\tau = \left\lfloor \frac{v}{2} + \Big(\tau_{\max} - \frac{v}{2}\Big) \cdot (1 - \rho)^2 \right\rceil,
\end{equation}
where $\rho$ denotes the missing rate and $\tau_{\max}$ is a defined upper bound typically ranging from $V/2$ to $V$.

After applying the pruning strategy, we obtain effective $|\mathcal{C}| = C(V, \tau)$ missing patterns, which are represented by the set $\mathcal{C} = \{\mathbf{m}_1, \mathbf{m}_2, \dots, \mathbf{m}_{|\mathcal{C}|}\}$.
Based on the missing patterns in $\mathcal{C}$, we group the incomplete multi-view dataset ${\mathcal{X}}$ into $|\mathcal{C}|$ different decision sets $\mathcal{S}$:
\begin{equation}
 \left\{
\begin{aligned}
\mathcal{S}_j &= \left\{ i \mid \mathbf{m}_i \odot \mathbf{m}_j = \tau \right\}, \\
\mathcal{V}_j &= \left\{ v \mid \mathbf{m}_j^v = 1 \right\},
\end{aligned}
\right.
\quad j = 1, 2, \dots, |\mathcal{C}|,
\end{equation}
where $\odot$ denotes element-wise multiplication. In this way, all samples in $\mathcal{S}_j$ share the consistent missing pattern $\mathbf{m}_j$ (as illustrated in Figure~\ref{fig:missing_pattern}), ensuring the same configuration of $\tau$ available views (i.e., $v \in \mathcal{V}_j$) in the subset. Consequently, the available multi-view pairs in our method can be fully exploited to uncover the hidden multi-view complementary information.

\noindent\textbf{Group-Wise Multi-View Clustering.}
For each decision set $\mathcal{S}_j$, we can conduct conventional multi-view clustering to produce the multi-view decision $\mathbf{D}^{\mathcal{S}_j} \in \mathbb{R}^{|\mathcal{S}_j| \times K}$.
For simplicity, we directly employ previous multi-view clustering~\cite{Xu_2022_AAAI} which concatenates the representations and then performs $K$-Means to compute clustering results.

Specifically, for each decision set $\mathcal{S}_j$, we denote the concatenation of the samples' latent representations as $\mathbf{Z}^{\mathcal{S}_j} = \text{Concat}(\{\mathbf{Z}^{v}\}_{v \in \mathcal{V}_j}) \in \mathbb{R}^{|\mathcal{S}_j| \times \sum_{v \in \mathcal{V}_j}d_v}$. The soft cluster assignment of the $i$-th sample in $\mathcal{S}_j$ is obtained by:
\begin{equation}
p^{\mathcal{S}_j}_{i,k} = 
\frac{\left(1 + \lVert \mathbf{z}_i^{\mathcal{S}_j} - \mathbf{c}_k^{\mathcal{S}_j} \rVert_2^2 \right)^{-1}}
{\sum_{k=1}^{K} \left(1 + \lVert \mathbf{z}_i^{\mathcal{S}_j} - \mathbf{c}_{k}^{\mathcal{S}_j} \rVert_2^2 \right)^{-1}},
\label{eq:deep_clustering}
\end{equation}
where $\mathbf{z}_i^{\mathcal{S}_j} \in \mathbf{Z}^{\mathcal{S}_j}$ and $\mathbf{c}_k^{\mathcal{S}_j}$ denotes the $k$-th cluster centroid obtained by applying $K$-means on $\mathbf{Z}^{\mathcal{S}_j}$.
Then, we adopt the similar operation~\cite{Xu_2022_AAAI} of amplifying the high-probability predictions to obtain our multi-view clustering decision:
\begin{equation}
d^{\mathcal{S}_j}_{i,k} = 
\frac{\left(p^{\mathcal{S}_j}_{i,k}\right)^2/\sum_{i=1}^{|\mathcal{S}_j|}p^{\mathcal{S}_j}_{i,k}}
{\sum_{k=1}^{K}\left( \left(p^{\mathcal{S}_j}_{i,k}\right)^2/\sum_{i=1}^{|\mathcal{S}_j|}p^{\mathcal{S}_j}_{i,k}\right)} \in \mathbf{D}^{\mathcal{S}_j}.
\label{eq:sharpen}
\end{equation}

\subsection{Uncertainty-based Weighing for Multi-Decision Ensemble}
Given the multi-view clustering decision $\{\mathbf{D}^{\mathcal{S}_j}\}_{j=1}^{\mathcal{|C|}}$ across $\mathcal{|C|}$ independent decision sets,
we propose an uncertainty-based weighing strategy to obtain a robust multi-view decision ensemble as shown in Figure~\ref{fig:framework}(b).

To be specific, we first align the predictions into a common label space with Hungarian algorithm~\cite{jonker1986improving}, 
and then stack the aligned predictions from all decision sets to obtain a tensor ${\mathbf{D}} \in \mathbb{R}^{|\mathcal{C}| \times |\mathcal{U}| \times K}$ (the entries of absent clustering predictions are padded with $0$),
where $\mathcal{U} = \bigcup_{j=1}^{|\mathcal{C}|} \mathcal{S}_j$ denotes the union set of the samples in all decision subsets.
Furthermore, we introduce the uncertainty-based weight matrix $\mathbf{W} \in \mathbb{R}^{|\mathcal{C}|\times |\mathcal{U}|}$ to mitigate the negative effects from low-quality clustering decisions.
Specifically, for the $i$-th sample in $\mathcal{U}$, the uncertainty-based weight of the $j$-th decision set is computed by the inverse of entropy:
\begin{equation}\label{entropy}
{e}_{i}^j
= \left(- \sum\nolimits_{k=1}^{K} {d}_{i,k}^{\mathcal{S}_j}\log({d}_{i,k}^{\mathcal{S}_j} + \varepsilon) \right)^{-1},
\end{equation}
where $\varepsilon$ is a small constant to avoid numerical instability.
The motivation behind in Eq.~(\ref{entropy}) is that high entropy indicates more prediction ambiguity, whereas low entropy indicates confident predictions. This observation motivates us to use entropy to measure the uncertainty of clustering predictions.
The final uncertainty-based weight tensor is obtained by normalizing across all samples, i.e., ${{w}}_{i}^j 
= {{{e}}_{i}^j}/{\sum_{j=1}^{|\mathcal{S}_j|} {e}_{i}^j} \in \mathbf{W}$.

Afterwards, the uncertainty-based weighing strategy produces the ensemble clustering decision as:
\begin{equation}
\mathbf{P} \;=\;
\sum\nolimits_{j=1}^{|\mathcal{C}|}
\mathbf{W}[j,:] \odot{\mathbf{D}[j,:,:]} \in \mathbb{R}^{|\mathcal{U}| \times K},
\label{eq:ensemble}
\end{equation}
where $\odot$ is the element-wise multiplication.
$\mathbf{P} \in \mathbb{R}^{|\mathcal{U}| \times K}$ are then normalized over each class to obtain the prediction probability for robust multi-view ensemble decision.

\subsection{Ensemble-to-Individual Distillation}
Given the results of multi-view ensemble decision $\mathbf{P}$, we design an ensemble-to-individual distillation (E2I) scheme to improve view-specific models as shown in Figure~\ref{fig:framework}(c).

Specifically, we take $\mathbf{P}$ as the \emph{teacher} knowledge, and the predictions of the view-specific model $\mathbf{Q}^{v} \in \mathbb{R}^{N_v \times K}$ as the \emph{student} output.
Our distillation approach enforces the \emph{cross-view consistency} and \emph{inter-cluster discrimination} objectives to train the models as follows.

\noindent\textbf{Cross-View Consistency.}
We directly employ the mean squared error to achieve the consistency between the teacher knowledge and any student output:
\begin{equation}
\mathcal{L}_{\mathrm{cons}}
= \sum\nolimits_{v=1}^{V} \sum\nolimits_{i \in \mathcal{U}}
\big\|\mathbf{p}_{i}^{} - \mathbf{q}_i^{v} \big\|_2^2,
\label{eq:cons_loss_v}
\end{equation}
where $\mathbf{p}_i \in \mathbf{P}$ and $\mathbf{q}_i^{v} \in \mathbf{Q}^{v}$ denote the ensemble prediction and the individual view's prediction for the same sample.
This objective enforces the individual views to learn from the teacher’s reliable decision, thereby improving the individual views' clustering performance.

\noindent\textbf{Inter-Cluster Discrimination.}
Motivated by contrastive learning,
we employ the following InfoNCE~\cite{oord2018representation} loss to enhance inter-cluster separability of each student's clustering prediction based on teacher’s reliable decision:
\begin{equation}\label{eq:con_loss}
\mathcal{L}_{\text{InfoNCE}}^{(v)} = -\mathbb{E}_{s^+ \in \mathcal{P}^{(v)}} \left[ s^+ - \log \sum\nolimits_{s^- \in \mathcal{N}^{(v)}} e^{s^-} \right],
\end{equation}
where $\mathcal{P}^{(v)}$ and $\mathcal{N}^{(v)}$ respectively denote the sets of positive label-pair and negative label-pair constructed with $\mathbf{P}$ and $\mathbf{Q}^v$.
That is, $\mathcal{N}^{(v)} = \{<\mathbf{p}_{:,k}^{}, \mathbf{q}_{:,i}^v>,<\mathbf{q}_{:,k}^{v}, \mathbf{q}_{:,i}^v>\}_{k=1,\dots,K}^{k\neq i}$ is the set of negative label-pair for any $\mathbf{q}_{:,i}^{v}$.
Meanwhile, $\mathcal{P}^{(v)} = \{<\mathbf{p}_{:,k}^{}, \mathbf{q}_{:,k}^v>\}_{k=1}^K$ forms the set of positive label-pair.
$s^+$ and $s^-$ measure the distance computed by cosine similarity.
Our total clustering discrimination loss is:
\begin{equation}
\mathcal{L}_{\mathrm{disc}} =  \sum\nolimits_{v=1}^{V} \mathcal{L}_{\text{InfoNCE}}^{(v)}.
\label{eq:disc_total}
\end{equation} 
This objective pushes the view-specific models to learn well-clustered class predictions and thus encourages the models to learn more discriminative representations.

\noindent\textbf{E2I Optimization Objective.}
Finally, the E2I process has the following three losses:
\begin{equation}
\mathcal{L} 
= \mathcal{L}_{\mathrm{rec}} + \lambda_{\mathrm{1}}\mathcal{L}_{\mathrm{cons}} 
+ \lambda_{\mathrm{2}}\mathcal{L}_{\mathrm{disc}},
\label{eq:total_loss}
\end{equation}
where $\lambda_{\mathrm{1}}$ and $\lambda_{\mathrm{2}}$ are balance coefficients.
Minimizing $\mathcal{L}$ enables view-specific models to align with the ensemble clustering result as well as enhance their representation discrimination.
These improved representations in turn promote the multi-view ensemble decision process, thus achieving iterative optimization in our framework.

\subsection{Theoretical Analysis}
We provide theoretical analysis of the effectiveness for our missing-pattern tree based grouping and uncertainty-weighted ensemble strategies, along with an assessment of the algorithm's computational complexity.

\noindent\textbf{Missing-Pattern Tree based Grouping.} First, we provide a formal illustration of missing-pattern tree, and then analyze its advantages in utilizing pairs from a theoretical perspective.

Let \(\mathcal{I} = \{1,2,\dots,N\}\) be the set of sample indices and \(\mathcal{V} = \{1,2,\dots,V\}\) the set of view indices.
For each sample \(i \in \mathcal{I}\), define its \emph{available view set} as
\[
\mathcal{V}_i = \{ v \in \mathcal{V} \mid a_i^v = 1 \},
\]
where \(a_i^v \in \{0,1\}\) is the missing indicator (\(a_i^v = 1\) iff the \(v\)-th view is observed for sample \(i\)).
The \emph{view-pair set} of sample \(i\) is
\[
\mathcal{P}_i = \bigl\{ \{u,v\} \subseteq \mathcal{V}_i \mid u \neq v \bigr\},
\]
and its cardinality is \(p_i = |\mathcal{P}_i| = \binom{|\mathcal{V}_i|}{2}\).
The collection of all view-pairs over all samples forms a multiset
\[
\mathfrak{P} = \biguplus_{i \in \mathcal{I}} \bigl( \{i\} \times \mathcal{P}_i \bigr),
\]
with total count \(|\mathfrak{P}| = \sum_{i \in \mathcal{I}} p_i\).

Given a truncation parameter \(\tau \in [2, V]\) (which is adaptively determined by Eq.~(\ref{tau}) based on the missing rate), we define the \emph{family of admissible missing patterns} as
\[
\mathfrak{M}_\tau = \{ M \subseteq \mathcal{V} \mid |M| = \tau \},
\]
i.e., all \(\tau\)-element subsets of \(\mathcal{V}\).
For each pattern \(M \in \mathfrak{M}_\tau\), the corresponding \emph{decision set} is
\[
\mathcal{D}_M = \{ i \in \mathcal{I} \mid M \subseteq \mathcal{V}_i \}.
\]

In the TreeEIC framework, a view-pair \(\{u,v\} \subseteq \mathcal{V}_i\) is said to be \emph{utilized} if there exists a pattern \(M \in \mathfrak{M}_\tau\) such that
\[
\{u,v\} \subseteq M \subseteq \mathcal{V}_i.
\]
The multiset of all utilized sample-view-pair tuples is denoted by
\[
\mathfrak{U}_\tau = \Bigl\{ (i,\{u,v\}) \in \mathfrak{P} \;\Big|\; \exists M \in \mathfrak{M}_\tau:\ \{u,v\} \subseteq M \subseteq \mathcal{V}_i \Bigr\},
\]
and its cardinality is denoted by \(U(\tau) := |\mathfrak{U}_\tau|\), which measures the total number of view-pairs exploited by TreeEIC for a given \(\tau\).

In our method, \(\tau\) serves as a threshold, allowing only samples with at least \(\tau\) available views to contribute
to the ensemble decision process.
Each decision set comprises exactly \(\tau\) views, potentially containing the richness of cross‑view complementarity among these \(\tau\) views.
For comparison, the conventional approach (which performs multi-view learning only on the complete part of samples and is then extended to the incomplete part) utilizes the sub‑multiset
\[
\mathfrak{U}_{\text{cpt}} = \biguplus_{i: |\mathcal{V}_i| = V} \bigl( \{i\} \times \mathcal{P}_i \bigr),
\]
with cardinality \(U_{\text{cpt}} := |\mathfrak{U}_{\text{cpt}}|\):
\[
U_{\text{cpt}} =U(V)= \sum_{i: |\mathcal{V}_i| = V} \binom{V}{2}.
\]

Given the above notations, we have Theorem~\ref{utilization} as follows:
\begin{theorem}[Pair Utilization]\label{utilization}
For any threshold \(\tau \in [2, V]\), the number of view-pairs utilized by TreeEIC is given by
\[
U(\tau) = \sum_{\substack{i \in \mathcal{I} \\ |\mathcal{V}_i| \ge \tau}} \binom{|\mathcal{V}_i|}{2}.
\]
Then \(U(\tau)\) is non‑increasing on \([2,V]\), and we have the chain of inequalities
\begin{equation}\label{inequalities}
U(2) \;\ge\; U(\tau) \;\ge\; U(V) \;=\; U_{\text{cpt}},
\end{equation}
where \(U_{\text{cpt}}\) denotes the number of view-pairs utilized by the conventional method. The gap between \(U(\tau)\) and \(U_{\text{cpt}}\) is precisely
\begin{equation}\label{gap}
U(\tau) - U_{\text{cpt}} = \sum_{\substack{i \in \mathcal{I} \\ \tau \le |\mathcal{V}_i| < V}} \binom{|\mathcal{V}_i|}{2} \ge 0,
\end{equation}
with equality if and only if there is no sample \(i\) satisfying \(\tau \le |\mathcal{V}_i| < V\); i.e., the set $\mathcal{I}_\tau := \{ i \in \mathcal{I} \mid \tau \le |\mathcal{V}_i| < V \}$ is empty.
In particular, \(U(\tau) = U_{\text{cpt}}\) iff \(\mathcal{I}_\tau = \emptyset\).
\end{theorem}
\begin{proof}
    The proof is shown in Appendix I-A.
\end{proof}

This theoretical result reveals that lowering \(\tau\) incorporates more samples, thereby exploiting a larger number of view‑pairs, and raising \(\tau\) has the converse effect.
As indicated in Eq.~(\ref{gap}), our method $U(\tau)$ has a higher pair utilization property compared to traditional strategies $U_{\text{cpt}}$, effectively overcoming the pair under‑utilization problem that plagues conventional deep IMVC methods.
Furthermore, the monotonicity property as Eq.~(\ref{inequalities}) guarantees that the pair utilization decreases as \(\tau\) increases, with the extreme cases corresponding to \(\tau=2\) and \(\tau=V\).
When \(\tau=2\), the complementarity of multiple views is limited between two views; when \(\tau=V\), the complementary information across views is the most abundant, but incomplete samples do not meet this requirement.
As a result, our TreeEIC achieves a data‑driven trade‑off between view coverage and view‑pair utilization by the adaptive \(\tau\).
Under high missing rates, a smaller \(\tau\) retains a higher proportion of view‑pairs, which is crucial for maintaining clustering performance when data are severely incomplete.

\noindent\textbf{Uncertainty-Weighted Ensemble.}
Second, we consider use the true labels as a reference without incorporating them into the actual clustering process,
and then evaluate the empirical risks before and after using our uncertainty-weighted ensemble.

For each sample \(i \in \mathcal{I}\), there exists an unknown true label vector \(\mathbf{y}_i \in \{0,1\}^K\) (one-hot encoding of the cluster assignment).
Through the missing-pattern tree, we obtain a collection of decision sets \(\{\mathcal{S}_j\}_{j=1}^{|\mathcal{C}|}\). For sample \(i\), we denote \(J(i) = \{ j : i \in \mathcal{S}_j \}\) as the set of decision sets that contain \(i\); each such decision set produces a soft cluster assignment \(\mathbf{d}_i^j \in \Delta^{K-1}\) (the probability simplex in \(\mathbb{R}^K\)).

Then, we measure the discrepancy between a clustering prediction and a true label by the squared loss, and the risk loss of a \emph{base learner} \(j\) from one decision set on sample \(i\) is 
\[
L(\mathbf{y}_i, \mathbf{d}_i^j) = \|\mathbf{y}_i - \mathbf{d}_i^j\|_2^2.
\]
The ensemble prediction for sample \(i\) is the weighted combination $\mathbf{p}_i = \sum_{j \in J(i)} w_i^j \mathbf{d}_i^j$ by the entropy uncertainty based weight $w_i^j$.
Given the adequate assumption, we have Theorem~\ref{superiority}:
\begin{theorem}[Ensemble Risk Inequality]\label{superiority}
Under the assumption of entropy-loss positive correlation, if there exists at least one sample \(i\) for which the base learners are not all identical (i.e., \(\mathbf{d}_i^j \neq \mathbf{p}_i\) for some \(j\)), then we have
\[R_{\mathrm{ens}} < \mathbb{E}_i\!\left[ \sum_{j \in J(i)} w_i^j L(\mathbf{y}_i, \mathbf{d}_i^j) \right] \le \mathbb{E}_i\!\left[ \frac{1}{|J(i)|} \sum_{j \in J(i)} L(\mathbf{y}_i, \mathbf{d}_i^j) \right],\]
where $R_{\mathrm{ens}} = \mathbb{E}_i\big[ L(\mathbf{y}_i, \mathbf{p}_i) \big]$ denotes the ensemble risk across multiple decision sets.
\end{theorem}
\begin{proof}
The proof is shown in Appendix I-B.
\end{proof}
In particular, the ensemble risk $R_{\mathrm{ens}}$ is strictly smaller than the weighted average risk of the base learners, and consequently it is possible for the ensemble to achieve a risk lower than that of any individual base learner.
This theoretically supports the effectiveness of our proposed decision sets partitioning and ensemble strategies.

\begin{algorithm}[!t]
\caption{Training Procedure of TreeEIC}
\label{alg:TreeEIC}
\KwIn{Incomplete multi-view data $\left\{\mathbf{X}^v\right\}^V_{v=1}$, E2I training epochs $E$, ensemble interval $T$, $\tau_{\max}$, $\lambda_1$, $\lambda_2$, $K$.}
\BlankLine
\textbf{Initialization:} Initialize $\{\mathcal{E}^v, \mathcal{D}^v\}^V_{v=1}$ using Eq.~(1); 
initialize cluster centers $\{\mathbf{U}^v\}_{v=1}^V$ via $K$-means; initialize missing indicator matrix $\mathbf{A}$. \\
Calculate pruning threshold $\tau$ by Eq.~(5);\\
Obtain valid missing patterns $\mathcal{C}$ by Eq.~(4).
\BlankLine
\For{$e = 1$ \KwTo $E / T$}{
    Group decision sets according to Eq.~(6); \\
    Compute multi-view decisions $\{\mathbf{D}^{\mathcal{S}_j}\}$ by Eqs.~(7)--(8); \\
    Align multi-view decisions using the Hungarian algorithm and stack them into the decision tensor $\mathbf{D}$;\\
    Compute the uncertainty-based weight tensor $\mathbf{E}$ by Eq.~(9); \\
    Obtain the ensemble decision $\mathbf{P}$ by Eq.~(10);\\
    \For{$t = 1$ \KwTo $T$}{
        \For{$v = 1$ \KwTo $V$}{
            Compute individual prediction $\mathbf{Q}^v$ by Eq.~(2);\\
            Compute reconstruction loss $\mathcal{L}_{\mathrm{rec}}$ by Eq.~(1); \\
            Compute cross-view consistency loss $\mathcal{L}_{\mathrm{cons}}$ by Eq.~(11);\\
            Compute inter-cluster discrimination loss $\mathcal{L}_{\mathrm{disc}}$ by Eqs.~(12)--(13); \\
            Compute the overall loss $\mathcal{L}$ by Eq.~(14); \\
            Optimize $\{\mathcal{E}^v, \mathcal{D}^v\}$ and update cluster centers $\{\mathbf{U}^v\}$ to minimize $\mathcal{L}$ with mini-batch Adam. \\
        }
    }
}
\KwOut{Clustering result $\hat{y}_i\in \hat{\mathbf{Y}}$ computed by Eq.~(3)}
\end{algorithm}

\noindent\textbf{Complexity Analysis.}
Let $E$ denote the number of training epochs, $\mathcal{B}$ the batch size.
In the fully-observed case, for the decision ensemble process with $\tau$, each sample can be repetitively assigned into $|\mathcal{C}|$ decision groups.
The total number of samples within all groups is at most $N|\mathcal{C}|$.
The decision ensemble stage employs K-means with a cost of $\mathcal{O}(\tau N|\mathcal{C}|K D)$, where $D$ denotes the feature dimension.
Generating the uncertainty-aware weight tensor and performing ensemble require $2 \mathcal{O}(N|\mathcal{C}|K)$ operations, while aligning samples via the Hungarian algorithm adds an extra $\mathcal{O}(|\mathcal{C}|K^3)$ cost.
Therefore, the total cost of the ensemble step is $\mathcal{O}(\tau N|\mathcal{C}|K D) + 2\mathcal{O}(N|\mathcal{C}|K) + \mathcal{O}(|\mathcal{C}|K^3)$.
For the optimization stage, computing the reconstruction loss costs $\mathcal{O}(\mathcal{B}VD)$, computing the cross-view consistency loss costs $\mathcal{O}(\mathcal{B}VK)$, and the inter-cluster discrimination loss costs $\mathcal{O}(\mathcal{B}^2VK(2K-1))$. 
Thus, the total cost for these objectives is $\mathcal{O}(BVD + B^2VK(2K-1)+BVK)$.
Since the ensemble decision is computed every $T$ epochs, the overall computational complexity for $E$ epochs is
$(E/T)\mathcal{O}\left(\tau N|\mathcal{C}|K D  + 2N|\mathcal{C}|K+|\mathcal{C}|K^3\right) + (EN/B)\mathcal{O}\left(\mathcal{B}VD + \mathcal{B}^2VK(2K-1)+\mathcal{B}VK\right)$.

\begin{table}[t]
\centering
\caption{Statistics of multi-view datasets used in our experiments.}
\label{tab:dataset_info}
\resizebox{\linewidth}{!}{
\begin{tabular}{lccccc}
\toprule[2pt]
\textbf{Dataset} &\textbf{Type}& \textbf{\#Views} & \textbf{\#Samples} & \textbf{\#Clusters} \\
\midrule
HandWritten~\cite{lecun1989backpropagation}    & handwritten numerals & 6 & 2,000 & 10  \\
Caltech101-7~\cite{fei2004learning}   & single object images & 5 & 1,400 & 7  \\
OutdoorScene~\cite{hu2020multi}   & outdoor scene images & 4 & 2,688 & 8  \\
AWA-7~\cite{romera2015embarrassingly}         & animal images & 7 & 10,158  & 50  \\
ModelNet40~\cite{wu20153d}      &3D point cloud features & 2 & 12,311 & 40 & \\
MVP~\cite{pan2021variational}        & CAD model features & 8 & 1,600 & 8  \\
\bottomrule[2pt]
\end{tabular}
}
\end{table}
\section{Experiments}\label{et}
This section presents extensive experiments to evaluate the effectiveness of our proposed framework. 

\begin{table*}[!ht]
\caption{\textbf{Clustering Performance Comparison on Six Datasets.} We report the mean$\pm$std values of five runs. $\rho$ denotes the missing rate. ``N/A'' indicates that the method cannot run in the case of $\rho=1.0$. ``OOM'' denotes the method has failed to run due to out-of-memory.}\label{tab:comparison}
\centering
\renewcommand\tabcolsep{5.0pt} 
\small
\resizebox{\textwidth}{!}{
\begin{threeparttable}
\begin{tabular}{c|l|cc|cc|cc|cc|cc} 
\toprule[2pt]
\multirow{2}{*}{Dataset} & \multirow{2}{*}{~~~~Method} & \multicolumn{2}{c|}{$\rho=0.1$} & \multicolumn{2}{c|}{$\rho=0.3$} & \multicolumn{2}{c|}{$\rho=0.5$} & \multicolumn{2}{c|}{$\rho=0.7$} & \multicolumn{2}{c}{$\rho=1.0$} \\
\cmidrule(lr){3-4} \cmidrule(lr){5-6} \cmidrule(lr){7-8} \cmidrule(lr){9-10} \cmidrule(lr){11-12}
 & & ACC & NMI & ACC & NMI  & ACC & NMI  & ACC & NMI & ACC & NMI  \\
\midrule
\multirow{7}{*}{\rotatebox{90}{HandWritten}} 
 & APADC~\cite{xu2023adaptive} 
 & 79.13$\pm$1.02 & 80.41$\pm$1.03 
 & 78.81$\pm$1.42 & 78.05$\pm$1.30 
 & 79.42$\pm$2.21 & 78.10$\pm$0.91 
 & 70.94$\pm$4.28 & 69.06$\pm$3.32 
 & 37.12$\pm$2.51 & 33.65$\pm$1.39 \\

 & ProImp~\cite{li2023incomplete}
 & 85.92$\pm$2.55 & 82.14$\pm$1.38 
 & 82.65$\pm$1.79 & 77.66$\pm$1.57 
 & 82.05$\pm$3.61 & 74.33$\pm$3.04 
 & 80.45$\pm$4.35 & 70.05$\pm$3.48 
 & N/A & N/A \\
 
 & ICMVC~\cite{chao2024incomplete} 
 & 85.36$\pm$0.64 & 83.80$\pm$0.81 
 & 83.73$\pm$1.44 & 81.69$\pm$1.41 
 & 82.07$\pm$1.05 & 78.29$\pm$1.66 
 & 73.82$\pm$2.71 & 70.69$\pm$2.42 
 & 21.38$\pm$1.49 & 10.15$\pm$0.85 \\

 & DCG~\cite{zhang2025incomplete}
 &79.27$\pm$4.94 &78.25$\pm$3.35 
 &72.61$\pm$7.06 &74.35$\pm$3.06 
 &81.42$\pm$7.34 &77.50$\pm$4.71 
 &74.16$\pm$6.93 &69.46$\pm$4.14 
 &22.32$\pm$1.99 &12.88$\pm$1.81 \\
 
 & GHICMC~\cite{chao2025global}
 & 96.83$\pm$0.26 & 92.78$\pm$0.50 
 & 96.14$\pm$0.34 & 91.38$\pm$0.49 
 & 94.90$\pm$0.87 & 89.18$\pm$1.05 
 & 93.79$\pm$0.36 & 86.97$\pm$0.55 
 & 89.74$\pm$5.79 & 84.06$\pm$3.68 \\
 
 & FreeCSL~\cite{dai2025imputation}
 & 84.40$\pm$1.32 & 88.90$\pm$1.09 & 85.00$\pm$2.19 & 87.69$\pm$1.23 & 84.20$\pm$2.45 & 85.71$\pm$1.06 & 83.14$\pm$0.31 & 84.76$\pm$1.61 & 81.85$\pm$1.24 & 81.66$\pm$0.99 \\
 
 & \textbf{TreeEIC [ours]} & \textbf{97.20$\pm$0.19} & \textbf{93.77$\pm$0.20} & \textbf{96.69$\pm$0.24} & \textbf{92.75$\pm$0.41} & \textbf{95.86$\pm$0.37} & \textbf{91.14$\pm$0.88} & \textbf{94.41$\pm$0.67} & \textbf{88.54$\pm$1.03} & \textbf{93.77$\pm$1.04} & \textbf{87.09$\pm$1.87} \\

\hline
\multirow{7}{*}{\rotatebox{90}{Caltech101-7}} 
 & APADC~\cite{xu2023adaptive}
 & 55.79$\pm$0.65 & 42.65$\pm$0.44 
 & 44.14$\pm$4.59 & 36.13$\pm$8.45 
 & 47.01$\pm$5.19 & 47.30$\pm$5.18 
 & 48.81$\pm$4.71 & 46.55$\pm$4.43 
 & 38.63$\pm$0.40 & 29.77$\pm$1.04 \\

 & ProImp~\cite{li2023incomplete}
 & 81.71$\pm$0.70 & 71.56$\pm$2.22 & 75.72$\pm$4.61 & 64.23$\pm$5.05 & 71.87$\pm$7.00 & 59.89$\pm$6.20 & 70.23$\pm$5.55 & 58.47$\pm$4.80 & N/A & N/A \\

 & ICMVC~\cite{chao2024incomplete}
 & 80.94$\pm$4.72 & 74.07$\pm$3.23 
 & 81.24$\pm$6.91 & 73.46$\pm$5.21 
 & 80.70$\pm$3.97 & 72.16$\pm$3.07 
 & 65.41$\pm$4.65 & 59.07$\pm$4.76 
 & 27.24$\pm$0.60 & ~9.65$\pm$1.13 \\

 & DCG~\cite{zhang2025incomplete} & 79.84$\pm$3.77 & 72.22$\pm$4.02 & 79.46$\pm$3.80 & 71.65$\pm$3.57 & 73.42$\pm$5.57 & 64.12$\pm$4.62 & 78.14$\pm$5.37 & 67.10$\pm$4.90 & 30.13$\pm$4.48 & 13.11$\pm$4.04 \\

 & GHICMC~\cite{chao2025global} & 81.21$\pm$1.42 & 75.69$\pm$3.09 & 76.00$\pm$4.45 & 68.80$\pm$3.20 & 78.34$\pm$1.52 & 68.81$\pm$2.16 & 75.02$\pm$9.57 & 66.33$\pm$6.55 & 77.15$\pm$3.71 & 64.97$\pm$3.05 \\

 & FreeCSL~\cite{dai2025imputation} & 89.70$\pm$3.52 & 83.94$\pm$2.74 & 89.72$\pm$2.57 & 82.00$\pm$3.27 & 86.64$\pm$2.52 & 78.76$\pm$2.39 & 82.87$\pm$4.01 & 74.19$\pm$2.41 & 81.46$\pm$1.21 & 69.16$\pm$2.36 \\

 & \textbf{TreeEIC [ours]} & \textbf{92.31$\pm$1.15} & \textbf{85.90$\pm$1.47} & \textbf{90.41$\pm$2.12} & \textbf{82.57$\pm$2.72} & \textbf{90.61$\pm$1.86} & \textbf{82.96$\pm$2.16} & \textbf{89.99$\pm$1.57} & \textbf{81.94$\pm$2.34} & \textbf{88.27$\pm$1.18} & \textbf{78.70$\pm$1.58} \\

\hline
\multirow{7}{*}{\rotatebox{90}{OutdoorScene}} 
 & APADC~\cite{xu2023adaptive} 
 & 59.29$\pm$0.61 & 54.89$\pm$0.85 
 & 60.64$\pm$1.10 & 50.26$\pm$2.02 
 & 62.17$\pm$3.05 & 52.57$\pm$1.31 
 & 56.74$\pm$3.49 & 45.74$\pm$1.89 
 & 34.39$\pm$0.63 & 26.61$\pm$0.77 \\

& ProImp~\cite{li2023incomplete} 
& 63.15$\pm$1.71 & 50.35$\pm$1.63 
 & 59.67$\pm$1.47 & 47.42$\pm$0.64 
 & 57.02$\pm$2.08 & 45.51$\pm$0.58 
 & 53.21$\pm$3.05 & 42.35$\pm$1.22 
 & N/A & N/A\\

 & ICMVC~\cite{chao2024incomplete} 
 & 70.25$\pm$1.88 & 57.52$\pm$0.98 
 & 67.85$\pm$1.39 & 56.06$\pm$0.74 
 & 65.31$\pm$4.32 & 52.11$\pm$3.29 
 & 61.29$\pm$3.61 & 48.58$\pm$2.92 
 & 20.10$\pm$0.51 & ~4.39$\pm$0.12 \\

 & DCG~\cite{zhang2025incomplete} 
 & 65.08$\pm$3.29 & 54.61$\pm$1.95 
 & 64.48$\pm$6.36 & 53.88$\pm$4.89 
 & 61.39$\pm$5.36 & 50.87$\pm$3.27 
 & 60.61$\pm$5.24 & 49.55$\pm$2.92 
 & 24.87$\pm$1.86 & 10.60$\pm$3.14 \\

& GHICMC~\cite{chao2025global} 
& 72.38$\pm$3.01 & 59.93$\pm$1.37 
 & 71.32$\pm$2.82 & 57.40$\pm$1.41 
 & 69.03$\pm$1.11 & 55.42$\pm$0.81 
 & 66.26$\pm$3.77 & 52.28$\pm$1.77 
 & 62.16$\pm$4.46 & 47.54$\pm$2.71 \\

 &FreeCSL~\cite{dai2025imputation} 
 & 67.63$\pm$3.80 & 59.74$\pm$1.92 
 & 66.90$\pm$5.12 & 58.25$\pm$2.27 
 & 63.35$\pm$3.92 & 55.76$\pm$1.31 
 & 59.43$\pm$2.68 & 53.06$\pm$1.18 
 & 58.40$\pm$2.63 & 49.01$\pm$1.15 \\

& \textbf{TreeEIC [ours]} 
& \textbf{73.14$\pm$3.30} & \textbf{60.92$\pm$2.26} 
 & \textbf{72.49$\pm$4.80} & \textbf{61.14$\pm$1.69} 
 & \textbf{72.85$\pm$1.78} & \textbf{59.79$\pm$1.00} 
 & \textbf{71.12$\pm$1.55} & \textbf{57.41$\pm$1.43} 
 & \textbf{68.27$\pm$1.23} & \textbf{54.63$\pm$1.18} \\

\hline
\multirow{7}{*}{\rotatebox{90}{AWA-7}} 
& APADC~\cite{xu2023adaptive}
& 27.12$\pm$1.01 & 43.09$\pm$0.91 
& 27.29$\pm$0.61 & 42.88$\pm$0.43 
& 28.26$\pm$1.33 & 42.00$\pm$0.89 
& 29.24$\pm$0.96 & 42.19$\pm$0.93 
& 30.13$\pm$1.29 & 44.31$\pm$1.07 \\

& ProImp~\cite{li2023incomplete} 
& 40.84$\pm$1.11 & 54.24$\pm$0.63 
& 40.25$\pm$1.30 & 52.24$\pm$0.59 
& 36.47$\pm$2.47 & 49.64$\pm$1.02 
& 37.59$\pm$1.36 & 48.66$\pm$0.56 
& N/A & N/A \\

& ICMVC~\cite{chao2024incomplete} 
& 48.26$\pm$0.94 & 63.34$\pm$0.35 
& 48.38$\pm$1.61 & 63.31$\pm$0.43 
& 48.59$\pm$1.74 & \textbf{60.44$\pm$0.45} 
& 44.33$\pm$0.90 & \textbf{57.15$\pm$0.34} 
& ~7.05$\pm$0.17 & 10.59$\pm$0.61 \\ 

& DCG~\cite{zhang2025incomplete} 
& 32.76$\pm$1.83 & 47.34$\pm$1.09 
& 30.77$\pm$2.05 & 45.21$\pm$2.35 
& 30.29$\pm$1.48 & 43.79$\pm$1.69 
& 28.01$\pm$2.16 & 41.76$\pm$2.05 
& ~9.10$\pm$2.77 & 13.83$\pm$1.63 \\

& GHICMC~\cite{chao2025global} 
& OOM & OOM 
& OOM & OOM 
& OOM & OOM 
& OOM & OOM 
& OOM & OOM \\

& FreeCSL~\cite{dai2025imputation} 
& 10.67$\pm$0.24 & 17.30$\pm$0.26 
& 10.86$\pm$0.52 & 17.56$\pm$0.26 
& 11.18$\pm$0.52 & 17.33$\pm$0.43 
& 11.09$\pm$0.59 & 17.16$\pm$0.56 
& 10.47$\pm$0.32 & 16.42$\pm$0.36 \\

& \textbf{TreeEIC [ours]} 
& \textbf{63.53$\pm$0.44} & \textbf{69.39$\pm$0.34} 
& \textbf{58.32$\pm$0.64} & \textbf{63.69$\pm$0.64} 
& \textbf{54.63$\pm$0.79} & 58.48$\pm$0.42 
& \textbf{52.04$\pm$1.59} & 54.34$\pm$0.70 
& \textbf{46.45$\pm$0.21} & \textbf{46.67$\pm$0.49} \\
\hline

\multirow{7}{*}{\rotatebox{90}{ModelNet40}} 
& APADC~\cite{xu2023adaptive} 
& 39.28$\pm$3.00 & 52.29$\pm$1.15 
& 39.61$\pm$1.10 & 51.79$\pm$0.74 
& 39.27$\pm$0.92 & 51.34$\pm$0.42 
& 33.20$\pm$0.45 & 45.09$\pm$0.76 
& 26.98$\pm$0.65 & 38.32$\pm$0.21 \\

& ProImp~\cite{li2023incomplete} 
& 42.20$\pm$0.48 & 58.64$\pm$0.65 
& 40.26$\pm$0.96 & 55.77$\pm$0.58 
& 34.32$\pm$2.35 & 51.11$\pm$1.94 
& 33.83$\pm$0.56 & 51.67$\pm$0.63 
& N/A & N/A \\

& ICMVC~\cite{chao2024incomplete} 
& 47.26$\pm$1.27 & \textbf{61.55$\pm$0.57} 
& 45.81$\pm$1.58 & \textbf{60.73$\pm$0.76} 
& 41.42$\pm$1.07 & \textbf{58.19$\pm$0.57} 
& 30.63$\pm$1.07 & 50.39$\pm$1.17 
& ~7.42$\pm$0.08 & ~9.26$\pm$0.22 \\

& DCG~\cite{zhang2025incomplete} 
& 42.01$\pm$1.02 & 53.69$\pm$0.83 
& 42.90$\pm$1.02 & 53.89$\pm$0.75 
& 41.74$\pm$1.41 & 51.74$\pm$1.38 
& 40.38$\pm$2.27 & 50.32$\pm$0.80 
& 14.05$\pm$0.98 & 13.12$\pm$0.80 \\

& GHICMC~\cite{chao2025global} 
& OOM & OOM 
& OOM & OOM 
& OOM & OOM 
& OOM & OOM 
& OOM & OOM \\

& FreeCSL~\cite{dai2025imputation} 
& 43.44$\pm$1.11 & 58.21$\pm$0.47 
& 42.93$\pm$0.41 & 57.56$\pm$0.20 
& 42.45$\pm$0.83 & 56.43$\pm$0.94 
& 41.09$\pm$1.23 & \textbf{55.18$\pm$0.70} 
& 25.90$\pm$0.50 & 37.73$\pm$0.66 \\

& \textbf{TreeEIC [ours]}
& \textbf{48.05$\pm$1.97} & 59.23$\pm$1.17 
& \textbf{48.25$\pm$0.99} & 57.26$\pm$0.66 
& \textbf{45.37$\pm$1.71} & 55.28$\pm$1.00 
& \textbf{42.60$\pm$2.22} & 51.31$\pm$0.93 
& \textbf{39.46$\pm$1.40} & \textbf{53.79$\pm$0.57}  \\
\hline
\multirow{7}{*}{\rotatebox{90}{MVP}} 
& APADC~\cite{xu2023adaptive} 
& 83.10$\pm$0.48 & 72.47$\pm$0.74 
& 84.20$\pm$2.71 & 74.97$\pm$3.74 
& 86.30$\pm$1.86 & 76.78$\pm$2.61 
& 82.38$\pm$3.51 & 72.72$\pm$1.64 
& 49.51$\pm$3.50 & 44.63$\pm$5.33 \\

& ProImp~\cite{li2023incomplete} 
& 83.71$\pm$3.39 & 74.32$\pm$3.02 
& 83.20$\pm$2.28 & 72.88$\pm$2.21 
& 74.64$\pm$7.36 & 68.32$\pm$3.94 
& 77.37$\pm$6.25 & 67.26$\pm$4.02 
& N/A & N/A \\

& ICMVC~\cite{chao2024incomplete} 
& 88.66$\pm$3.94 & 81.35$\pm$2.81 
& 88.17$\pm$3.38 & 80.89$\pm$2.07 
& 80.26$\pm$4.58 & 75.38$\pm$3.27 
& 85.66$\pm$3.77 & 77.42$\pm$3.07 
& 22.78$\pm$0.66 & ~9.47$\pm$0.99 \\

& DCG~\cite{zhang2025incomplete} 
& 84.25$\pm$5.38 & 76.22$\pm$4.81 
& 83.41$\pm$4.91 & 76.83$\pm$3.71 
& 84.34$\pm$5.37 & 76.45$\pm$4.71 
& 82.26$\pm$5.41 & 73.50$\pm$2.98 
& 33.75$\pm$5.61 & 22.16$\pm$3.32 \\

& GHICMC~\cite{chao2025global}
&86.09$\pm$4.31 &79.07$\pm$2.96 &84.31$\pm$6.49 &79.78$\pm$3.68 &85.62$\pm$3.74 &79.14$\pm$2.03 &89.27$\pm$2.82 &81.17$\pm$2.32 &88.04$\pm$5.56 &81.95$\pm$1.69 \\

& FreeCSL~\cite{dai2025imputation} 
& 90.86$\pm$0.60 & \textbf{83.10$\pm$0.76} 
& 87.36$\pm$4.79 & 80.75$\pm$2.80 
& 89.20$\pm$1.83 & 81.31$\pm$1.99 
& 89.82$\pm$2.02 & 81.85$\pm$1.76 
& 89.03$\pm$1.44 & 80.33$\pm$1.43 \\
& \textbf{TreeEIC [ours]} 
& \textbf{91.31$\pm$0.50} & 82.87$\pm$0.65 
& \textbf{91.18$\pm$0.16} & \textbf{82.46$\pm$0.17} 
& \textbf{90.80$\pm$0.70} & \textbf{81.77$\pm$1.33} 
& \textbf{90.80$\pm$0.91} & \textbf{82.00$\pm$1.30} 
& \textbf{91.07$\pm$0.59} & \textbf{82.13$\pm$1.02} \\

\bottomrule[2pt]
\end{tabular}
\end{threeparttable}

}
\end{table*}

\subsection{Experimental Setup}\label{sec:setup}

\noindent\textbf{Datasets}.
Our experiments are conducted on six public multi-view datasets, whose detailed statistics are provided in Table~\ref{tab:dataset_info}.
The number of views varies across these datasets, providing diverse experimental scenarios that offer strong support for evaluating our method's ability to alleviate the pair underutilization issue.
To be specific,
HandWritten~\cite{lecun1989backpropagation} is a dataset of handwritten digits from $0$ to $9$, where each sample contains six handcrafted feature views, including Pixel, Fourier, Profile, Zer, Kar, and Mor.
Caltech101-7 is a subset of the Caltech101-7~\cite{fei2004learning} dataset with seven object categories, where each sample is represented by five visual views, namely Wavelet\_moments, Cenhist, HOG, Gist, and LBP.
OutdoorScene is a dataset of outdoor scene images divided into eight groups, where each sample contains four feature views: GIST , HOG , LBP , and Gabor.
AWA-7~\cite{romera2015embarrassingly} is an animal image dataset where each sample contains seven descriptor-based views, including PixColorHistogram, Profile, Local Self-Similarity, PHOG, colorSIFT, SURF, and SIFT.
ModelNet40~\cite{wu20153d} is a 3D point cloud dataset where each sample is represented by two heterogeneous feature views extracted from the point cloud using PointNet\_KAN and PointNet.
MVP~\cite{pan2021variational} is a multi-view partial point cloud dataset, and in our experiments we construct a subset containing eight object categories. For each object, we uniformly sample eight point clouds of 2048 points from different camera viewpoints, and extract a global feature from each point cloud using PointNet, resulting in eight homogeneous feature views.
We construct incomplete multi-view data with different missing rates $\rho \in [0.1, 0.3, 0.5, 0.7, 1.0]$.
For example, $\rho = 0.5$ represents that there are random $50\%$ samples are incomplete among all $N$ samples.
For each incomplete sample, we randomly mask its $1$ to $V-1$ views to produce missing views to simulate the inconsistent missing patterns for IMVC.

\noindent \textbf{Implementation Details}.
For a fair comparison, we implement view-specific autoencoders for each view following previous deep IMVC methods~\cite{Xu_2022_AAAI,dai2025imputation}.
Specifically, the autoencoder for the $v$-th view maps the input data $\mathbf{X}^v$ into a latent embedding $\mathbf{Z}^v$ and reconstruct the output data $\hat{\mathbf{X}}^v$ through a fully connected network with the architecture of $\mathbf{X}^v - 500 - 500 - 2000 - \mathbf{Z}^v - 2000 - 500 - 500 - \hat{\mathbf{X}}^v$.
In our experiments, the maximum pruning threshold $\tau_{\max}$ is not greater than $6$, and the minimum value of $\tau$ is set to 2 because a single view cannot form valid multi-view pairs.
We set $\lambda_1 = 0.01$ for HandWritten, Caltech101-7, OutdoorScene, AWA-7 and MVP datasets, $\lambda_1 = 0.001$ for ModelNet40 datasets, and $\lambda_2 = 0.2$ for all datasets.
The maximal epochs for pre-training autoencoders and for training the E2I model are $200$ and $700$, respectively.
The model is optimized with the Adam optimizer~\cite{kingma2014adam} with a learning rate of 0.0001, and the batch size is set to 256 for training.
The code of TreeEIC is implemented in PyTorch 2.7.1 (source code is provided in supplementary materials), and all experiments are conducted on Ubuntu 22.04.3 using an NVIDIA 3090 GPU with 24GB memory.
For comparison methods, we adopt their open-source codes with their recommended settings.

\noindent\textbf{Compared Methods.} 
We compare our TreeEIC with six state-of-the-art deep IMVC methods, categorized into imputation-based and imputation-free approaches.
The imputation-based IMVC methods include:
1)~ProImp~\cite{li2023incomplete}, which employs dual attention and contrastive objectives to recover missing information through prototype interactions,
2)~DCG~\cite{zhang2025incomplete}, which performs diffusion-based imputation with contrastive consistency for robust reconstruction,
3)~GHICMC~\cite{chao2025global}, which couples view-specific GCNs with global propagation for unified optimization.
The imputation-free IMVC methods include:
4)~APADC~\cite{xu2023adaptive}, which aligns cross-view distributions via adaptive projection,
5)~ICMVC~\cite{chao2024incomplete}, which integrates instance-level contrastive learning with high-confidence guidance,
and
6)~FreeCSL~\cite{dai2025imputation}, which constructs a shared prototype space to learn consensus semantics across views.


\noindent\textbf{Evaluation Metrics.} 
In this paper, we leverage two popular metrics, i.e., clustering accuracy (ACC) and normalized mutual information (NMI) to evaluate the clustering performance.
The reported results are mean values of five independent runs.
We provided more clustering evaluation metrics and standard deviations of results in the Appendix.

\subsection{Comparison Results}\label{sec:comparison}
\textbf{Quantitative Comparison.}
As shown in Table~\ref{tab:comparison}, the results demonstrate the effectiveness of our method which consistently achieves superior performance across various datasets and missing rates.
For instance, compared with the second-best method, our TreeEIC attains ACC improvements of $15.27\%$, $9.94\%$, $6.04\%$, $7.71\%$, $16.32\%$ on the AWA-7 dataset, and $0.76\%$, $1.17\%$, $3.82\%$, $4.86\%$, $6.11\%$ on the OutdoorScene dataset, respectively at the missing rates across $0.1, 0.3, 0.5, 0.7, 1.0$.
Furthermore, the challenge faced by IMVC the low robustness to different datasets, resulting in that current methods can only exhibit dataset-specific advantages.
For example, CHICMC performs favorably on small-scale datasets like HandWritten but is constrained by memory limitations on larger datasets, while FreeCSL achieves good results on Caltech101-7 but its performance degrades sharply on AWA-7.
Nevertheless, our TreeEIC exhibits robustness across the six datasets with heterogeneous and complex data characteristics.

\begin{figure}[!t]
    \centering
    \begin{subfigure}[b]{0.23\textwidth}
        \centering
        \includegraphics[width=\linewidth]{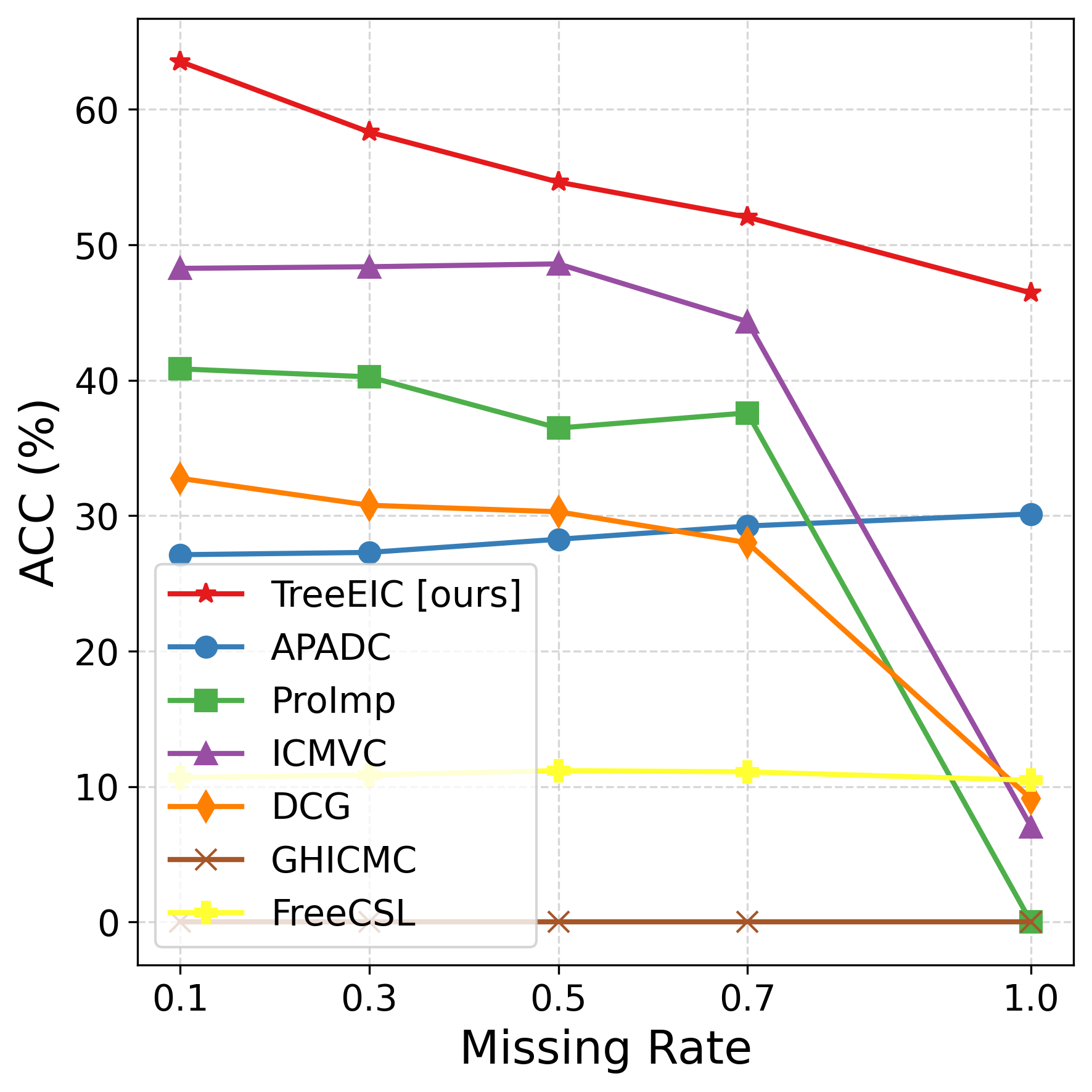}
        \caption{AWA-7}
        \label{fig:awa_acc}
    \end{subfigure}
    \begin{subfigure}[b]{0.23\textwidth}
        \centering
        \includegraphics[width=\linewidth]{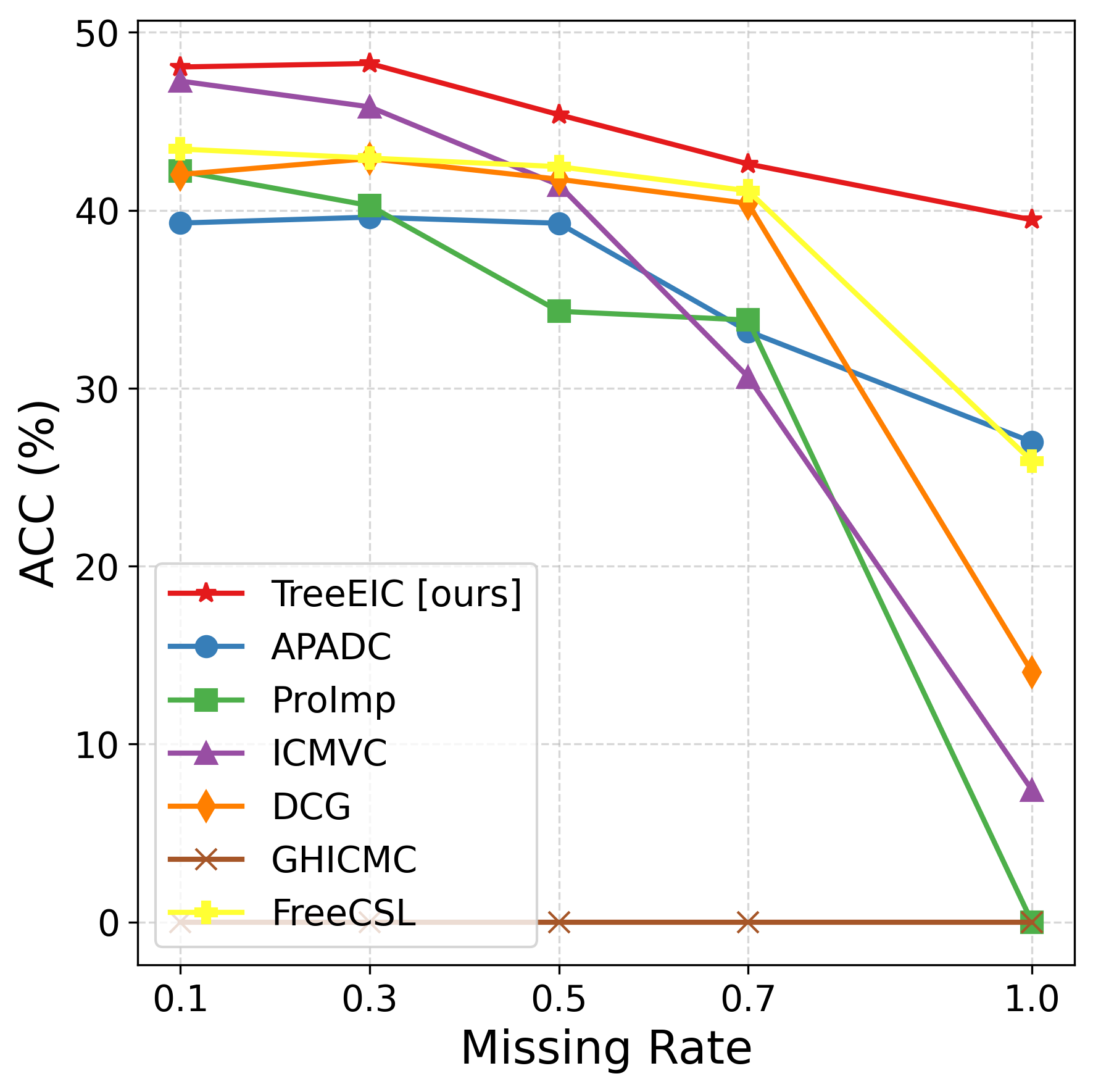}
        \caption{ModelNet40}
        \label{fig:ModelNet40}
    \end{subfigure}
    \caption{ACC $vs.$ Missing Rate on AWA-7 and ModelNet40. When the missing rate $\tau =1.0$, i.e., the highly inconsistent missing patterns, we can observe most IMVC methods have heavy performance degradation while our method TreeEIC is still robust.}
    \label{fig:acctrend}
\end{figure}

\begin{table}[!t]\caption{\textbf{Ablation Study on Model Variants.} MPT denotes the missing pattern tree for multi-view decision and MDE is the uncertainty-based weighing for multi-view decision ensemble.}\label{tab:two_components}
\centering
\renewcommand\tabcolsep{4.5pt}
\resizebox{\linewidth}{!}{
\begin{threeparttable}
    \begin{tabular}{c|cc|cc|cc|cc}
    \toprule[2pt]
    \multicolumn{1}{l|}{} &\multicolumn{2}{c|}{HandWritten} &\multicolumn{2}{c|}{Caltech101-7} &\multicolumn{2}{c|}{OutdoorScene} &\multicolumn{2}{c}{AWA-7} \cr
    \hline
     Variant & ACC & NMI & ACC & NMI& ACC & NMI& ACC & NMI  \\
    \hline
    without MPT        
    &83.24 &73.63 
    &84.91 &74.31 
    &63.88 &50.05
    &10.06 &12.89
    \\
    without MDE      
    &92.07 &84.57
    &87.34 &77.14
    &67.68 &54.07
    &40.58 &42.64 
    \\ 
    MPT+MDE        
    & \textbf{93.77} & \textbf{87.09}
    &\textbf{88.27} &\textbf{78.70}
    & \textbf{68.27} & \textbf{54.63}
    & \textbf{46.45} & \textbf{46.67}
    \\
    \bottomrule[2pt]
    \end{tabular}
\end{threeparttable}
}
\end{table}

\begin{table}[!t]\caption{\textbf{Ablation Study on Loss Components.} $\mathcal{L}_{cons}$ is cross-view consistency loss and $\mathcal{L}_{disc}$ is inter-cluster discrimination loss.}\label{tab:two_loss}
\centering
\renewcommand\tabcolsep{4.5pt}
\resizebox{\linewidth}{!}{
\begin{threeparttable}
    \begin{tabular}{cc|cc|cc|cc|cc}
    \toprule[2pt]
    \multicolumn{2}{c|}{Components} &\multicolumn{2}{c|}{HandWritten} &\multicolumn{2}{c|}{Caltech101-7} &\multicolumn{2}{c|}{OutdoorScene} &\multicolumn{2}{c}{AWA-7} \cr
    \hline
    $\mathcal{L}_{cons}$ &$\mathcal{L}_{disc}$   & ACC & NMI & ACC & NMI& ACC & NMI& ACC & NMI  \\
    \hline
    \ding{55} & \ding{55}   
    &35.37 &28.24
    &37.11 &19.16
    &26.38 &12.06 
    &11.10 &10.86
    \\
    \ding{51} & \ding{55}   
    &91.04 &83.16
    &52.19 &35.33
    &51.34 &35.83 
    &42.95 &42.29 
    
    \\
    \ding{55} & \ding{51}           
    &92.49 &86.69
    &87.53 &77.43
    &67.99 &54.48 
    &41.25 &43.26
    \\
    \ding{51} & \ding{51}           
    & \textbf{93.77} & \textbf{87.09}
    & \textbf{88.27} & \textbf{78.70}
    & \textbf{68.27} & \textbf{54.63}
    & \textbf{46.45} & \textbf{46.67} 
    \\
    \bottomrule[2pt]
    \end{tabular}
\end{threeparttable}
}
\end{table}

\begin{table}[!t]\caption{\textbf{Analysis on Missing-Pattern Tree.} Subscript $_{\mathcal{U}}$ denotes the clustering results of the sample union set $\mathcal{U}$, which will impact the performance of all $N$ samples marked with $_{N}$.}\label{tab:different_tau}
\centering
\resizebox{\linewidth}{!}{
\begin{threeparttable}
    \begin{tabular}{l|c|c|c|c|c}
    \toprule[2pt]
    \multicolumn{1}{c|}{$\tau$}&$|\mathcal{C}|$ & $|\mathcal{S}_j|$ &$|\mathcal{U}|$ & ACC$_{\mathcal{U}}$ $\rightarrow$ ACC$_{N}$ & NMI$_{\mathcal{U}}$ $\rightarrow$ NMI$_{N}$ \cr
    \hline
    1      
    &6 &1189  &2000 
    &81.40 $\rightarrow$ 82.45  &70.94 $\rightarrow$ 72.98
    \\
    2     
    &15 &687 &1957
    &86.20 $\rightarrow$ 89.40 &76.89 $\rightarrow$ 81.36
    \\
    3     
    &20 &382 &1658
    &87.82 $\rightarrow$ 91.85  &78.93 $\rightarrow$ 83.79
    \\
    4     
    &15 &196 &1045
    &84.78 $\rightarrow$ 88.70 &77.02 $\rightarrow$ 81.01
    \\
    \bottomrule[2pt]
    \end{tabular}
\end{threeparttable}
}
\end{table}

\begin{figure*}[!t]
    \centering
    \begin{subfigure}[b]{0.16\textwidth}
        \centering
        \includegraphics[width=\linewidth]{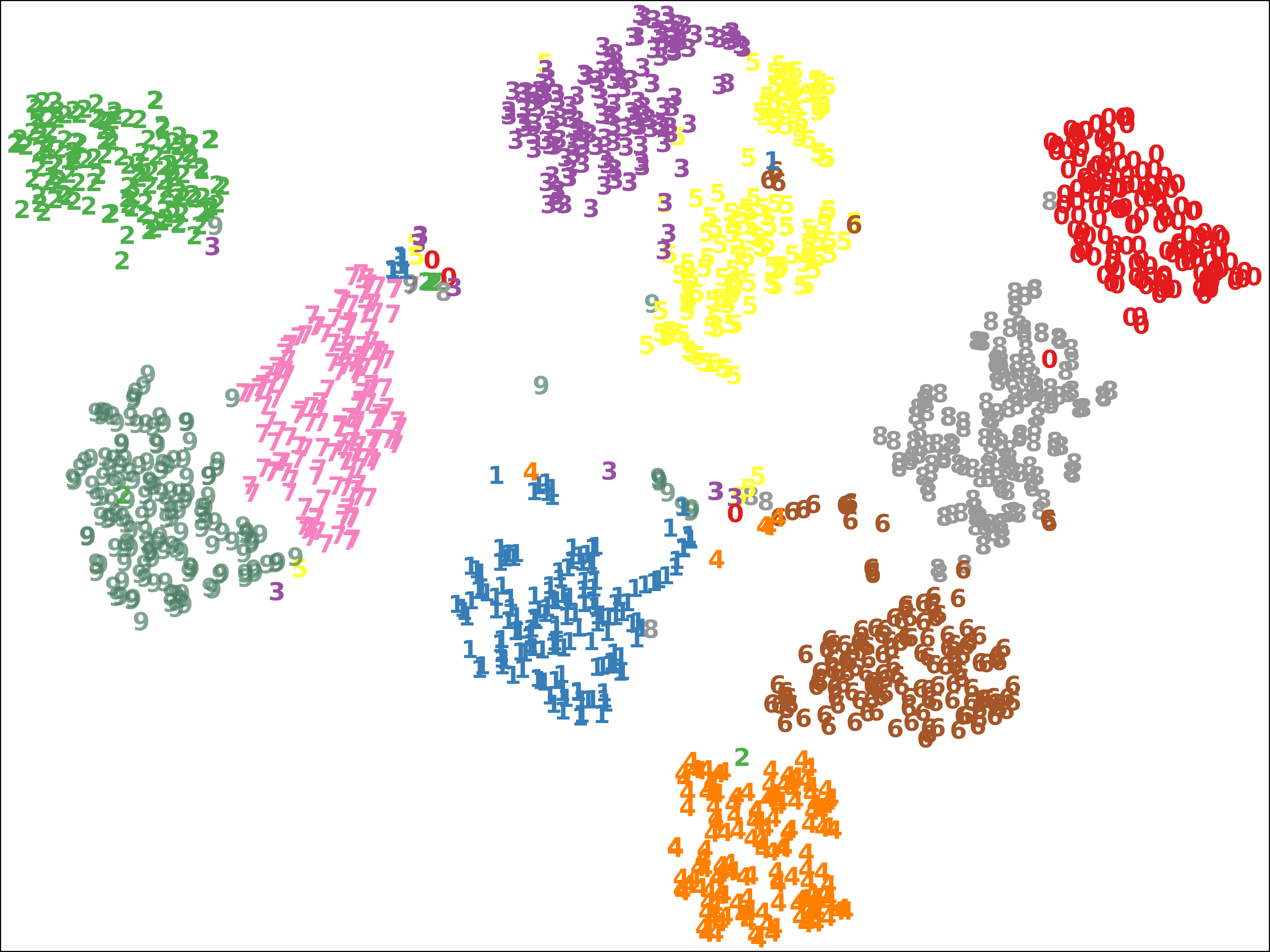}
        \caption{View 1}
        \label{fig:pre_view1}
    \end{subfigure}
    \hfill
    \begin{subfigure}[b]{0.16\textwidth}
        \centering
        \includegraphics[width=\linewidth]{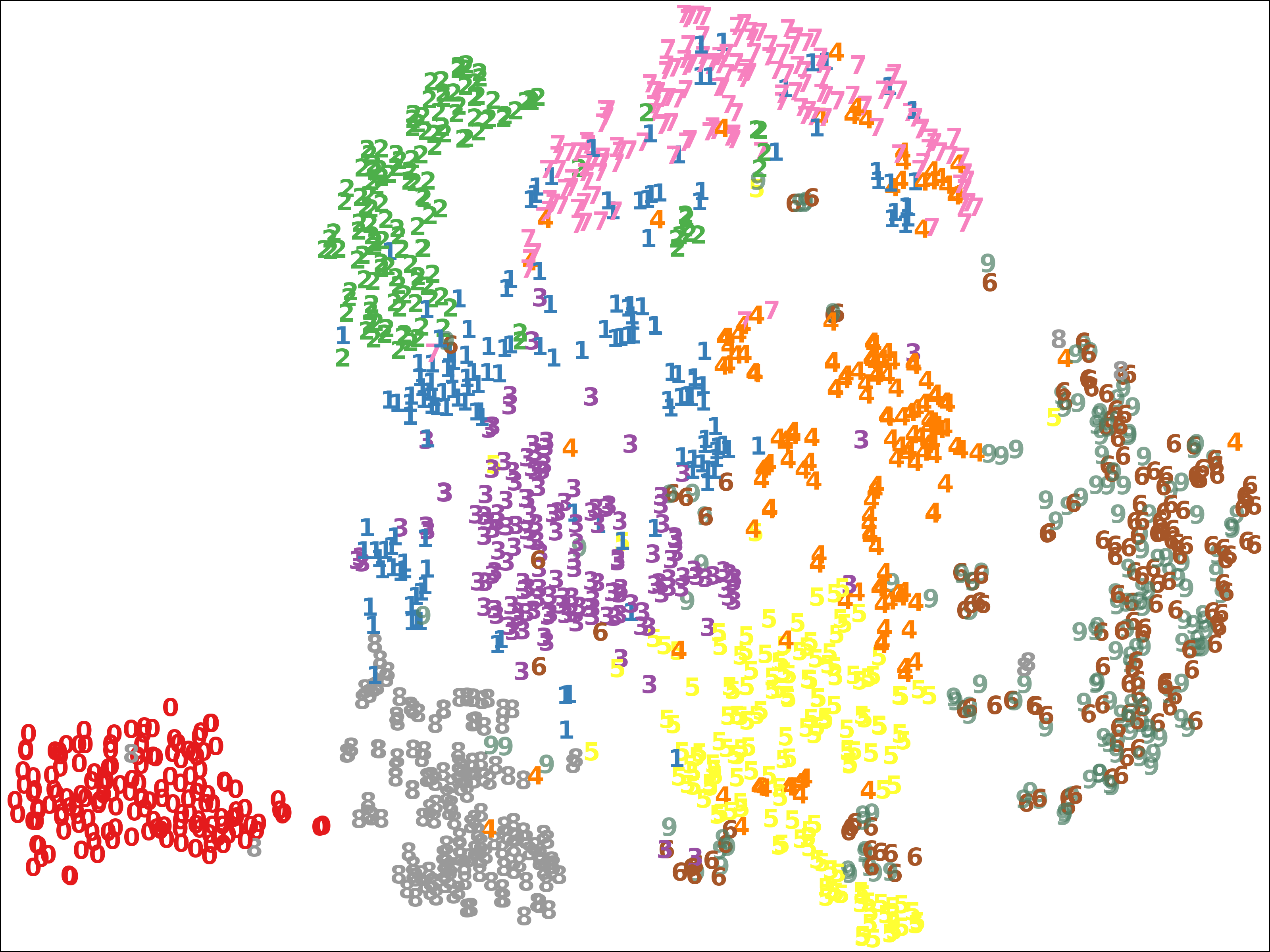}
        \caption{View 2}
        \label{fig:pre_view2}
    \end{subfigure}
    \hfill
    \begin{subfigure}[b]{0.16\textwidth}
        \centering
        \includegraphics[width=\linewidth]{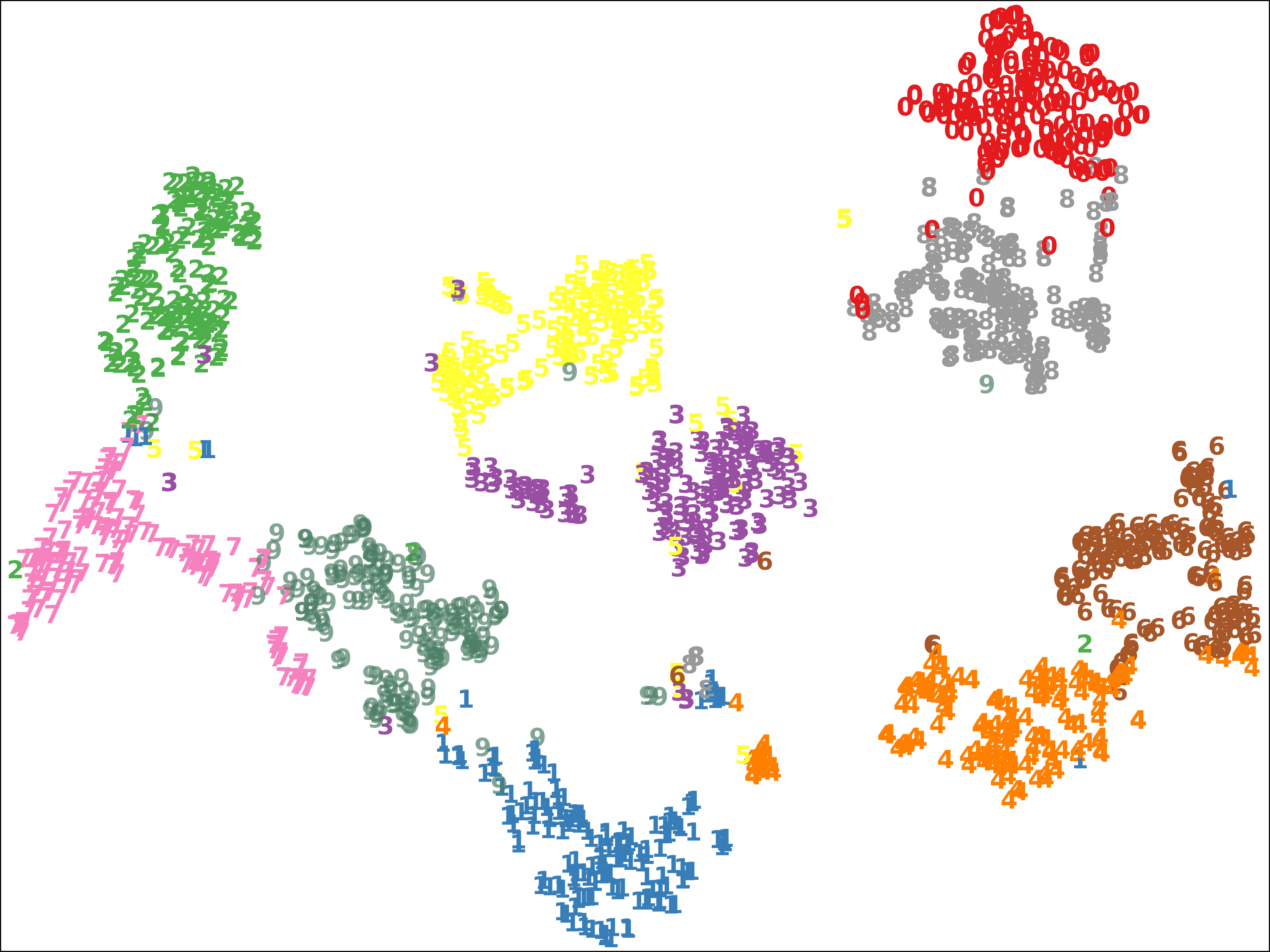}
        \caption{View 3}
        \label{fig:pre_view3}
    \end{subfigure}
    \hfill
    \begin{subfigure}[b]{0.16\textwidth}
        \centering
        \includegraphics[width=\linewidth]{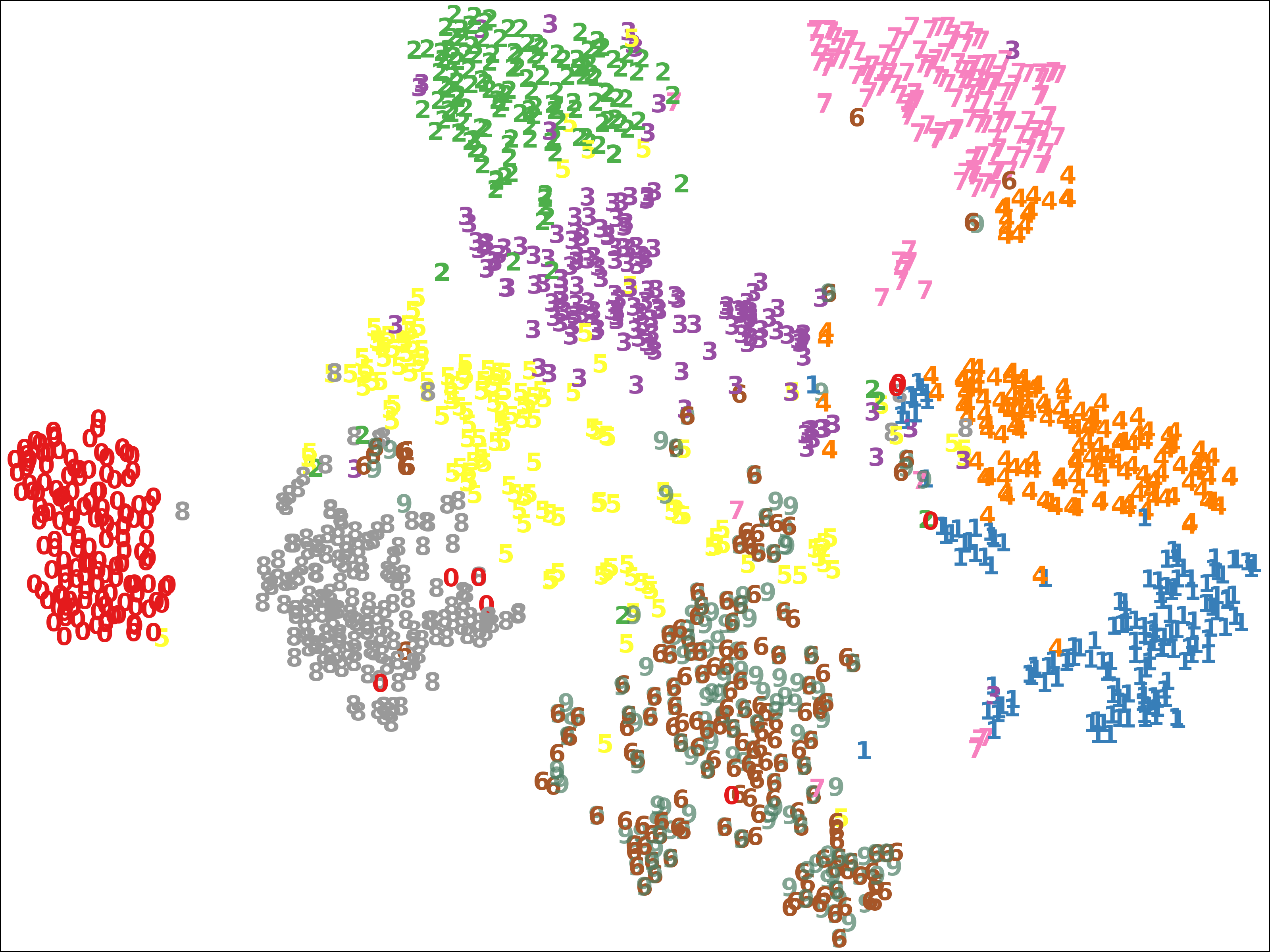}
        \caption{View 4}
        \label{fig:pre_view4}
    \end{subfigure}
    \hfill
    \begin{subfigure}[b]{0.16\textwidth}
        \centering
        \includegraphics[width=\linewidth]{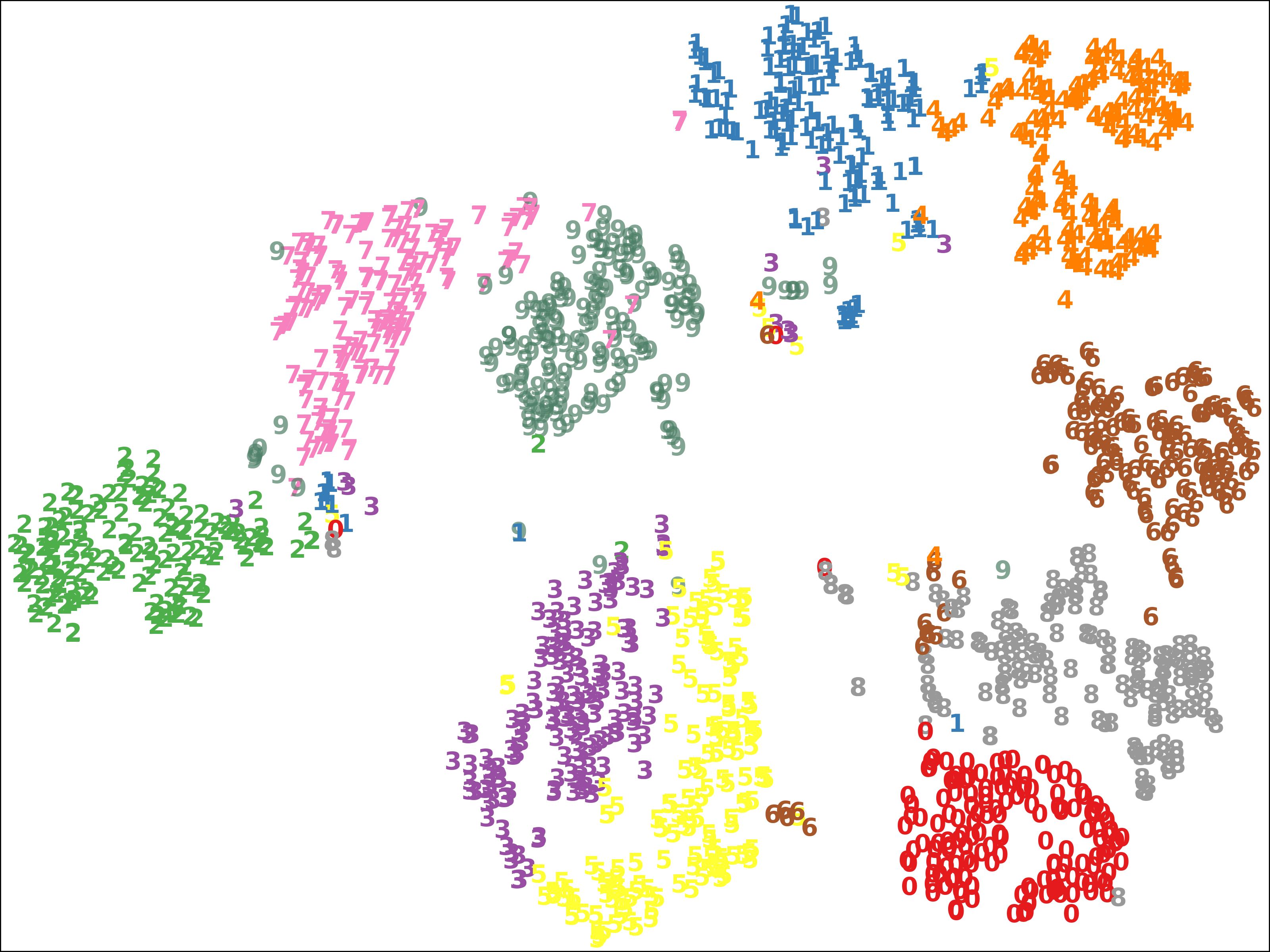}
        \caption{View 5}
        \label{fig:pre_view5}
    \end{subfigure}
    \hfill
    \begin{subfigure}[b]{0.16\textwidth}
        \centering
        \includegraphics[width=\linewidth]{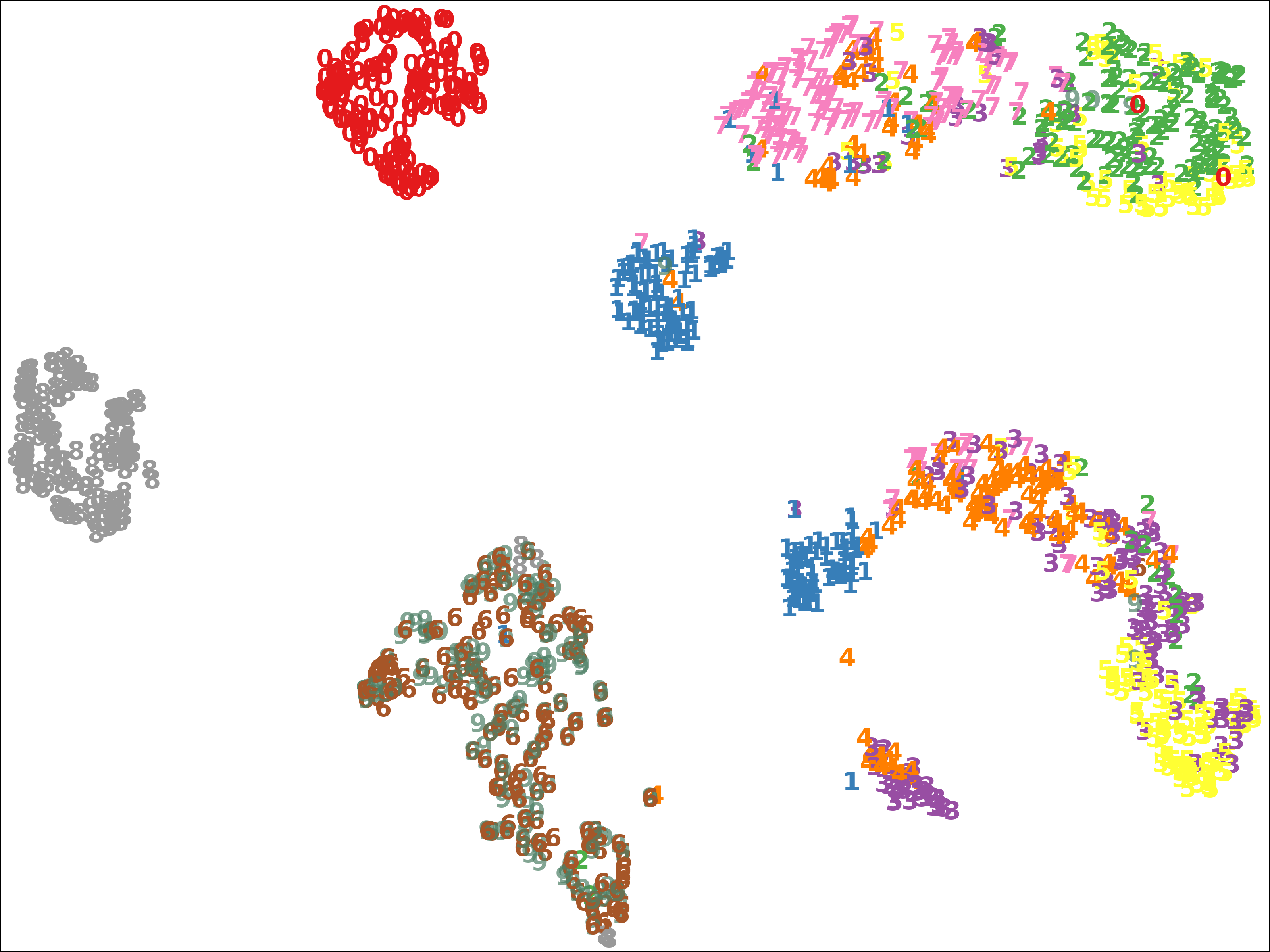}
        \caption{View 6}
        \label{fig:pre_view6}
    \end{subfigure}
    
    \vspace{0.25cm} %
    \begin{subfigure}[b]{0.16\textwidth}
        \centering
        \includegraphics[width=\linewidth]{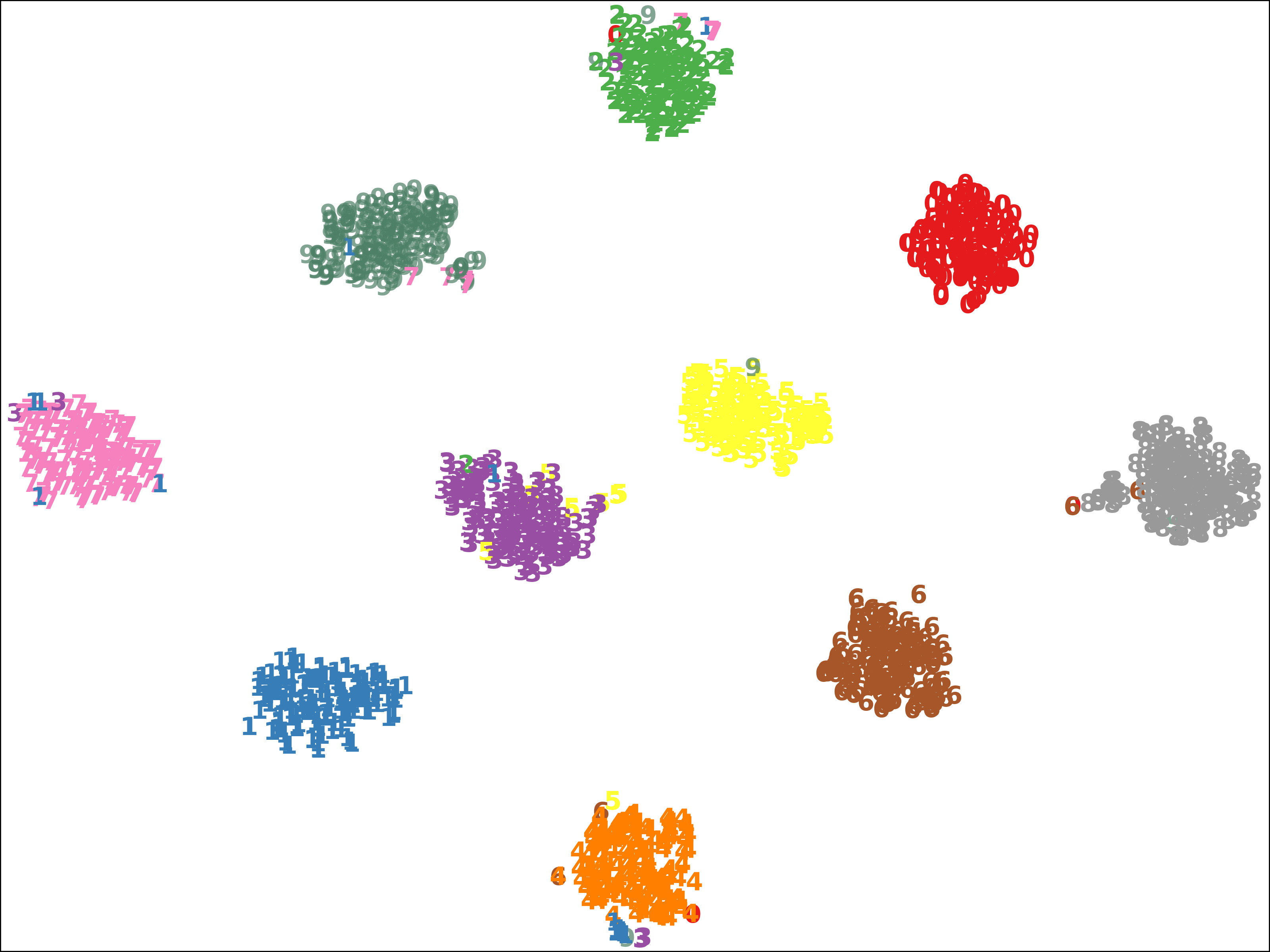}
        \caption{View 1}
        \label{fig:trained_view1}
    \end{subfigure}
    \hfill
    \begin{subfigure}[b]{0.16\textwidth}
        \centering
        \includegraphics[width=\linewidth]{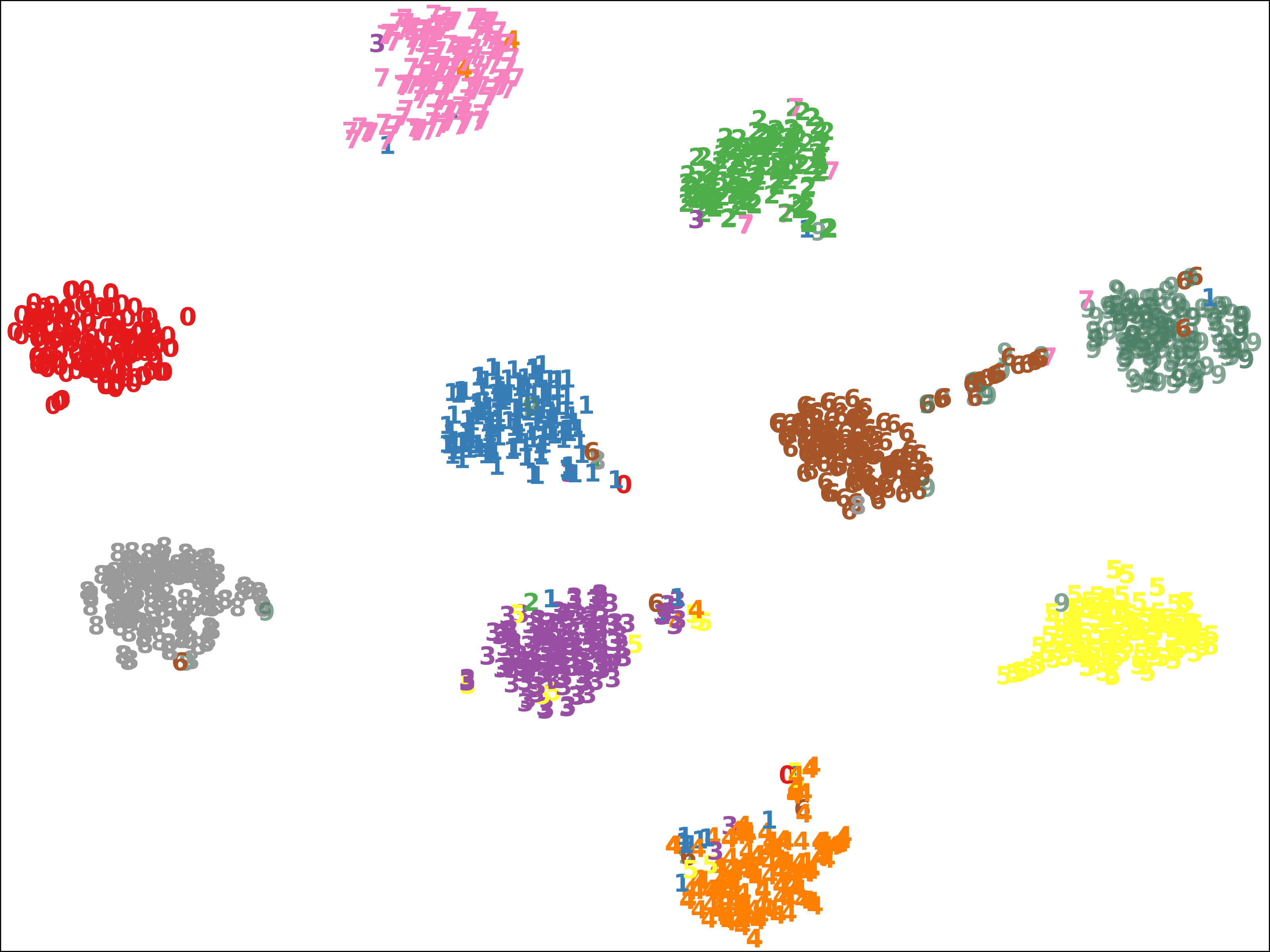}
        \caption{View 2}
        \label{fig:trained_view2}
    \end{subfigure}
    \hfill
    \begin{subfigure}[b]{0.16\textwidth}
        \centering
        \includegraphics[width=\linewidth]{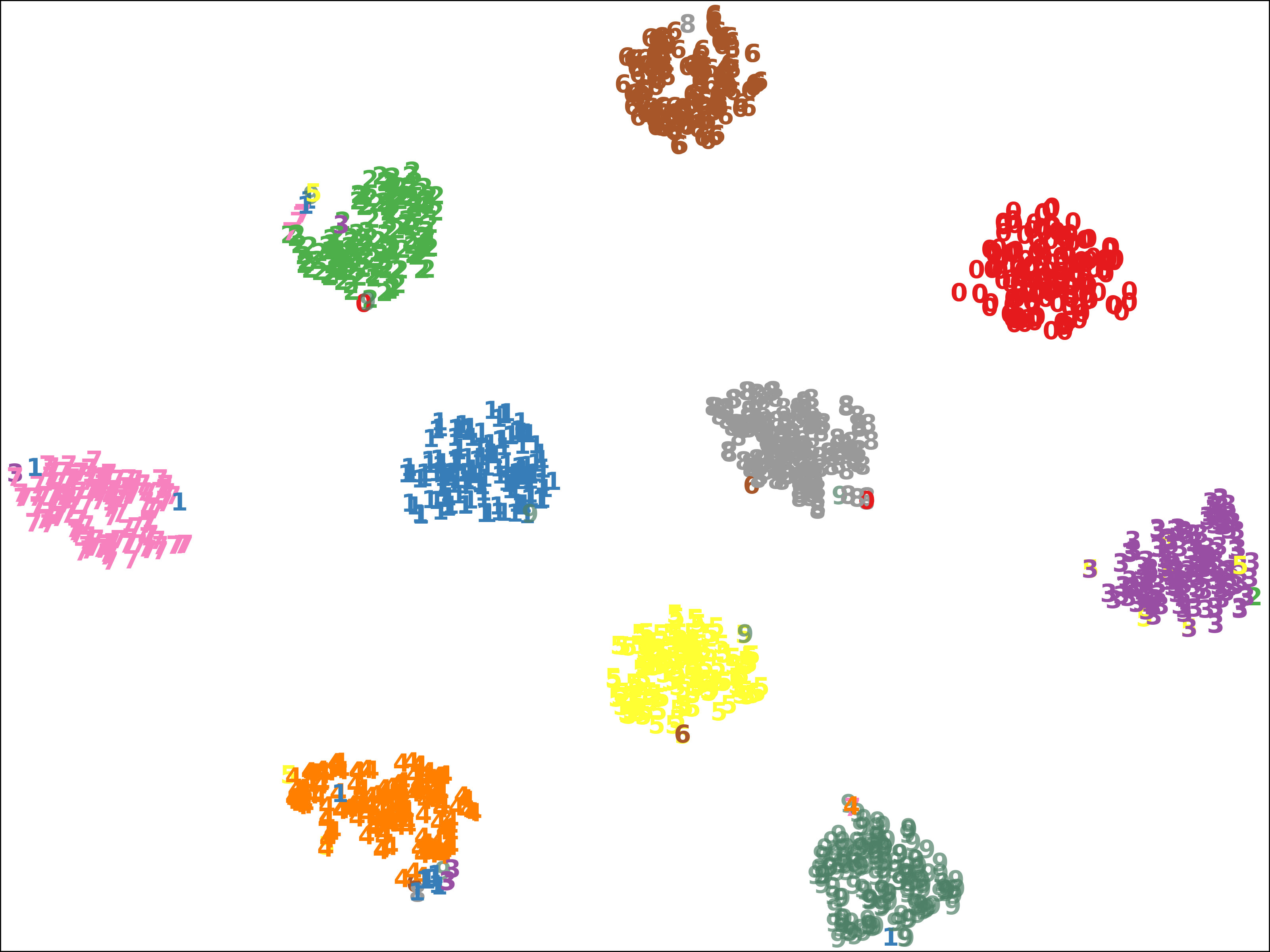}
        \caption{View 3}
        \label{fig:trained_view3}
    \end{subfigure}
    \hfill
    \begin{subfigure}[b]{0.16\textwidth}
        \centering
        \includegraphics[width=\linewidth]{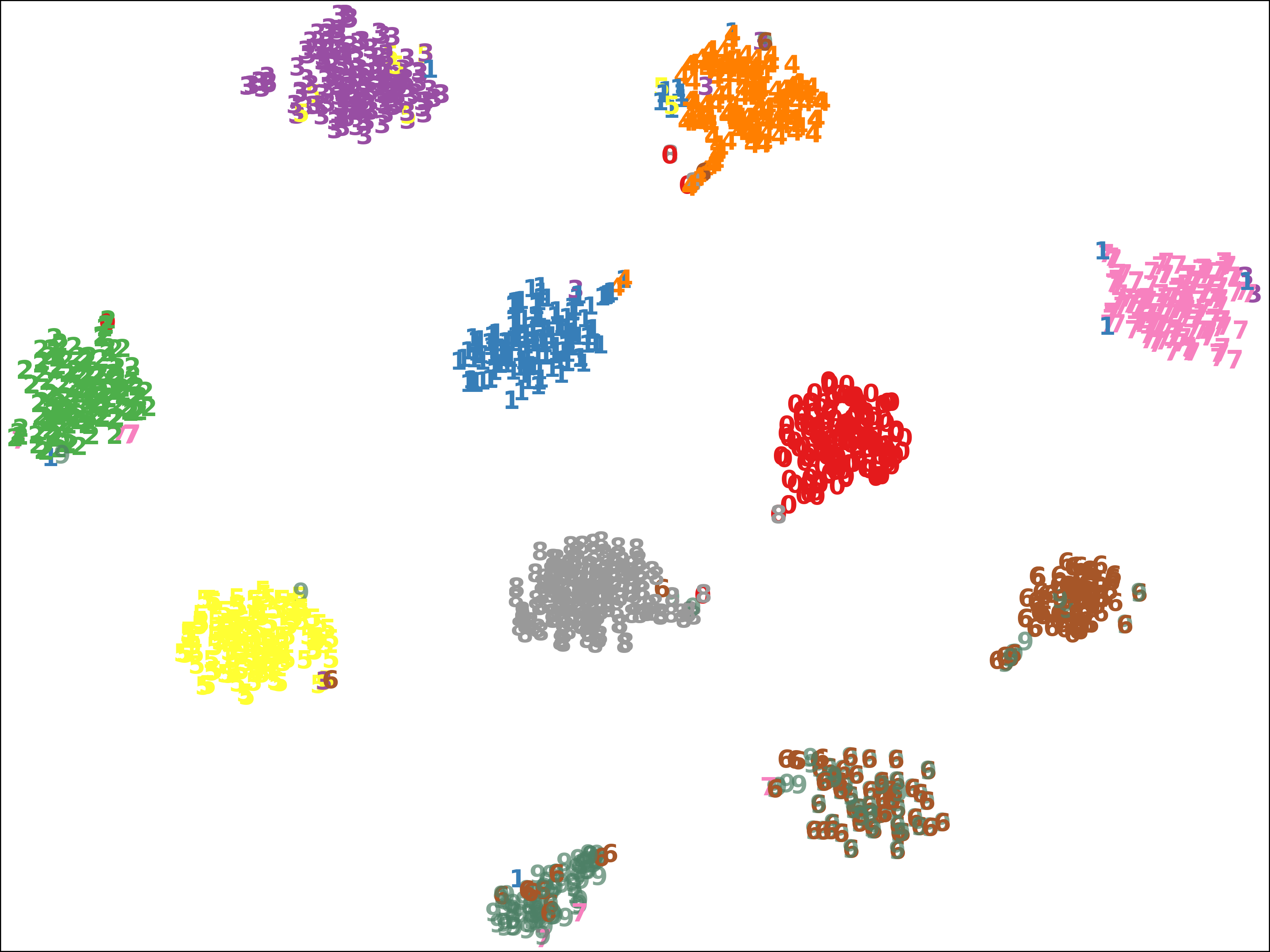}
        \caption{View 4}
        \label{fig:trained_view4}
    \end{subfigure}
    \hfill
    \begin{subfigure}[b]{0.16\textwidth}
        \centering
        \includegraphics[width=\linewidth]{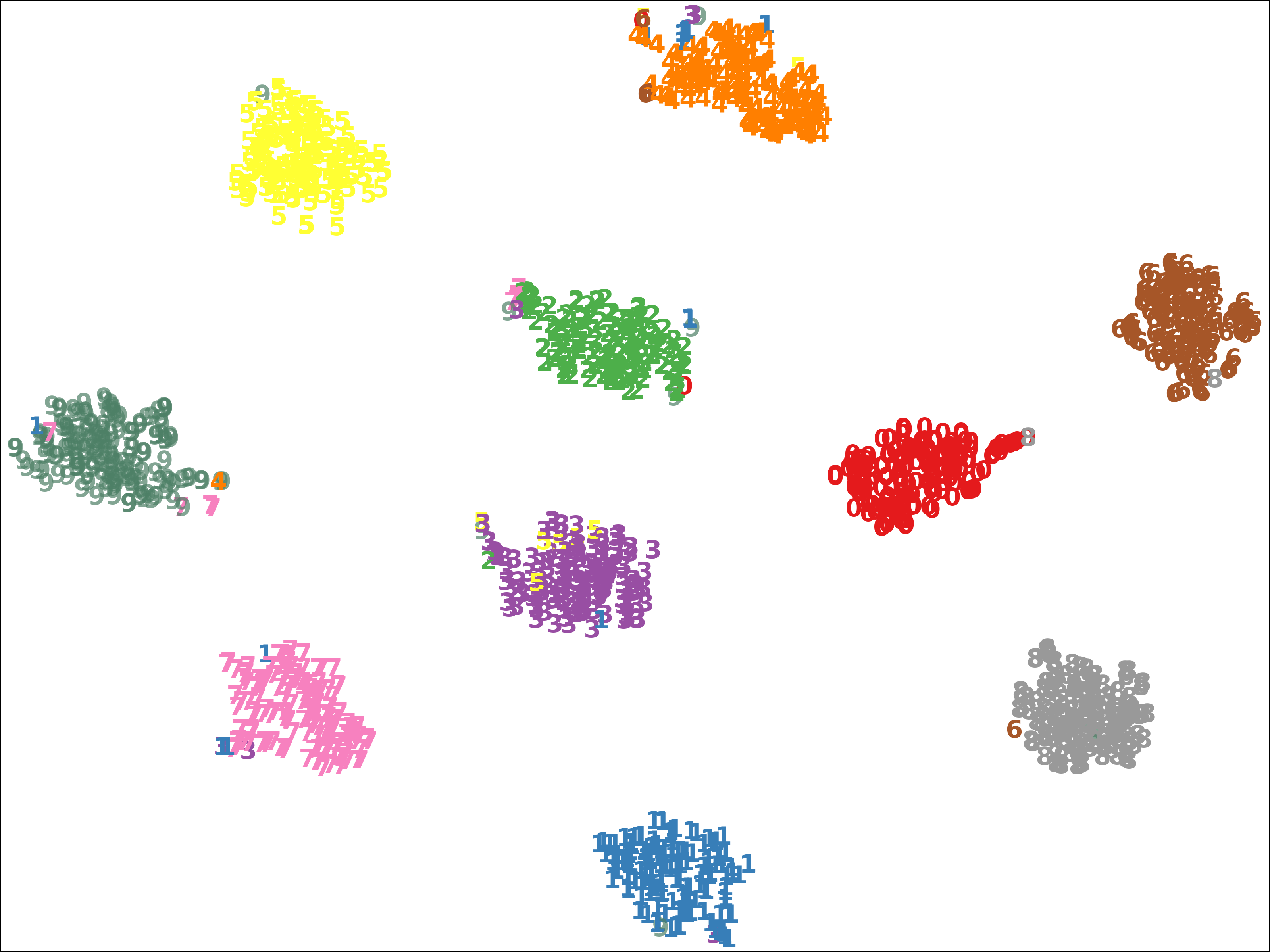}
        \caption{View 5}
        \label{fig:trained_view5}
    \end{subfigure}
    \hfill
    \begin{subfigure}[b]{0.16\textwidth}
        \centering
        \includegraphics[width=\linewidth]{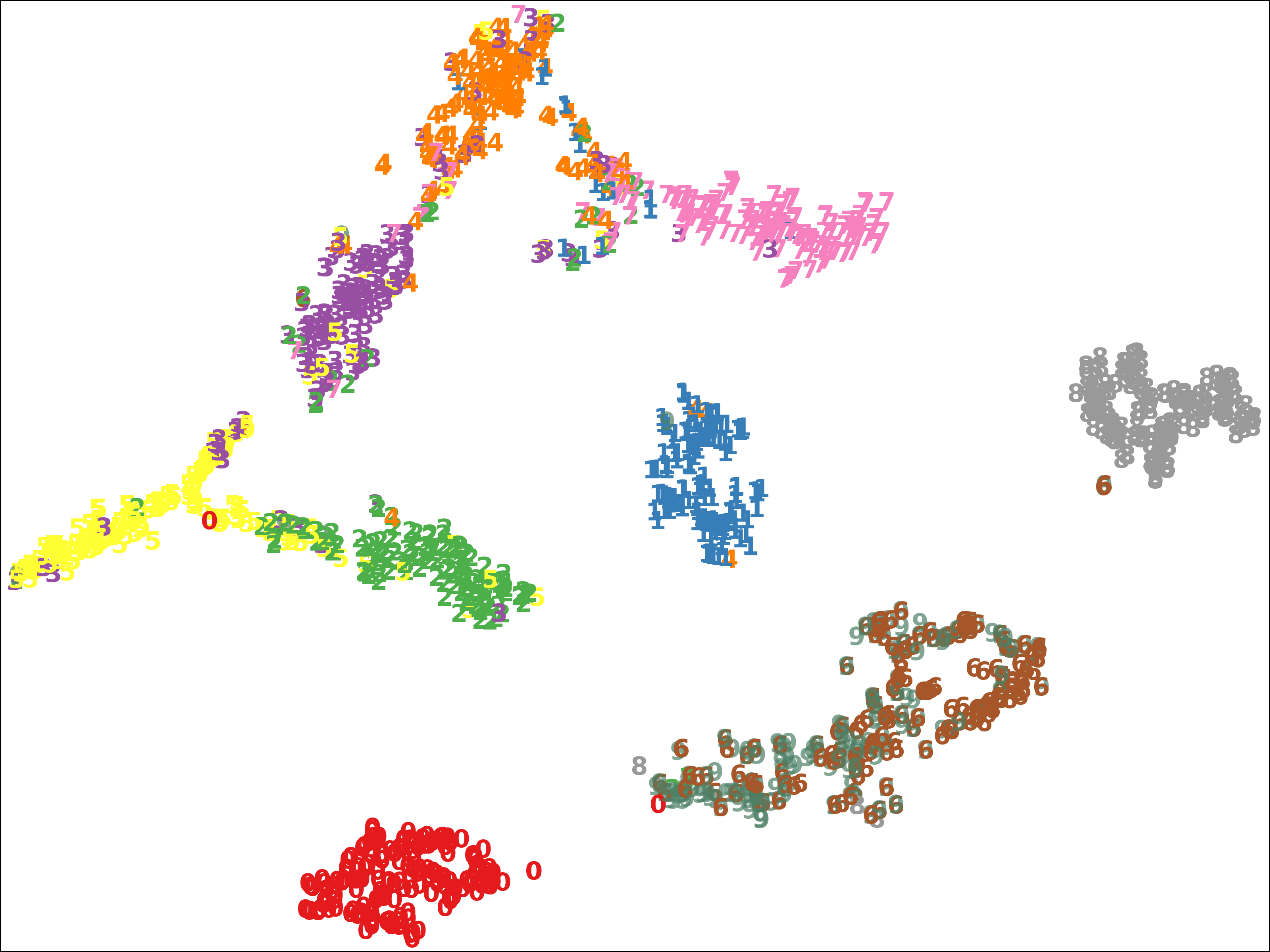}
        \caption{View 6}
        \label{fig:trained_view6}
    \end{subfigure}
    \caption{Representation learned without (first row) or with (second row) our ensemble-to-individual distillation on 6 views of HandWritten.}
    \label{fig:12_views}
\end{figure*}

\begin{figure}[!t]
    \centering
    \begin{subfigure}[b]{0.235\textwidth}
        \centering
        \includegraphics[width=\linewidth]{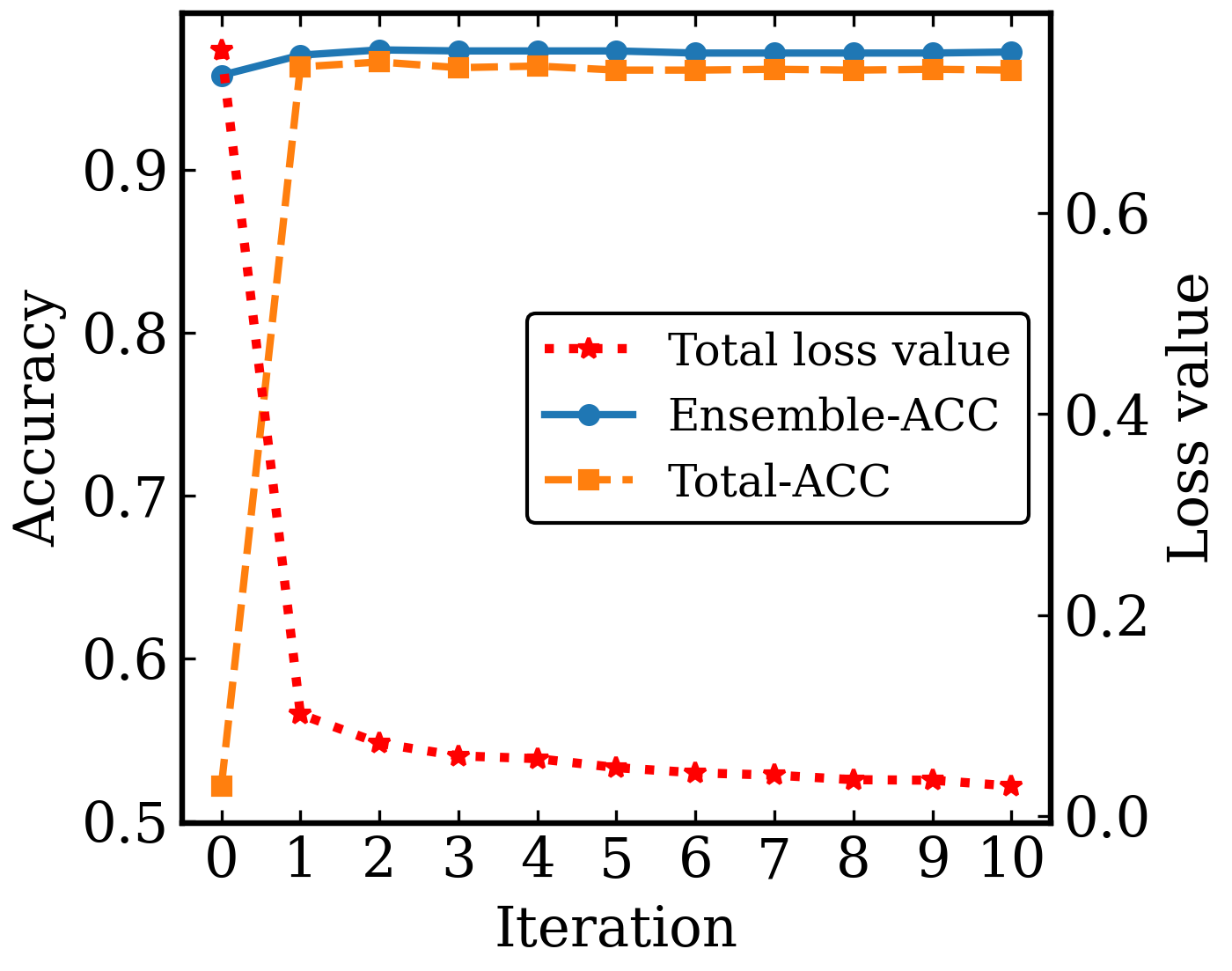}
        \caption{HandWritten}
        \label{fig:handwritten_acc_curve}
    \end{subfigure}
    \hfill
    \begin{subfigure}[b]{0.235\textwidth}
        \centering
        \includegraphics[width=\linewidth]{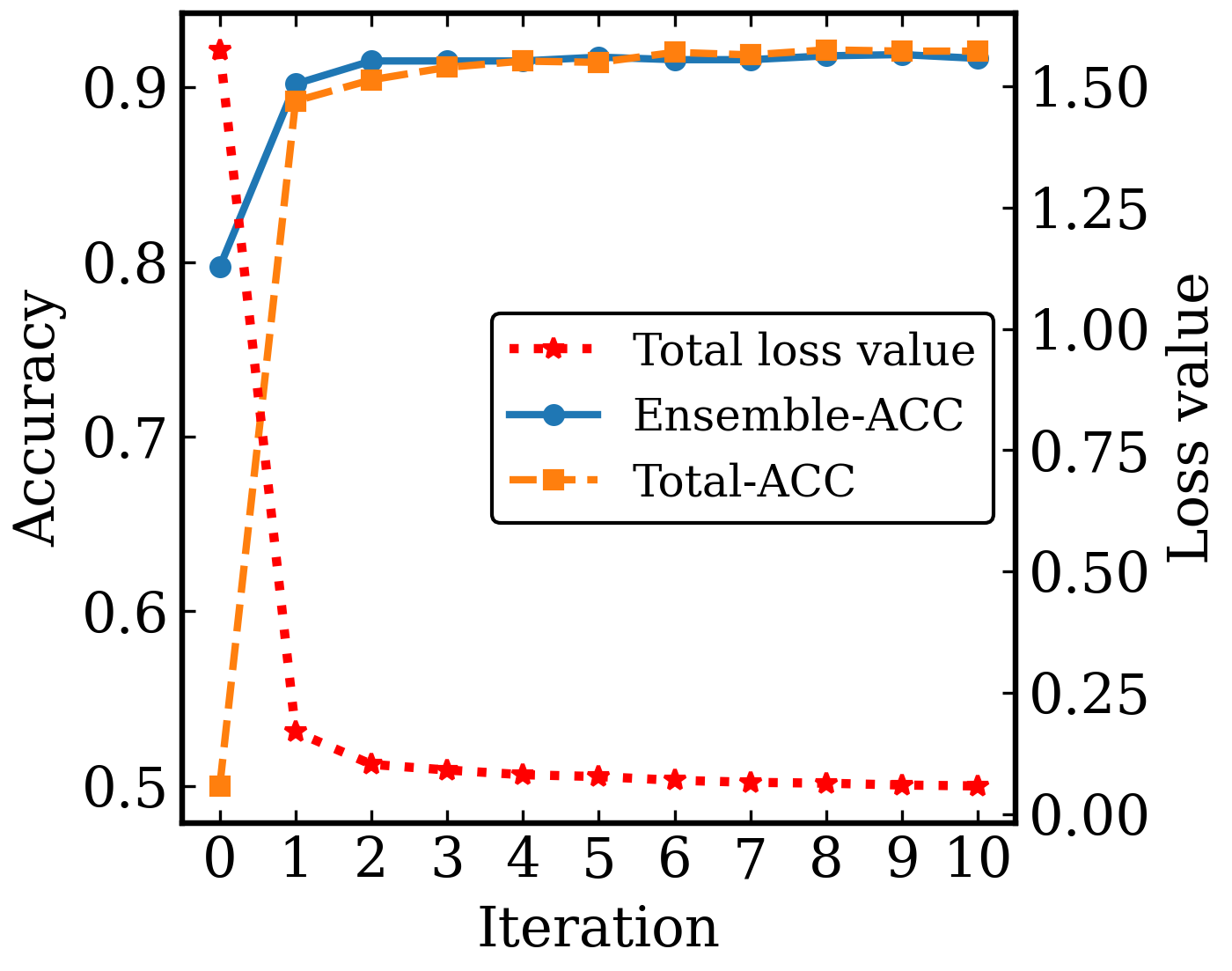}
        \caption{Caltech101-7}
        \label{fig:caltech_acc_curve}
    \end{subfigure}
    \caption{ACC $vs.$ Loss on HandWritten and Caltech101-7.} 
    \label{fig:loss_curve}
\end{figure}

\begin{figure}[!t]
    \centering
    \begin{subfigure}[b]{0.49\columnwidth}
        \centering
        \includegraphics[width=\linewidth]{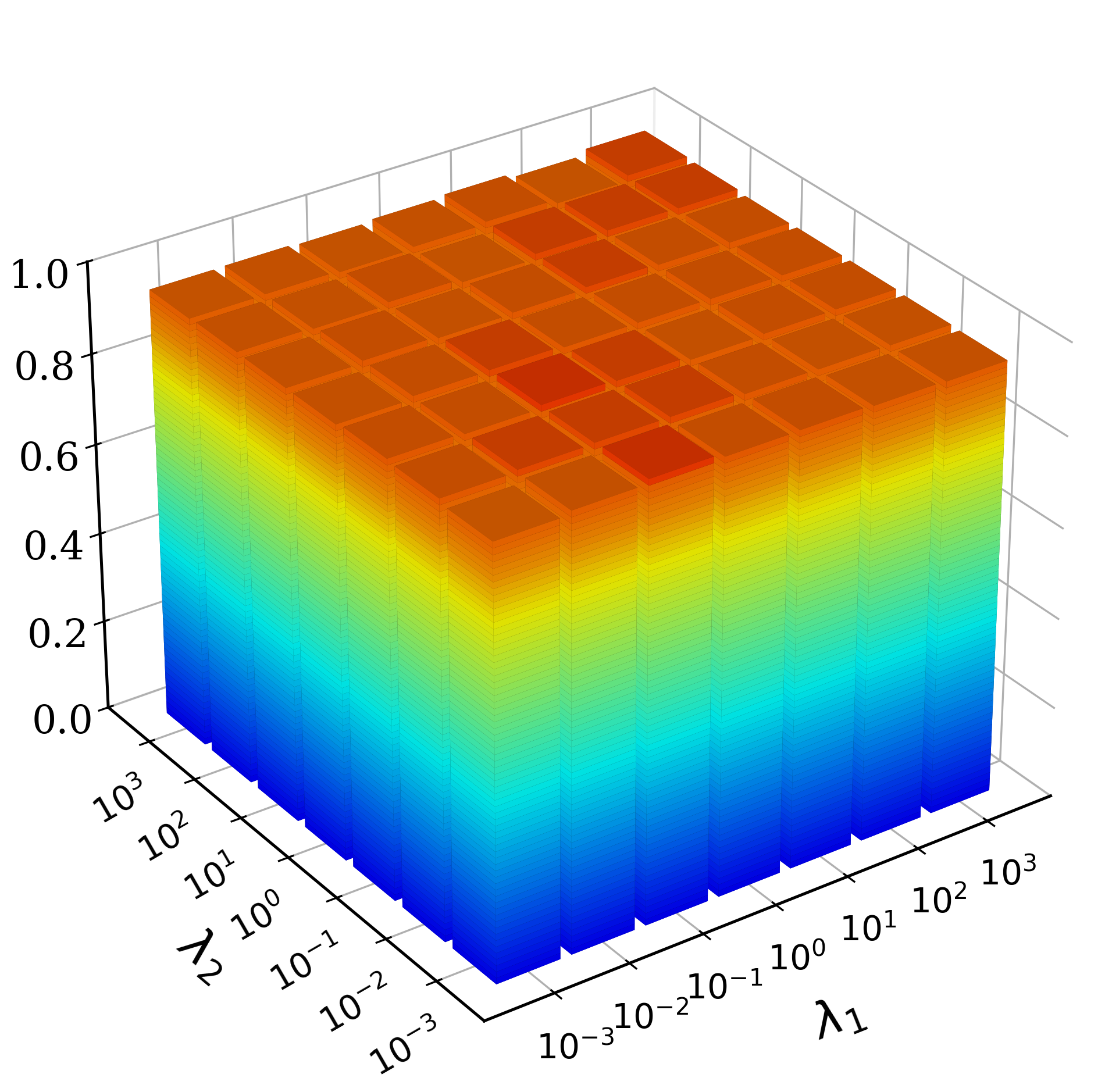}
        \caption{HandWritten}
        \label{fig:handwritten_hot}
    \end{subfigure}
    \hfill
    \begin{subfigure}[b]{0.49\columnwidth}
        \centering
        \includegraphics[width=\linewidth]{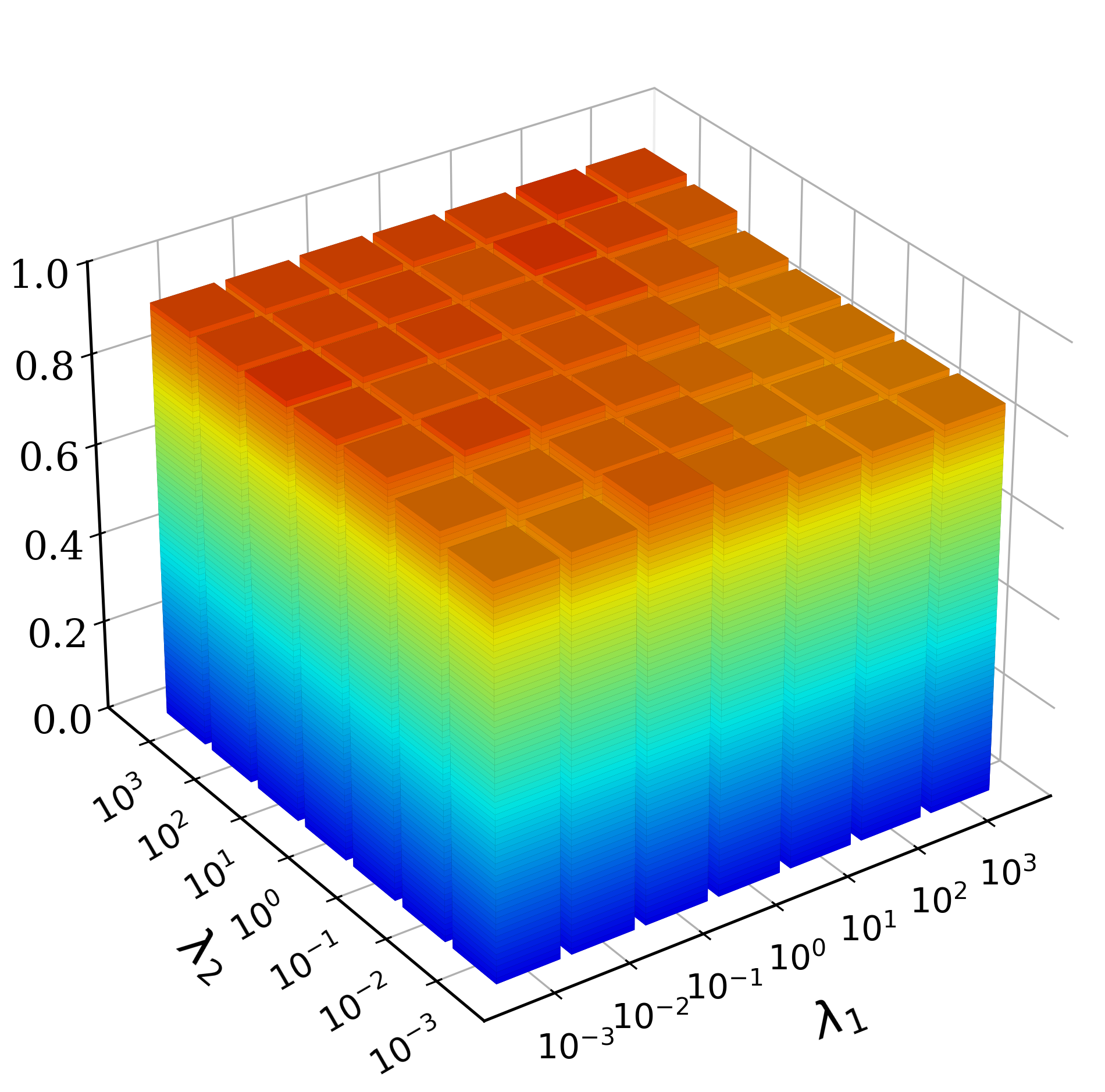}
        \caption{Caltech101-7}
        \label{fig:caltech_hot}
    \end{subfigure}
    \caption{Parameter analysis on HandWritten and Caltech101-7.} 
    \label{fig:parameter_analysis_horizontal}
\end{figure}

\noindent\textbf{Qualitative Comparison.}
As shown in Figure~\ref{fig:acctrend}, all IMVC methods experience performance degradation as the missing rate increases.
This is because higher missing rates result in fewer available multi-view pairs limiting the algorithms' ability to explore complementary information across views.
Moreover, previous methods have overlooked the pair utilization issue, reducing the model's opportunities to mine useful information and causing sharp performance drops at high missing rates, e.g., $\rho > 0.7$.
In contrast, our TreeEIC is a novel missing pattern tree based IMVC method for addressing the pair utilization issue, and thus it can achieve robust clustering performance under high missing rates (even at $\rho = 1.0$).

\subsection{Ablation Studies}\label{sec:ablation}
Our TreeEIC framework can be divided into two main processes, i.e., \emph{Decision Ensemble} and \emph{Knowledge Distillation}.
In this part, we conduct ablation studies on four datasets with the missing rate $\rho=1.0$, to examine the contribution of each component these two processes.

\noindent\textbf{Decision Ensemble in TreeEIC.}
The decision ensemble process has two key designs, including the \emph{Missing Pattern Tree for Multi-View Decision} (MPT) and the \emph{Uncertainty-based Weighing for Multi-View Decision Ensemble} (MDE),
resulting in three model variants as shown in Table~\ref{tab:two_components}.

Specifically,
``without MPT'' represents the variant that directly employs the MDE to obtain ensemble clustering from all view-specific models without performing MPT.
Compared with our full framework (``MPT+MDE''), ``without MPT'' causes significant performance degradation.
For example, on the HandWritten dataset, its ACC and NMI decrease by $10.53\%$ and $13.46\%$ respectively; on the AWA-7 dataset, they drop by $36.39\%$ and $33.78\%$.
This is because the multi-view pairs are not fully utilized without the MPT module, thereby limiting the model performance. 
``without MDE'' represents the variant that applies MPT to group samples but directly averages the clustering labels without performing MDE.
Compared with our full framework (``MPT+MDE''), ``without MDE'' also shows degraded performance.
For example, on the AWA-7 dataset, the ACC and NMI decreases by $5.87\%$ and $4.03\%$.
This is because the MDE module is designed to mitigate the negative influence of ambiguous decisions, and without MDE might make model vulnerable to low-quality data.

\noindent\textbf{Knowledge Distillation in TreeEIC.} 
The knowledge distillation process has two major losses, including the {\emph{Cross-View Consistency} loss ($\mathcal{L}_{cons}$) and the \emph{Inter-Cluster Discrimination} loss ($\mathcal{L}_{disc}$)}.
Table~\ref{tab:two_loss} shows their ablation experiments.

Specifically, the first row of Table~\ref{tab:two_loss} reports the basic result which is obtained by only optimizing the reconstruction loss.
Compared with the basic result, we can observe that both $\mathcal{L}_{cons}$ and $\mathcal{L}_{disc}$ can individually improve the model performance.
For example, on the HandWritten and AWA-7 datasets, the introduction of $ \mathcal{L}_{\text{cons}} $ brings accuracy improvements of $55.67\%$ and $31.85\%$, respectively, while $ \mathcal{L}_{\text{disc}} $ contributes additional gains of $57.12\%$ and $30.15\%$.
This is because the cross-view consistency objective aligns individual view representations with the ensemble decisions, enabling them to learn more clustering-friendly representations.
Meanwhile, the inter-cluster discrimination objective guides each view to refine its own prediction according to the ensemble supervision and enhances inter-cluster separability, thus promoting representation learning.
When our full framework considers both the consistency and discrimination objectives, it achieves further improvements, demonstrating that these two objectives complement each other for IMVC tasks.

\subsection{Model Analysis}\label{sec:analsis}
In this part, we provide a detailed model analysis to gain deeper insights into how our ensemble decision and knowledge distillation enhance clustering under the highly inconsistent missing patterns.

\noindent\textbf{Addressing Pair Underutilization Issue.}
We conduct experiment on HandWritten with $\rho=1.0$ to illustrate the mechanism of our TreeEIC.
In our TreeEIC, the decision ensemble process will perform multi-view clustering within each decision subset $\mathcal{S}_j$ to exploit the cross-view association, and then obtain ensemble clustering results for the unioned samples $\mathcal{U}$.
The average number of available samples $|\mathcal{S}_j|$ in each subset is controlled by the truncation parameter $\tau$ in the missing pattern tree illustrated by Eq.~\ref{eq:prune}.
As shown in Table~\ref{tab:different_tau}, when $\tau=1$ ($1$ view does not constitute pair), each subset contains only single available view samples without any paired information, leading to significantly lower $\text{ACC}_{\mathcal{U}}$ and $\text{NMI}_{\mathcal{U}}$ for the samples in $\mathcal{U}$ compared with the cases with $\tau=2,3,4$ (allowing $2, 3, 4$ views to constitute pairs).
This observation demonstrates that TreeEIC leverages the missing pattern tree based grouping to achieve effective utilization of the paired multi-view data.
Notably, incorporating more paired views has a boundary. 
That is, with the increase of $\tau$, the number of paired multi-view data that satisfy the corresponding missing pattern decreases, which reduces both the average number of samples within each subset $|\mathcal{S}_j|$ and the number of the unioned samples $|\mathcal{U}|$.
Consequently, the inherent multi-view pairs are under-utilized, thereby limiting the model's capability.
To address this, we introduce the missing pattern tree with a truncation parameter to addressing the pair underutilization issue.
$V \ge \tau \ge V/2$ is empirically set to balance the full utilization of paired multi-view data and the practical available pairs.

We also experimentally compare our tree-based grouping strategy against the random partition strategy over the same decision sets.
As shown in Table~\ref{RP}, the clustering performance of random partition (RP) degrade significantly with the increasing of missing rates, while our tree-based grouping (TG) strategy consistently yields reliable clustering decisions, supporting our design choice for addressing the pair underutilization issue.

\noindent\textbf{Loss $vs.$ Clustering Performance.}
In our TreeEIC, the decision ensemble process exploits intrinsic paired information in incomplete multi-view data for robust ensemble clustering, while the knowledge distillation process transfers the robust clustering knowledge to individual views.
This iteration is illustrated in Figure~\ref{fig:loss_curve}, where individual views with initially poor clustering performance (Total-ACC) can leverage the ensemble decisions (Ensemble-ACC) to produce more reliable clustering results.
During the iterative optimization, both Total-ACC and Ensemble-ACC are improved as the total loss decreases.
As a consequence, our method can make the individual views' models learn clustering-friendly representations as illustrated in Figure~\ref{fig:12_views}.
These observations demonstrate that TreeEIC effectively achieves iterative optimization through the collaboration of the decision ensemble and knowledge distillation processes.

\noindent\textbf{Parameter Analysis.}
We investigate the sensitivity of the two balance parameters $\lambda_1$ and $\lambda_2$ in Eq.~\ref{eq:total_loss} during the knowledge distillation process.
Figure~\ref{fig:parameter_analysis_horizontal} visualizes the results as a heatmap by varying the parameters within the range $[10^{-3}, 10^{-2}, 10^{-1}, 10^{0}, 10^{1}, 10^{2}, 10^{3}]$.
As observed, the knowledge distillation process exhibits relative insensitivity to $\lambda_1$ and $\lambda_2$.
For the two parameters, the range of $[10^{-1},10^{1}]$ can will balance the multi-view consistency and clustering discrimination objectives.

\noindent\textbf{Computational Cost.}
In Table~\ref{Cost}, we provide the experimental results to empirically verify the computational complexity and memory cost for the proposed framework, which suggests that the runtime and memory usage of our method grows linearly with the sample number $N$, consistent with our complexity analysis.

\begin{table}[!t]
\centering
\renewcommand\tabcolsep{3pt} 
\caption{Random partition (RP) $vs.$ Tree-based grouping (TG) on HandWritten over different missing rates $\rho=[0.1,0.3,0.5,0.7,1.0]$.}\label{RP}
\resizebox{0.48\textwidth}{!}{
\begin{threeparttable}
    \begin{tabular}{l|cc|cc|cc|cc|cc}
    \toprule[2pt]
     & \multicolumn{2}{c|}{$\rho=0.1$} & \multicolumn{2}{c|}{$\rho=0.3$} & \multicolumn{2}{c|}{$\rho=0.5$} & \multicolumn{2}{c|}{$\rho=0.7$} & \multicolumn{2}{c}{$\rho=1.0$}\cr
    \hline
     & ACC & NMI & ACC & NMI & ACC & NMI & ACC & NMI & ACC & NMI \\
    \hline
    RP    & 95.05  & 90.52 & 82.70 & 77.91 & 79.70 & 76.08 & 72.35 & 65.19 & 61.55 & 57.32 \\
    TG    &97.20  & 93.77 & 96.69 & 92.75 & 95.86 & 91.14 & 94.41 & 88.54 & 93.77 & 87.09 \\
    \bottomrule[2pt]
    \end{tabular}
\end{threeparttable}
}
\end{table}
\begin{table}[!t]
\centering
\caption{Computational runtime and memory cost.}\label{Cost}
\renewcommand\tabcolsep{4pt} 
\resizebox{0.48\textwidth}{!}{
\begin{threeparttable}
    \begin{tabular}{c|c|c|c|c}
    \toprule[2pt]
     & \multicolumn{1}{c|}{HandWritten} & \multicolumn{1}{c|}{Caltech101-7} & \multicolumn{1}{c|}{OutdoorScene} & \multicolumn{1}{c}{AWA-7}\cr
    \hline
    samples ($N$) & 2,000  & 1,400  & 2,688  & 10,158  \\
    views ($V$)   & 6      & 5     & 4     & 7 \\
    \hline
    runtime (min) & 11   & 6    & 8    & 67  \\
    memory (GB)   & 0.50 & 0.43 & 0.41 & 4.25 \\
    \bottomrule[2pt]
    \end{tabular}
\end{threeparttable}
}
\end{table}

\section{Conclusion}\label{Conclusion}
In this paper, we propose a novel imputation-free IMVC framework \textbf{TreeEIC}, based on the innovative concepts of missing pattern tree and ensemble decisions designed for deep IMVC domain.
TreeEIC first groups samples into multiple decision subsets based on their missing patterns, ensuring that the samples within each subset share a consistent missing pattern. 
Multi-view clustering is then performed within each subset to fully exploit the available multi-view pairs, addressing the pair underutilization issue overlooked by previous IMVC methods.
The multiple decisions are then aggregated via an uncertainty-weighted ensemble to produce robust ensemble clustering results, which are further distilled back to individual views to transfer ensemble knowledge.
Experiments demonstrate that TreeEIC achieves superior clustering performance across six diverse datasets under high missing rates.
The future work may investigate more robust and efficient methods for addressing the pair underutilization issue in IMVC, and extend this insight to other multi-view learning domains.


{
\small
\bibliographystyle{unsrt}
\bibliography{main}

\begin{thebibliography}{10}

\bibitem{chaudhuri2009multi}
Kamalika Chaudhuri, Sham~M Kakade, Karen Livescu, and Karthik Sridharan.
\newblock Multi-view clustering via canonical correlation analysis.
\newblock In {\em ICML}, pages 129--136, 2009.

\bibitem{kumar2011co}
Abhishek Kumar, Piyush Rai, and Hal Daume.
\newblock Co-regularized multi-view spectral clustering.
\newblock {\em Advances in neural information processing systems}, 24, 2011.

\bibitem{lin2025multi}
Fangfei Lin, Jie Xu, Yazhou Ren, Junjie Chen, Irwin King, and Zenglin Xu.
\newblock Multi-modal hierarchical clustering network for cancer subtype identification of multi-omics data.
\newblock In {\em 2025 International Joint Conference on Neural Networks (IJCNN)}, pages 1--8, 2025.

\bibitem{9782584}
Peng Hu, Hongyuan Zhu, Jie Lin, Dezhong Peng, Yin-Ping Zhao, and Xi~Peng.
\newblock Unsupervised contrastive cross-modal hashing.
\newblock {\em TPAMI}, 45(3):3877--3889, 2023.

\bibitem{vazquez2020multigraph}
Miguel~Angel V{\'a}zquez and Ana~I P{\'e}rez-Neira.
\newblock Multigraph spectral clustering for joint content delivery and scheduling in beam-free satellite communications.
\newblock In {\em ICASSP 2020-2020 IEEE International Conference on Acoustics, Speech and Signal Processing (ICASSP)}, pages 8802--8806, 2020.

\bibitem{wen2023highly}
Jie Wen, Chengliang Liu, Gehui Xu, Zhihao Wu, Chao Huang, Lunke Fei, and Yong Xu.
\newblock Highly confident local structure based consensus graph learning for incomplete multi-view clustering.
\newblock In {\em CVPR}, pages 15712--15721, 2023.

\bibitem{wan2024fast}
Xinhang Wan, Bin Xiao, Xinwang Liu, Jiyuan Liu, Weixuan Liang, and En~Zhu.
\newblock Fast continual multi-view clustering with incomplete views.
\newblock {\em TIP}, 33:2995--3008, 2024.

\bibitem{Jiang_2025_ICCV}
Xiaorui Jiang, Buyun He, Peng~Yuan Zhou, Xinyue Chen, Jingcai Guo, Jie Xu, and Yong Liao.
\newblock A unified framework to bridge complete and incomplete deep multi-view clustering under non-iid missing patterns.
\newblock In {\em ICCV}, pages 594--603, 2025.

\bibitem{wen2018incomplete}
Jie Wen, Zheng Zhang, Yong Xu, and Zuofeng Zhong.
\newblock Incomplete multi-view clustering via graph regularized matrix factorization.
\newblock In {\em Proceedings of the European conference on computer vision (ECCV) workshops}, pages 0--0, 2018.

\bibitem{wang2022highly}
Siwei Wang, Xinwang Liu, Li~Liu, Wenxuan Tu, Xinzhong Zhu, Jiyuan Liu, Sihang Zhou, and En~Zhu.
\newblock Highly-efficient incomplete large-scale multi-view clustering with consensus bipartite graph.
\newblock In {\em CVPR}, pages 9776--9785, 2022.

\bibitem{wang2020icmsc}
Qianqian Wang, Huanhuan Lian, Gan Sun, Quanxue Gao, and Licheng Jiao.
\newblock icmsc: Incomplete cross-modal subspace clustering.
\newblock {\em TIP}, 30:305--317, 2020.

\bibitem{wen2022survey}
Jie Wen, Zheng Zhang, Lunke Fei, Bob Zhang, Yong Xu, Zhao Zhang, and Jinxing Li.
\newblock A survey on incomplete multiview clustering.
\newblock {\em IEEE Transactions on Systems, Man, and Cybernetics: Systems}, 53(2):1136--1149, 2022.

\bibitem{wen2020dimc}
Jie Wen, Zheng Zhang, Zhao Zhang, Zhihao Wu, Lunke Fei, Yong Xu, and Bob Zhang.
\newblock Dimc-net: Deep incomplete multi-view clustering network.
\newblock In {\em ACM MM}, pages 3753--3761, 2020.

\bibitem{xue2021clustering}
Zhe Xue, Junping Du, Changwei Zheng, Jie Song, Wenqi Ren, and Meiyu Liang.
\newblock Clustering-induced adaptive structure enhancing network for incomplete multi-view data.
\newblock In {\em IJCAI}, pages 3235--3241, 2021.

\bibitem{zhang2026structure}
Yuanyang Zhang, Yijie Lin, Xinhang Wan, Jie Xu, Li~Yao, Weiqing Yan, and Chang Tang.
\newblock Structure-aware conditional diffusion generation for incomplete multi-view clustering.
\newblock {\em TKDE}, 2026.

\bibitem{yan2024deep}
Weiqing Yan, Kanglong Liu, Wujie Zhou, and Chang Tang.
\newblock Deep incomplete multi-view clustering via dynamic imputation and triple alignment with dual optimization.
\newblock {\em TCSVT}, 2024.

\bibitem{wen2020adaptive}
Jie Wen, Ke~Yan, Zheng Zhang, Yong Xu, Junqian Wang, Lunke Fei, and Bob Zhang.
\newblock Adaptive graph completion based incomplete multi-view clustering.
\newblock {\em TMM}, 23:2493--2504, 2020.

\bibitem{Xu_2022_AAAI}
Jie Xu, Chao Li, Yazhou Ren, Liang Peng, Yujie Mo, Xiaoshuang Shi, and Xiaofeng Zhu.
\newblock Deep incomplete multi-view clustering via mining cluster complementarity.
\newblock In {\em AAAI}, pages 8761--8769, 2022.

\bibitem{sun2025roll}
Yuan Sun, Yongxiang Li, Zhenwen Ren, Guiduo Duan, Dezhong Peng, and Peng Hu.
\newblock Roll: Robust noisy pseudo-label learning for multi-view clustering with noisy correspondence.
\newblock In {\em CVPR}, pages 30732--30741, 2025.

\bibitem{xu2023adaptive}
Jie Xu, Chao Li, Liang Peng, Yazhou Ren, Xiaoshuang Shi, Heng~Tao Shen, and Xiaofeng Zhu.
\newblock Adaptive feature projection with distribution alignment for deep incomplete multi-view clustering.
\newblock {\em TIP}, 32:1354--1366, 2023.

\bibitem{dai2025imputation}
Yuzhuo Dai, Jiaqi Jin, Zhibin Dong, Siwei Wang, Xinwang Liu, En~Zhu, Xihong Yang, Xinbiao Gan, and Yu~Feng.
\newblock Imputation-free and alignment-free: Incomplete multi-view clustering driven by consensus semantic learning.
\newblock In {\em CVPR}, pages 5071--5081, 2025.

\bibitem{dietterich2002ensemble}
Thomas~G Dietterich et~al.
\newblock Ensemble learning.
\newblock {\em The handbook of brain theory and neural networks}, 2(1):110--125, 2002.

\bibitem{nie2017self}
Feiping Nie, Jing Li, and Xuelong Li.
\newblock Self-weighted multiview clustering with multiple graphs.
\newblock In {\em IJCAI}, pages 2564--2570, 2017.

\bibitem{peng2019comic}
Xi~Peng, Zhenyu Huang, Jiancheng Lv, Hongyuan Zhu, and Joey~Tianyi Zhou.
\newblock {COMIC}: Multi-view clustering without parameter selection.
\newblock In {\em ICML}, pages 5092--5101, 2019.

\bibitem{8502831}
Changqing Zhang, Huazhu Fu, Qinghua Hu, Xiaochun Cao, Yuan Xie, Dacheng Tao, and Dong Xu.
\newblock Generalized latent multi-view subspace clustering.
\newblock {\em TPAMI}, 42(1):86--99, 2020.

\bibitem{NIE2020107207}
Feiping Nie, Shaojun Shi, and Xuelong Li.
\newblock Auto-weighted multi-view co-clustering via fast matrix factorization.
\newblock {\em Pattern Recognition}, 102:107207, 2020.

\bibitem{8611131}
Xinwang Liu, Xinzhong Zhu, Miaomiao Li, Lei Wang, En~Zhu, Tongliang Liu, Marius Kloft, Dinggang Shen, Jianping Yin, and Wen Gao.
\newblock Multiple kernel $k$k-means with incomplete kernels.
\newblock {\em TPAMI}, 42(5):1191--1204, 2020.

\bibitem{11370248}
Zhibin Gu and Songhe Feng.
\newblock Twin tensor learning for consistency and inconsistency: A unified affinity learning framework for multi-view clustering.
\newblock {\em TMM}, pages 1--12, 2026.

\bibitem{gu2024edison}
Zhibin Gu, Zhendong Li, and Songhe Feng.
\newblock Edison: Enhanced dictionary-induced tensorized incomplete multi-view clustering with gaussian error rank minimization.
\newblock In {\em ICML}, pages 16548--16567, 2024.

\bibitem{10595464}
Yuan Sun, Yang Qin, Yongxiang Li, Dezhong Peng, Xi~Peng, and Peng Hu.
\newblock Robust multi-view clustering with noisy correspondence.
\newblock {\em TKDE}, 36(12):9150--9162, 2024.

\bibitem{li2014partial}
Shao-Yuan Li, Yuan Jiang, and Zhi-Hua Zhou.
\newblock Partial multi-view clustering.
\newblock In {\em AAAI}, pages 1968--1974, 2014.

\bibitem{xu2015multi}
Chang Xu, Dacheng Tao, and Chao Xu.
\newblock Multi-view learning with incomplete views.
\newblock {\em TIP}, 24(12):5812--5825, 2015.

\bibitem{liu2020efficient}
Xinwang Liu, Miaomiao Li, Chang Tang, Jingyuan Xia, Jian Xiong, Li~Liu, Marius Kloft, and En~Zhu.
\newblock Efficient and effective regularized incomplete multi-view clustering.
\newblock {\em TPAMI}, 43(8):2634--2646, 2020.

\bibitem{wen2020cdimc}
Jie Wen, Zheng Zhang, Yong Xu, Bob Zhang, Lunke Fei, and Guo-Sen Xie.
\newblock {CDIMC}-net: Cognitive deep incomplete multi-view clustering network.
\newblock In {\em IJCAI}, pages 3230--3236, 2020.

\bibitem{wang2021generative}
Qianqian Wang, Zhengming Ding, Zhiqiang Tao, Quanxue Gao, and Yun Fu.
\newblock Generative partial multi-view clustering with adaptive fusion and cycle consistency.
\newblock {\em TIP}, 30:1771--1783, 2021.

\bibitem{lin2021completer}
Yijie Lin, Yuanbiao Gou, Zitao Liu, Boyun Li, Jiancheng Lv, and Xi~Peng.
\newblock Completer: Incomplete multi-view clustering via contrastive prediction.
\newblock In {\em CVPR}, pages 11174--11183, 2021.

\bibitem{wang2018partial}
Qianqian Wang, Zhengming Ding, Zhiqiang Tao, Quanxue Gao, and Yun Fu.
\newblock Partial multi-view clustering via consistent gan.
\newblock In {\em ICDM}, pages 1290--1295, 2018.

\bibitem{yan2023gcfagg}
Weiqing Yan, Yuanyang Zhang, Chenlei Lv, Chang Tang, Guanghui Yue, Liang Liao, and Weisi Lin.
\newblock {GCFAgg}: Global and cross-view feature aggregation for multi-view clustering.
\newblock In {\em CVPR}, pages 19863--19872, 2023.

\bibitem{tang2022deep}
Huayi Tang and Yong Liu.
\newblock Deep safe multi-view clustering: Reducing the risk of clustering performance degradation caused by view increase.
\newblock In {\em CVPR}, pages 202--211, 2022.

\bibitem{Hinton2006Reducing}
Geoffrey~E Hinton and Ruslan~R Salakhutdinov.
\newblock Reducing the dimensionality of data with neural networks.
\newblock {\em Science}, 313(5786):504--507, 2006.

\bibitem{oord2018representation}
Aaron van~den Oord, Yazhe Li, and Oriol Vinyals.
\newblock Representation learning with contrastive predictive coding.
\newblock {\em arXiv preprint arXiv:1807.03748}, 2018.

\bibitem{xu2022self}
Jie Xu, Yazhou Ren, Huayi Tang, Zhimeng Yang, Lili Pan, Yang Yang, Xiaorong Pu, Philip~S Yu, and Lifang He.
\newblock Self-supervised discriminative feature learning for deep multi-view clustering.
\newblock {\em TKDE}, 35(7):7470--7482, 2023.

\bibitem{zhang2017latent}
Changqing Zhang, Qinghua Hu, Huazhu Fu, Pengfei Zhu, and Xiaochun Cao.
\newblock Latent multi-view subspace clustering.
\newblock In {\em CVPR}, pages 4279--4287, 2017.

\bibitem{8387808}
Xi~Peng, Jiashi Feng, Shijie Xiao, Wei-Yun Yau, Joey~Tianyi Zhou, and Songfan Yang.
\newblock Structured autoencoders for subspace clustering.
\newblock {\em TIP}, 27(10):5076--5086, 2018.

\bibitem{tang2022deepi}
Huayi Tang and Yong Liu.
\newblock Deep safe incomplete multi-view clustering: Theorem and algorithm.
\newblock In {\em ICML}, pages 21090--21110, 2022.

\bibitem{9852291}
Yijie Lin, Yuanbiao Gou, Xiaotian Liu, Jinfeng Bai, Jiancheng Lv, and Xi~Peng.
\newblock Dual contrastive prediction for incomplete multi-view representation learning.
\newblock {\em TPAMI}, 45(4):4447--4461, 2023.

\bibitem{macqueen1967some}
James MacQueen.
\newblock Some methods for classification and analysis of multivariate observations.
\newblock In {\em Proceedings of the Berkeley Symposium on Mathematical Statistics and Probability}, pages 281--297, 1967.

\bibitem{jonker1986improving}
Roy Jonker and Ton Volgenant.
\newblock Improving the hungarian assignment algorithm.
\newblock {\em Operations Research Letters}, 5(4):171--175, 1986.

\bibitem{lecun1989backpropagation}
Yann LeCun, Bernhard Boser, John~S Denker, Donnie Henderson, Richard~E Howard, Wayne Hubbard, and Lawrence~D Jackel.
\newblock Backpropagation applied to handwritten zip code recognition.
\newblock {\em Neural computation}, 1(4):541--551, 1989.

\bibitem{fei2004learning}
Li~Fei-Fei, Rob Fergus, and Pietro Perona.
\newblock Learning generative visual models from few training examples: An incremental bayesian approach tested on 101 object categories.
\newblock In {\em CVPR}, pages 178--178, 2004.

\bibitem{hu2020multi}
Zhanxuan Hu, Feiping Nie, Rong Wang, and Xuelong Li.
\newblock Multi-view spectral clustering via integrating nonnegative embedding and spectral embedding.
\newblock {\em Information Fusion}, 55:251--259, 2020.

\bibitem{romera2015embarrassingly}
Bernardino Romera-Paredes and Philip Torr.
\newblock An embarrassingly simple approach to zero-shot learning.
\newblock In {\em ICML}, pages 2152--2161. PMLR, 2015.

\bibitem{wu20153d}
Zhirong Wu, Shuran Song, Aditya Khosla, Fisher Yu, Linguang Zhang, Xiaoou Tang, and Jianxiong Xiao.
\newblock 3d shapenets: A deep representation for volumetric shapes.
\newblock In {\em CVPR}, pages 1912--1920, 2015.

\bibitem{pan2021variational}
Liang Pan, Xinyi Chen, Zhongang Cai, Junzhe Zhang, Haiyu Zhao, Shuai Yi, and Ziwei Liu.
\newblock Variational relational point completion network.
\newblock In {\em CVPR}, pages 8524--8533, 2021.

\bibitem{li2023incomplete}
Haobin Li, Yunfan Li, Mouxing Yang, Peng Hu, Dezhong Peng, and Xi~Peng.
\newblock Incomplete multi-view clustering via prototype-based imputation.
\newblock In {\em IJCAI}, pages 3911--3919, 2023.

\bibitem{chao2024incomplete}
Guoqing Chao, Yi~Jiang, and Dianhui Chu.
\newblock Incomplete contrastive multi-view clustering with high-confidence guiding.
\newblock In {\em AAAI}, pages 11221--11229, 2024.

\bibitem{zhang2025incomplete}
Yuanyang Zhang, Yijie Lin, Weiqing Yan, Li~Yao, Xinhang Wan, Guangyuan Li, Chao Zhang, Guanzhou Ke, and Jie Xu.
\newblock Incomplete multi-view clustering via diffusion contrastive generation.
\newblock In {\em AAAI}, pages 22650--22658, 2025.

\bibitem{chao2025global}
Guoqing Chao, Kaixin Xu, Xijiong Xie, and Yongyong Chen.
\newblock Global graph propagation with hierarchical information transfer for incomplete contrastive multi-view clustering.
\newblock In {\em AAAI}, pages 15713--15721, 2025.

\bibitem{kingma2014adam}
Diederik~P Kingma and Jimmy Ba.
\newblock Adam: {A} method for stochastic optimization.
\newblock {\em arXiv preprint arXiv:1412.6980}, 2014.

\bibitem{krogh1995neural}
Anders Krogh and Jesper Vedelsby.
\newblock Neural network ensembles, cross validation, and active learning.
\newblock {\em Advances in neural information processing systems}, 7:231--238, 1994.

\bibitem{gradshteyn2014table}
Izrail~Solomonovich Gradshteyn and Iosif~Moiseevich Ryzhik.
\newblock {\em Table of integrals, series, and products}.
\newblock Academic press, 2014.

\end{thebibliography}
}

\iftrue

\newpage

\onecolumn

\setcounter{figure}{0}
\setcounter{table}{0}
\setcounter{equation}{0}

\section*{Appendix I: Theoretical Analysis and Proof}

\subsection{Missing-Pattern Tree based Grouping}
In this part, we attempt to provide a formal illustration of missing-pattern tree, and then analyze its advantages in utilizing pairs from a theoretical perspective.

\noindent\textbf{Notations.}
Let \(\mathcal{I} = \{1,2,\dots,N\}\) be the set of sample indices and \(\mathcal{V} = \{1,2,\dots,V\}\) the set of view indices.
For each sample \(i \in \mathcal{I}\), define its \emph{available view set} as
\[
\mathcal{V}_i = \{ v \in \mathcal{V} \mid a_i^v = 1 \},
\]
where \(a_i^v \in \{0,1\}\) is the missing indicator (\(a_i^v = 1\) iff the \(v\)-th view is observed for sample \(i\)).
The \emph{view-pair set} of sample \(i\) is
\[
\mathcal{P}_i = \bigl\{ \{u,v\} \subseteq \mathcal{V}_i \mid u \neq v \bigr\},
\]
and its cardinality is \(p_i = |\mathcal{P}_i| = \binom{|\mathcal{V}_i|}{2}\).
The collection of all view-pairs over all samples forms a multiset
\[
\mathfrak{P} = \biguplus_{i \in \mathcal{I}} \bigl( \{i\} \times \mathcal{P}_i \bigr),
\]
with total count \(|\mathfrak{P}| = \sum_{i \in \mathcal{I}} p_i\).

Given a truncation parameter \(\tau \in [2, V]\) (which is adaptively determined by Eq.~(5) based on the missing rate), we define the \emph{family of admissible missing patterns} as
\[
\mathfrak{M}_\tau = \{ M \subseteq \mathcal{V} \mid |M| = \tau \},
\]
i.e., all \(\tau\)-element subsets of \(\mathcal{V}\).
For each pattern \(M \in \mathfrak{M}_\tau\), the corresponding \emph{decision set} is
\[
\mathcal{D}_M = \{ i \in \mathcal{I} \mid M \subseteq \mathcal{V}_i \}.
\]

In the TreeEIC framework, a view-pair \(\{u,v\} \subseteq \mathcal{V}_i\) is said to be \emph{utilized} if there exists a pattern \(M \in \mathfrak{M}_\tau\) such that
\[
\{u,v\} \subseteq M \subseteq \mathcal{V}_i.
\]
The multiset of all utilized sample-view-pair tuples is denoted by
\[
\mathfrak{U}_\tau = \Bigl\{ (i,\{u,v\}) \in \mathfrak{P} \;\Big|\; \exists M \in \mathfrak{M}_\tau:\ \{u,v\} \subseteq M \subseteq \mathcal{V}_i \Bigr\},
\]
and its cardinality is denoted by \(U(\tau) := |\mathfrak{U}_\tau|\), which measures the total number of view-pairs exploited by TreeEIC for a given \(\tau\).

For comparison, the conventional approach (which performs multi-view learning only on the complete part of samples) utilizes the sub‑multiset
\[
\mathfrak{U}_{\text{cpt}} = \biguplus_{i: |\mathcal{V}_i| = V} \bigl( \{i\} \times \mathcal{P}_i \bigr),
\]
with cardinality
\[
U_{\text{cpt}} = \sum_{i: |\mathcal{V}_i| = V} \binom{V}{2}.
\]

\begin{theorem}[Pair Utilization]\label{thm:utilization}
For any threshold \(\tau \in [2, V]\), the number of view-pairs utilized by TreeEIC is given by
\begin{equation}\label{eq:utilization}
U(\tau) = \sum_{\substack{i \in \mathcal{I} \\ |\mathcal{V}_i| \ge \tau}} \binom{|\mathcal{V}_i|}{2}.
\end{equation}
Then \(U(\tau)\) is non‑increasing on \([2,V]\), and we have the chain of inequalities
\[
U(2) \;\ge\; U(\tau) \;\ge\; U(V) \;=\; U_{\text{cpt}},
\]
where \(U_{\text{cpt}}\) denotes the number of view-pairs utilized by the conventional method. The gap between \(U(\tau)\) and \(U_{\text{cpt}}\) is precisely
\[
U(\tau) - U_{\text{cpt}} = \sum_{\substack{i \in \mathcal{I} \\ \tau \le |\mathcal{V}_i| < V}} \binom{|\mathcal{V}_i|}{2} \ge 0,
\]
with equality if and only if there is no sample \(i\) satisfying \(\tau \le |\mathcal{V}_i| < V\); i.e., the set
\[
\mathcal{I}_\tau := \{ i \in \mathcal{I} \mid \tau \le |\mathcal{V}_i| < V \}
\]
is empty. In particular, \(U(\tau) = U_{\text{cpt}}\) iff \(\mathcal{I}_\tau = \emptyset\).
\end{theorem}

\begin{proof}
We establish the theorem through a series of steps.

{Step 1 (Covering Property).}  
For any \(i \in \mathcal{I}\) and any \(\{u,v\} \subseteq \mathcal{V}_i\) with \(|\mathcal{V}_i| \ge \tau\), there exists a pattern \(M \in \mathfrak{M}_\tau\) such that
\[
\{u,v\} \subseteq M \subseteq \mathcal{V}_i. \tag{A}
\]
\textit{Proof of Step 1.} If \(\tau = 2\), take \(M = \{u,v\}\); clearly \(M \subseteq \mathcal{V}_i\) and \(|M| = \tau\). If \(\tau > 2\), since \(|\mathcal{V}_i| \ge \tau\), the set \(\mathcal{V}_i \setminus \{u,v\}\) contains at least \(\tau-2\) elements. Choose any \((\tau-2)\)-element subset \(R \subseteq \mathcal{V}_i \setminus \{u,v\}\) and set \(M = \{u,v\} \cup R\). Then \(|M| = \tau\) and \(M \subseteq \mathcal{V}_i\), hence \(M \in \mathfrak{M}_\tau\).

{Step 2 (Sufficiency).}  
If \(|\mathcal{V}_i| \ge \tau\), then for every \(\{u,v\} \subseteq \mathcal{V}_i\), we have
\[
(i,\{u,v\}) \in \mathfrak{U}_\tau. \tag{B}
\]
\textit{Proof of Step 2.} By Step 1, there exists \(M \in \mathfrak{M}_\tau\) with \(\{u,v\} \subseteq M \subseteq \mathcal{V}_i\). This directly implies \((i,\{u,v\}) \in \mathfrak{U}_\tau\) by definition.

{Step 3 (Necessity).}  
If \(|\mathcal{V}_i| < \tau\), then for any \(\{u,v\} \subseteq \mathcal{V}_i\), we have
\[
(i,\{u,v\}) \notin \mathfrak{U}_\tau. \tag{C}
\]
\textit{Proof of Step 3.} Assume for contradiction that \((i,\{u,v\}) \in \mathfrak{U}_\tau\). Then there exists \(M \in \mathfrak{M}_\tau\) with \(\{u,v\} \subseteq M \subseteq \mathcal{V}_i\), which yields \(|\mathcal{V}_i| \ge |M| = \tau\), contradicting \(|\mathcal{V}_i| < \tau\).

{Step 4 (Characterization of} \(\mathfrak{U}_\tau\)).  
Combining (B) and (C) we obtain the disjoint decomposition
\[
\mathfrak{U}_\tau = \biguplus_{\substack{i \in \mathcal{I} \\ |\mathcal{V}_i| \ge \tau}} \bigl( \{i\} \times \mathcal{P}_i \bigr). \tag{D}
\]
\textit{Proof of Step 4.} (B) shows that the right‑hand side is contained in \(\mathfrak{U}_\tau\). Conversely, if \((i,\{u,v\}) \in \mathfrak{U}_\tau\), then by (C) we must have \(|\mathcal{V}_i| \ge \tau\), so the element belongs to the right‑hand side. Disjointness follows from the distinctness of the sample indices.

{Step 5 (Expression for} \(U(\tau)\)).  
Taking cardinalities in (D) gives
\[
U(\tau) = \sum_{\substack{i \in \mathcal{I} \\ |\mathcal{V}_i| \ge \tau}} |\{i\} \times \mathcal{P}_i|
      = \sum_{\substack{i \in \mathcal{I} \\ |\mathcal{V}_i| \ge \tau}} |\mathcal{P}_i|
      = \sum_{\substack{i \in \mathcal{I} \\ |\mathcal{V}_i| \ge \tau}} \binom{|\mathcal{V}_i|}{2},
\]
which is exactly Eq.~\eqref{eq:utilization}. This establishes the closed‑form expression for \(U(\tau)\).

{Step 6 (Comparison with} \(U_{\text{cpt}}\)).  
Observe that \(\{ i \mid |\mathcal{V}_i| = V \} \subseteq \{ i \mid |\mathcal{V}_i| \ge \tau \}\), and for any complete sample we have \(\binom{|\mathcal{V}_i|}{2} = \binom{V}{2}\). Hence
\[
U(\tau) = \sum_{\substack{i: |\mathcal{V}_i| \ge \tau}} \binom{|\mathcal{V}_i|}{2}
      \ge \sum_{\substack{i: |\mathcal{V}_i| = V}} \binom{V}{2}
      = U_{\text{cpt}}.
\]
Splitting the left sum into complete and partially observed parts,
\[
U(\tau) = \sum_{\substack{i: |\mathcal{V}_i| = V}} \binom{V}{2}
      + \sum_{\substack{i: \tau \le |\mathcal{V}_i| < V}} \binom{|\mathcal{V}_i|}{2}
      = U_{\text{cpt}} + \Delta(\tau),
\]
where \(\Delta(\tau) := \sum_{i: \tau \le |\mathcal{V}_i| < V} \binom{|\mathcal{V}_i|}{2} \ge 0\). Clearly \(\Delta(\tau) > 0\) iff \(\mathcal{I}_\tau \neq \emptyset\). This yields the gap formula.

{Step 7 (Monotonicity).}  
Let \(\tau_1, \tau_2\) satisfy \(2 \le \tau_1 < \tau_2 \le V\). Applying the expression of \(U(\tau)\) to both thresholds,
\[
U(\tau_1) = \sum_{\substack{i: |\mathcal{V}_i| \ge \tau_1}} \binom{|\mathcal{V}_i|}{2},\qquad
U(\tau_2) = \sum_{\substack{i: |\mathcal{V}_i| \ge \tau_2}} \binom{|\mathcal{V}_i|}{2}.
\]
Since the condition \(|\mathcal{V}_i| \ge \tau_1\) is weaker than \(|\mathcal{V}_i| \ge \tau_2\), the first sum includes all terms of the second sum plus possibly additional terms corresponding to samples with \(\tau_1 \le |\mathcal{V}_i| < \tau_2\). Thus
\[
U(\tau_1) - U(\tau_2) = \sum_{\substack{i: \tau_1 \le |\mathcal{V}_i| < \tau_2}} \binom{|\mathcal{V}_i|}{2} \ge 0,
\]
so \(U(\tau_1) \ge U(\tau_2)\). Hence \(U\) is non‑increasing. The extremal cases \(U(2)\) and \(U(V)\) follow directly from the definition, and \(U(V) = U_{\text{cpt}}\) by construction.

All claims of Theorem~\ref{thm:utilization} are now proved.
\end{proof}

\noindent\textbf{Discussion.}
Theorem~\ref{thm:utilization} reveals that \(\tau\) acts as a selectivity threshold: only samples whose available view count meets or exceeds \(\tau\) contribute to the ensemble decision process. Lowering \(\tau\) incorporates more samples, thereby exploiting a larger number of view‑pairs, but each decision set then comprises exactly \(\tau\) views, potentially reducing the richness of cross‑view complementarity. Raising \(\tau\) has the converse effect. By adaptively tuning \(\tau\) via Eq.~(5), TreeEIC achieves a data‑driven trade‑off between coverage and view‑pair utility, effectively overcoming the under‑utilization problem that plagues conventional IMVC methods.

Furthermore, the monotonicity property guarantees that the utilization decreases as \(\tau\) increases, with the extreme cases corresponding to maximal exploitation (\(\tau=2\)) and the conventional baseline (\(\tau=V\)). Under high missing rates, a smaller \(\tau\) retains a higher proportion of view‑pairs, which is crucial for maintaining clustering performance when data are severely incomplete.

\subsection{Uncertainty-Weighted Ensemble}
\label{sec:theory}

In this part, we provide a statistical learning perspective on why the uncertainty-weighted ensemble strategy could reduce generalization error.
We first establish the notation, then derive an upper bound for the ensemble risk that is no larger than the average risk of the base learners.
Finally, we show that the ensemble can potentially outperform individual base learners.

\noindent\textbf{Notations.}
Let \(\mathcal{I}\) be the set of all samples. For each sample \(i \in \mathcal{I}\), there exists an unknown true label vector \(\mathbf{y}_i \in \{0,1\}^K\) (one-hot encoding of the cluster assignment). Through the missing-pattern tree, we obtain a collection of decision sets \(\{\mathcal{S}_j\}_{j=1}^{|\mathcal{C}|}\). For sample \(i\), denote by \(J(i) = \{ j : i \in \mathcal{S}_j \}\) the set of indices of decision sets that contain \(i\). For each \(j \in J(i)\), the corresponding decision set \(\mathcal{S}_j\) produces a soft cluster assignment \(\mathbf{d}_i^j \in \Delta^{K-1}\) (the probability simplex in \(\mathbb{R}^K\)). The entropy of this assignment is
\[
h_i^j = -\sum_{k=1}^{K} d_{i,k}^j \log d_{i,k}^j.
\]
The uncertainty weight for sample \(i\) in decision set \(j\) is defined as the inverse of its entropy ($h_i^j$), normalized across all available decision sets:
\[
w_i^j = \frac{1/h_i^j}{\sum_{j' \in J(i)} 1/h_i^{j'}}, \qquad \sum_{j \in J(i)} w_i^j = 1.
\]
The ensemble prediction for sample \(i\) is then the weighted combination
\[
\mathbf{p}_i = \sum_{j \in J(i)} w_i^j \mathbf{d}_i^j.
\]
We measure the discrepancy between a prediction \(\mathbf{q}\) and the true label \(\mathbf{y}\) by the squared loss
\[
L(\mathbf{y},\mathbf{q}) = \|\mathbf{y} - \mathbf{q}\|_2^2,
\]
which is convex in \(\mathbf{q}\). The expected risk of a \emph{base learner} \(j\) from one decision set on sample \(i\) is
\[
R_i^j = \mathbb{E}\big[ L(\mathbf{y}_i, \mathbf{d}_i^j) \big],
\]
and the ensemble risk is
\[
R_{\mathrm{ens}} = \mathbb{E}_i\big[ L(\mathbf{y}_i, \mathbf{p}_i) \big],
\]
where the outer expectation is taken over the random draw of a sample and the randomness in the clustering process.

We first recall a classic result for ensemble learning.
\begin{lemma}[Error-Ambiguity Decomposition]
\label{lem:error-ambiguity}
For a sample \(i\) and any set of non‑negative weights \(\{w_i^j\}_{j \in J(i)}\) summing to one,
\[
L(\mathbf{y}_i, \mathbf{p}_i) = \sum_{j \in J(i)} w_i^j L(\mathbf{y}_i, \mathbf{d}_i^j) - \sum_{j \in J(i)} w_i^j \big\|\mathbf{d}_i^j - \mathbf{p}_i\big\|_2^2.
\]
\end{lemma}
\begin{proof}
The proof follows by expanding the squared norm and rearranging terms; see e.g., \cite{krogh1995neural}.
\end{proof}

Since the ambiguity term \(\sum_j w_i^j \|\mathbf{d}_i^j - \mathbf{p}_i\|^2\) is always non‑negative, we immediately obtain an upper bound:
\begin{equation}\label{eq:bound1}
    L(\mathbf{y}_i, \mathbf{p}_i) \le \sum_{j \in J(i)} w_i^j L(\mathbf{y}_i, \mathbf{d}_i^j).
\end{equation}
The uncertainty weights \(w_i^j\) are constructed to be decreasing functions of the entropy \(h_i^j\). Intuitively, a higher entropy indicates a less confident prediction, which is more likely to incur a larger loss. To formalize this intuition, we introduce the following assumption.
Let \(\mathcal{F}_i\) be the \(\sigma\)-field generated by the predictions \(\{\mathbf{d}_i^j\}_{j \in J(i)}\). Note that both \(h_i^j\) and \(w_i^j\) are \(\mathcal{F}_i\)-measurable. Define the conditional expected loss
\[
r_i^j := \mathbb{E}\big[ L(\mathbf{y}_i, \mathbf{d}_i^j) \mid \mathcal{F}_i \big].
\]
\begin{assumption}[Entropy-Loss Positive Correlation]
\label{ass:corr}
For every sample \(i\) and any two decision sets \(j, j' \in J(i)\),
\[
h_i^j \ge h_i^{j'} \quad\Longrightarrow\quad r_i^j \ge r_i^{j'}.
\]
\end{assumption}
In words, conditionally on the observed predictions, a higher entropy implies a higher expected loss. This assumption is plausible in unsupervised clustering because ambiguous predictions are more prone to misassignment, leading to larger expected error. Because \(w_i^j\) is a decreasing function of \(h_i^j\), Assumption~\ref{ass:corr} implies that the sequences \(\{r_i^j\}_{j \in J(i)}\) and \(\{w_i^j\}_{j \in J(i)}\) are oppositely ordered (i.e., larger \(r_i^j\) corresponds to smaller \(w_i^j\)).

Now we show that the weighted average of the conditional expected losses is bounded by their arithmetic average.
\begin{lemma}[Weighted Average $\leq$ Arithmetic Average]
\label{lem:neg-correlation}
For any sample \(i\), under Assumption~\ref{ass:corr},
\[
\sum_{j \in J(i)} w_i^j r_i^j \le \frac{1}{|J(i)|} \sum_{j \in J(i)} r_i^j.
\]
\end{lemma}
\begin{proof}
Because \(\{r_i^j\}_{j \in J(i)}\) and \(\{w_i^j\}_{j \in J(i)}\) are oppositely ordered, we can apply Chebyshev's sum inequality \cite{gradshteyn2014table} for finite sequences:
\[
\frac{1}{|J(i)|} \sum_{j \in J(i)} w_i^j r_i^j \le
\left( \frac{1}{|J(i)|} \sum_{j \in J(i)} w_i^j \right)
\left( \frac{1}{|J(i)|} \sum_{j \in J(i)} r_i^j \right)
= \frac{1}{|J(i)|} \sum_{j \in J(i)} r_i^j,
\]
where we used \(\sum_j w_i^j=1\). Multiplying both sides by \(|J(i)|\) yields the desired inequality.
\end{proof}
Observe that \(\sum_j w_i^j r_i^j\) is exactly the conditional expectation of the weighted loss:
\[
\mathbb{E}\Big[ \sum_{j \in J(i)} w_i^j L(\mathbf{y}_i, \mathbf{d}_i^j) \;\Big|\; \mathcal{F}_i \Big]
= \sum_{j \in J(i)} w_i^j \mathbb{E}\big[ L(\mathbf{y}_i, \mathbf{d}_i^j) \mid \mathcal{F}_i \big]
= \sum_{j \in J(i)} w_i^j r_i^j.
\]
Therefore, Lemma~\ref{lem:neg-correlation} implies
\[
\mathbb{E}\Big[ \sum_{j \in J(i)} w_i^j L(\mathbf{y}_i, \mathbf{d}_i^j) \;\Big|\; \mathcal{F}_i \Big]
\le \frac{1}{|J(i)|} \sum_{j \in J(i)} r_i^j.
\]
Taking expectations on both sides and using the law of total expectation,
\[
\mathbb{E}\Big[ \sum_{j \in J(i)} w_i^j L(\mathbf{y}_i, \mathbf{d}_i^j) \Big]
\le \mathbb{E}\Big[ \frac{1}{|J(i)|} \sum_{j \in J(i)} r_i^j \Big]
= \frac{1}{|J(i)|} \sum_{j \in J(i)} \mathbb{E}[ r_i^j ]
= \frac{1}{|J(i)|} \sum_{j \in J(i)} R_i^j.
\]
Notice that \(|J(i)|\) is a deterministic constant (it depends only on the missing pattern of sample \(i\)), so the right‑hand side can also be written as
\[
\mathbb{E}\Big[ \frac{1}{|J(i)|} \sum_{j \in J(i)} L(\mathbf{y}_i, \mathbf{d}_i^j) \Big]
= \frac{1}{|J(i)|} \sum_{j \in J(i)} R_i^j.
\]
Thus we have established
\begin{equation}\label{eq:bound2}
\mathbb{E}\Big[ \sum_{j \in J(i)} w_i^j L(\mathbf{y}_i, \mathbf{d}_i^j) \Big]
\le \mathbb{E}\Big[ \frac{1}{|J(i)|} \sum_{j \in J(i)} L(\mathbf{y}_i, \mathbf{d}_i^j) \Big].
\end{equation}
Combining the upper bound from the error-ambiguity decomposition \eqref{eq:bound1} with inequality \eqref{eq:bound2}, we obtain the following theorem.
\begin{theorem}[Ensemble Risk Bound]
\label{thm:main}
Under Assumption~\ref{ass:corr}, the ensemble risk satisfies
\[
R_{\mathrm{ens}} \le \mathbb{E}_i\!\left[ \frac{1}{|J(i)|} \sum_{j \in J(i)} L(\mathbf{y}_i, \mathbf{d}_i^j) \right].
\]
In words, the ensemble risk is no larger than the average risk of the base learners, where the average is taken over all available decisions for each sample and then averaged over samples.
\end{theorem}
\begin{proof}
From \eqref{eq:bound1} and \eqref{eq:bound2},
\[
R_{\mathrm{ens}} = \mathbb{E}_i\big[ L(\mathbf{y}_i, \mathbf{p}_i) \big]
\le \mathbb{E}_i\!\left[ \sum_{j \in J(i)} w_i^j L(\mathbf{y}_i, \mathbf{d}_i^j) \right]
\le \mathbb{E}_i\!\left[ \frac{1}{|J(i)|} \sum_{j \in J(i)} L(\mathbf{y}_i, \mathbf{d}_i^j) \right].
\]
The second inequality is exactly \eqref{eq:bound2} with the sample index \(i\) made explicit. This completes the proof.
\end{proof}
The error-ambiguity decomposition (Lemma~\ref{lem:error-ambiguity}) is an equality, not merely an inequality. Therefore we can write
\[
R_{\mathrm{ens}} = \mathbb{E}_i\!\left[ \sum_{j \in J(i)} w_i^j L(\mathbf{y}_i, \mathbf{d}_i^j) \right]
- \mathbb{E}_i\!\left[ \sum_{j \in J(i)} w_i^j \|\mathbf{d}_i^j - \mathbf{p}_i\|^2 \right].
\]
The second term (the expected ambiguity) is non‑negative and is strictly positive whenever the base learners disagree on a sample. This observation leads to the following corollary.
\begin{theorem}[Ensemble Risk Inequality]
\label{cor:superiority}
If there exists at least one sample \(i\) for which the base learners are not all identical (i.e., \(\mathbf{d}_i^j \neq \mathbf{p}_i\) for some \(j\)), then
\[
R_{\mathrm{ens}} < \mathbb{E}_i\!\left[ \sum_{j \in J(i)} w_i^j L(\mathbf{y}_i, \mathbf{d}_i^j) \right] \le \mathbb{E}_i\!\left[ \frac{1}{|J(i)|} \sum_{j \in J(i)} L(\mathbf{y}_i, \mathbf{d}_i^j) \right].
\]
In particular, the ensemble risk is strictly smaller than the weighted average risk of the base learners, and consequently it is possible for the ensemble to achieve a risk lower than that of any individual base learner.
\end{theorem}
\begin{proof}
The first inequality follows from Lemma~\ref{lem:error-ambiguity} because the ambiguity term is positive when base learners disagree. The second inequality is from \eqref{eq:bound2}. Therefore, the ensemble risk is strictly smaller than the upper bound in Theorem~\ref{thm:main}, implying that it may fall below all individual risks.
\end{proof}

\noindent\textbf{Discussion.} In the proof, the error-ambiguity decomposition holds for any convex loss (here we consider the squared loss as an example).
The entropy-loss positive correlation assumption is natural in clustering, i.e., ambiguous predictions (high entropy) are more likely to be erroneous.
The analysis shows that the ensemble risk is upper bounded by the average risk of the base learners, and the ambiguity term further contributes to a potential reduction beyond that bound.
This theoretical result supports the uncertainty-weighted design of the TreeEIC framework and explains its effectiveness for multi‑view decision ensemble.
Moreover, the experimental results in Section~IV corroborate the theoretical finding, i.e., removing the uncertainty-weighted ensemble leads to a significant drop in clustering accuracy, confirming its crucial role in improving learning performance.

\newpage
\section*{Appendix II: More Details and Results} 

We evaluate the IMVC performance on six public multi-view datasets as listed in Table~\ref{tab:dataset_info2},
including HandWritten~\cite{lecun1989backpropagation}, Caltech101-7~\cite{fei2004learning}, OutdoorScene~\cite{hu2020multi}, AWA-7~\cite{romera2015embarrassingly}, ModelNet-40~\cite{wu20153d}, and MVP~\cite{pan2021variational}.

\begin{table*}[!ht]
\centering
\renewcommand\tabcolsep{7.0pt} 
\small
\caption{Statistics of multi-view datasets used in our experiments.}
\label{tab:dataset_info2}
\resizebox{\linewidth}{!}{
\begin{tabular}{lcccccc}
\toprule
\textbf{Dataset} &\textbf{Type}& \textbf{\#Views} & \textbf{\#Samples} & \textbf{\#Clusters} & \textbf{Dimensions} \\
\midrule
HandWritten~\cite{lecun1989backpropagation}    &handwritten numerals & 6 & 2,000 & 10 & 240/76/216/47/64/6 \\
Caltech101-7~\cite{fei2004learning}   &single object images & 5 & 1,400 & 7 & 1984/512/928/254/40 \\
OutdoorScene~\cite{hu2020multi}   &outdoor scene images & 4 & 2,688 & 8 & 512/432/256/48 \\
AWA-7~\cite{romera2015embarrassingly}         &animal images & 7 & 10,158  & 50 &   2688/2000/2000/2000/2000/4096/4096 \\
ModelNet40~\cite{wu20153d}      &3D point cloud features & 2 & 12,311 & 40 & 3072/1024 \\
MVP~\cite{pan2021variational}        &CAD model features & 8 & 1,600 & 8 & 256/256/256/256/256/256/256/256 \\
\bottomrule
\end{tabular}
}
\end{table*}

We provide more experimental results which can not be shown in the paper due to its space limitation.
The effectiveness is evaluated by clustering accuracy (ACC), normalized mutual information (NMI), and adjusted rand index (ARI).

\begin{table*}[!ht]
\caption{\textbf{Clustering Performance Comparison on Six Datasets.} We report the mean$\pm$std values of five runs. $\rho$ denotes the missing rate. ``N/A'' indicates that the method cannot run in the case of $\rho=1.0$. ``OOM'' denotes the method has failed to run due to out-of-memory.}
\centering
\renewcommand\tabcolsep{3.0pt} 
\small
\resizebox{\textwidth}{!}{
\begin{threeparttable}
\begin{tabular}{cl|ccc|ccc|ccc|ccc|ccc} 
\toprule
\multirow{2}{*}{Dataset} & \multirow{2}{*}{Method} & \multicolumn{3}{c|}{0.1} & \multicolumn{3}{c|}{0.3} & \multicolumn{3}{c|}{0.5} & \multicolumn{3}{c|}{0.7} & \multicolumn{3}{c}{1} \\
\cmidrule(lr){3-5} \cmidrule(lr){6-8} \cmidrule(lr){9-11} \cmidrule(lr){12-14} \cmidrule(lr){15-17}
 & & ACC & NMI & ARI & ACC & NMI & ARI & ACC & NMI & ARI & ACC & NMI & ARI & ACC & NMI & ARI \\
\midrule
\multirow{7}{*}{\rotatebox{90}{HandWritten}} 
 & APADC~\cite{xu2023adaptive}  
 & 79.13$\pm$1.02 & 80.41$\pm$1.03 & 70.41$\pm$2.04 
 & 78.81$\pm$1.42 & 78.05$\pm$1.30 & 68.48$\pm$2.83 
 & 79.42$\pm$2.21 & 78.10$\pm$0.91 & 70.29$\pm$2.54 
 & 70.94$\pm$4.28 & 69.06$\pm$3.32 & 58.60$\pm$4.75 
 & 37.12$\pm$2.51 & 33.65$\pm$1.39 & 14.24$\pm$0.75 \\

 & ProImp~\cite{li2023incomplete}  & 85.92$\pm$2.55 & 82.14$\pm$1.38 & 77.55$\pm$2.30 & 82.65$\pm$1.79 & 77.66$\pm$1.57 & 72.44$\pm$2.26 & 82.05$\pm$3.61 & 74.33$\pm$3.04 & 69.04$\pm$4.34 & 80.45$\pm$4.35 & 70.05$\pm$3.48 & 64.41$\pm$5.75 &N/A &N/A &N/A \\
 & ICMVC~\cite{chao2024incomplete} 
 & 85.36$\pm$0.64 & 83.80$\pm$0.81 & 78.29$\pm$1.02 
 & 83.73$\pm$1.44 & 81.69$\pm$1.41 & 75.65$\pm$2.00 
 & 82.07$\pm$1.05 & 78.29$\pm$1.66 & 72.62$\pm$1.96 
 & 73.82$\pm$2.71 & 70.69$\pm$2.42 & 61.99$\pm$3.54 
 & 21.38$\pm$1.49 & 10.15$\pm$0.85 & ~4.77$\pm$0.63 \\

 & DCG~\cite{zhang2025incomplete}  &79.27$\pm$4.94 &78.25$\pm$3.35 &70.87$\pm$5.40 &72.61$\pm$7.06 &74.35$\pm$3.06 &64.30$\pm$5.56 &81.42$\pm$7.34 &77.50$\pm$4.71 &71.62$\pm$7.71 &74.16$\pm$6.93 &69.46$\pm$4.14 &61.66$\pm$6.59 &22.32$\pm$1.99 &12.88$\pm$1.81 &~5.14$\pm$1.26 \\
& GHICMC~\cite{chao2025global} 
& 96.83$\pm$0.26 & 92.78$\pm$0.50 & 93.11$\pm$0.55 
& 96.14$\pm$0.34 & 91.38$\pm$0.49 & 91.67$\pm$0.67 
& 94.90$\pm$0.87 & 89.18$\pm$1.05 & 89.14$\pm$1.64 
& 93.79$\pm$0.36 & 86.97$\pm$0.55 & 86.83$\pm$0.69 
& 89.74$\pm$5.79 & 84.06$\pm$3.68 & 82.01$\pm$6.68 \\
& FreeCSL~\cite{dai2025imputation}
& 84.40$\pm$1.32 & 88.90$\pm$1.09 & 82.00$\pm$1.22 
& 85.00$\pm$2.19 & 87.69$\pm$1.23 & 81.42$\pm$1.91 
& 84.20$\pm$2.45 & 85.71$\pm$1.06 & 79.33$\pm$1.82 
& 83.14$\pm$0.31 & 84.76$\pm$1.61 & 78.38$\pm$1.48 
& 81.85$\pm$1.24 & 81.66$\pm$0.99 & 75.52$\pm$1.06 \\
& \textbf{TreeEIC [ours]} & \textbf{97.20$\pm$0.19} & \textbf{93.77$\pm$0.20} & \textbf{93.87$\pm$0.41} & \textbf{96.69$\pm$0.24} & \textbf{92.75$\pm$0.41} & \textbf{92.79$\pm$0.50} & \textbf{95.86$\pm$0.37} & \textbf{91.14$\pm$0.88} & \textbf{91.05$\pm$0.81} & \textbf{94.41$\pm$0.67} & \textbf{88.54$\pm$1.03} & \textbf{88.12$\pm$1.30} & \textbf{93.77$\pm$1.04} & \textbf{87.09$\pm$1.87} & \textbf{86.76$\pm$2.13} \\
\hline
\multirow{7}{*}{\rotatebox{90}{Caltech101-7}} 
 & APADC~\cite{xu2023adaptive} 
& 55.79$\pm$0.65 & 42.65$\pm$0.44 & 37.04$\pm$0.40 
& 44.14$\pm$4.59 & 36.13$\pm$8.45 & 21.08$\pm$1.83 
& 47.01$\pm$5.19 & 47.30$\pm$5.18 & 17.38$\pm$6.55 
& 48.81$\pm$4.71 & 46.55$\pm$4.43 & 18.43$\pm$4.13 
& 38.63$\pm$0.40 & 29.77$\pm$1.04 & 16.39$\pm$1.44 \\

 & ProImp~\cite{li2023incomplete}
 & 81.71$\pm$0.70 & 71.56$\pm$2.22 & 66.63$\pm$1.91 & 75.72$\pm$4.61 & 64.23$\pm$5.05 & 57.95$\pm$6.71 & 71.87$\pm$7.00 & 59.89$\pm$6.20 & 52.99$\pm$8.42 & 70.23$\pm$5.55 & 58.47$\pm$4.80 & 51.35$\pm$6.34 &N/A & N/A & N/A \\
 & ICMVC~\cite{chao2024incomplete} 
& 80.94$\pm$4.72 & 74.07$\pm$3.23 & 68.44$\pm$4.99 
& 81.24$\pm$6.91 & 73.46$\pm$5.21 & 68.01$\pm$7.86 
& 80.70$\pm$3.97 & 72.16$\pm$3.07 & 66.96$\pm$4.31 
& 65.41$\pm$4.65 & 59.07$\pm$4.76 & 49.05$\pm$5.31 
& 27.24$\pm$0.60 & ~9.65$\pm$1.13 & ~5.79$\pm$0.67 \\

& DCG~\cite{zhang2025incomplete} 
& 79.84$\pm$3.77 & 72.22$\pm$4.02 & 67.37$\pm$4.81 & 79.46$\pm$3.80 & 71.65$\pm$3.57 & 67.14$\pm$4.38 & 73.42$\pm$5.57 & 64.12$\pm$4.62 & 57.71$\pm$5.62 & 78.14$\pm$5.37 & 67.10$\pm$4.90 & 62.70$\pm$6.80 & 30.13$\pm$4.48 & 13.11$\pm$4.04 & ~7.46$\pm$3.47 \\
& GHICMC~\cite{chao2025global} 
&81.21$\pm$1.42 &75.69$\pm$3.09 &70.92$\pm$3.58 &76.00$\pm$4.45 &68.80$\pm$3.20 &62.34$\pm$5.13 &78.34$\pm$1.52 &68.81$\pm$2.16 &63.49$\pm$2.55 &75.02$\pm$9.57 &66.33$\pm$6.55 &60.65$\pm$9.29 &77.15$\pm$3.71 &64.97$\pm$3.05 & 61.44$\pm$3.67\\
 & FreeCSL~\cite{dai2025imputation} 
 & 89.70$\pm$3.52 & 83.94$\pm$2.74 & 80.70$\pm$4.15 & 89.72$\pm$2.57 & 82.00$\pm$3.27 & 79.35$\pm$4.73 & 86.64$\pm$2.52 & 78.76$\pm$2.39 & 75.29$\pm$3.61 
 & 82.87$\pm$4.01 & 74.19$\pm$2.41 & 70.63$\pm$3.43 & 81.46$\pm$1.21 & 69.16$\pm$2.36 & 65.75$\pm$2.51 \\
 & \textbf{TreeEIC [ours]} & \textbf{92.31$\pm$1.15} & \textbf{85.90$\pm$1.47} & \textbf{83.56$\pm$2.18} & \textbf{90.41$\pm$2.12} & \textbf{82.57$\pm$2.72} & \textbf{80.13$\pm$3.54} & \textbf{90.61$\pm$1.86} & \textbf{82.96$\pm$2.16} & \textbf{80.49$\pm$3.41} & \textbf{89.99$\pm$1.57} & \textbf{81.94$\pm$2.34} & \textbf{79.57$\pm$2.79} & \textbf{88.27$\pm$1.18} & \textbf{78.70$\pm$1.58} & \textbf{76.62$\pm$1.99} \\

\hline

\multirow{7}{*}{\rotatebox{90}{OutdoorScene}} 
 & APADC~\cite{xu2023adaptive}  
 & 59.29$\pm$0.61 & 54.89$\pm$0.85 & 40.86$\pm$1.12 
 & 60.64$\pm$1.10 & 50.26$\pm$2.02 & 38.71$\pm$1.43 
 & 62.17$\pm$3.05 & 52.57$\pm$1.31 & 41.06$\pm$2.55 
 & 56.74$\pm$3.49 & 45.74$\pm$1.89 & 34.54$\pm$1.86 
 & 34.39$\pm$0.63 & 26.61$\pm$0.77 & 13.85$\pm$1.12 \\

& ProImp~\cite{li2023incomplete}  & 63.15$\pm$1.71 & 50.35$\pm$1.63 & 40.84$\pm$1.73 & 59.67$\pm$1.47 & 47.42$\pm$0.64 & 37.30$\pm$0.94 & 57.02$\pm$2.08 & 45.51$\pm$0.58 & 34.78$\pm$1.24 & 53.21$\pm$3.05 & 42.35$\pm$1.22 & 30.89$\pm$1.56 & N/A &N/A &N/A\\
 & ICMVC~\cite{chao2024incomplete} 
 & 70.25$\pm$1.88 & 57.52$\pm$0.98 & 49.49$\pm$1.62 
 & 67.85$\pm$1.39 & 56.06$\pm$0.74 & 47.54$\pm$0.80 
 & 65.31$\pm$4.32 & 52.11$\pm$3.29 & 43.45$\pm$4.68 
 & 61.29$\pm$3.61 & 48.58$\pm$2.92 & 38.82$\pm$4.03 
 & 20.10$\pm$0.51 & ~4.39$\pm$0.12 & ~2.34$\pm$0.04 \\

 & DCG~\cite{zhang2025incomplete}  & 65.08$\pm$3.29 & 54.61$\pm$1.95 & 44.98$\pm$3.26 & 64.48$\pm$6.36 & 53.88$\pm$4.89 & 44.53$\pm$5.38 & 61.39$\pm$5.36 & 50.87$\pm$3.27 & 41.77$\pm$3.93 & 60.61$\pm$5.24 & 49.55$\pm$2.92 & 40.19$\pm$4.25 & 24.87$\pm$1.86 & 10.60$\pm$3.14 & ~4.69$\pm$1.84 \\
& GHICMC~\cite{chao2025global}
& 72.38$\pm$3.01 & 59.93$\pm$1.37 & 52.09$\pm$2.31 & 71.32$\pm$2.82 & 57.40$\pm$1.41 & 50.03$\pm$2.48 & 69.03$\pm$1.11 & 55.42$\pm$0.81 & 47.57$\pm$1.13 & 66.26$\pm$3.77 & 52.28$\pm$1.77 & 43.75$\pm$3.55 & 62.16$\pm$4.46 & 47.54$\pm$2.71 & 38.94$\pm$3.62 \\
 &FreeCSL~\cite{dai2025imputation} 
 & 67.63$\pm$3.80 & 59.74$\pm$1.92 & 49.09$\pm$3.19 & 66.90$\pm$5.12 & 58.25$\pm$2.27 & 47.98$\pm$3.93 & 63.35$\pm$3.92 & 55.76$\pm$1.31 & 44.57$\pm$2.28 & 59.43$\pm$2.68 & 53.06$\pm$1.18 & 40.95$\pm$1.78 & 58.40$\pm$2.63 & 49.01$\pm$1.15 & 37.13$\pm$1.52 \\
& \textbf{TreeEIC [ours]} & \textbf{73.14$\pm$3.30} & \textbf{60.92$\pm$2.26} & \textbf{52.58$\pm$3.10} & \textbf{72.49$\pm$4.80} & \textbf{61.14$\pm$1.69} & \textbf{52.94$\pm$2.93} & \textbf{72.85$\pm$1.78} & \textbf{59.79$\pm$1.00} & \textbf{51.66$\pm$1.86} & \textbf{71.12$\pm$1.55} & \textbf{57.41$\pm$1.43} & \textbf{49.39$\pm$1.42} & \textbf{68.27$\pm$1.23} & \textbf{54.63$\pm$1.18} & \textbf{46.17$\pm$1.18} \\

\hline
\multirow{7}{*}{\rotatebox{90}{AWA-7}} 
 & APADC~\cite{xu2023adaptive} 
 & 27.12$\pm$1.01 & 43.09$\pm$0.91 & 14.13$\pm$0.64 
 & 27.29$\pm$0.61 & 42.88$\pm$0.43 & 14.30$\pm$0.29 
 & 28.26$\pm$1.33 & 42.00$\pm$0.89 & 14.30$\pm$0.55 
 & 29.24$\pm$0.96 & 42.19$\pm$0.93 & 15.64$\pm$1.22 
 & 30.13$\pm$1.29 & 44.31$\pm$1.07 & 18.17$\pm$0.91 \\

 & ProImp~\cite{li2023incomplete}  & 40.84$\pm$1.11 & 54.24$\pm$0.63 & 30.44$\pm$0.94 & 40.25$\pm$1.30 & 52.24$\pm$0.59 & 28.77$\pm$0.84 & 36.47$\pm$2.47 & 49.64$\pm$1.02 & 26.65$\pm$1.00 & 37.59$\pm$1.36 & 48.66$\pm$0.56 & 25.87$\pm$1.05 &N/A &N/A &N/A \\
 & ICMVC~\cite{chao2024incomplete} 
 & 48.26$\pm$0.94 & 63.34$\pm$0.35 & 38.06$\pm$0.86 
 & 48.38$\pm$1.61 & 63.31$\pm$0.43 & 38.87$\pm$1.44 & 48.59$\pm$1.74 & \textbf{60.44$\pm$0.45} & 37.63$\pm$1.12 
 & 44.33$\pm$0.90 & \textbf{57.15$\pm$0.34} & 33.70$\pm$0.60 
& ~7.05$\pm$0.17 & 10.59$\pm$0.61 & ~1.32$\pm$0.12 \\ 
 & DCG~\cite{zhang2025incomplete}  & 32.76$\pm$1.83 & 47.34$\pm$1.09 & 23.76$\pm$1.76 & 30.77$\pm$2.05 & 45.21$\pm$2.35 & 21.01$\pm$1.57 & 30.29$\pm$1.48 & 43.79$\pm$1.69 & 20.70$\pm$1.23 & 28.01$\pm$2.16 & 41.76$\pm$2.05 & 18.92$\pm$1.85 & ~9.10$\pm$2.77 & 13.83$\pm$1.63 & ~1.94$\pm$0.84 \\
& GHICMC~\cite{chao2025global}
& OOM&OOM &OOM &OOM&OOM &OOM &OOM &OOM  &OOM &OOM &OOM &OOM &OOM &OOM &OOM \\
 & FreeCSL~\cite{dai2025imputation}
 & 10.67$\pm$0.24 & 17.30$\pm$0.26 & ~3.54$\pm$0.15 & 10.86$\pm$0.52 & 17.56$\pm$0.26 & ~3.78$\pm$0.17 & 11.18$\pm$0.52 & 17.33$\pm$0.43 & ~3.8$\pm$0.35 & 11.09$\pm$0.59 & 17.16$\pm$0.56 & ~3.72$\pm$0.33 & 10.47$\pm$0.32 & 16.42$\pm$0.36 & ~3.37$\pm$0.15 \\
& \textbf{TreeEIC [ours]} & \textbf{63.53$\pm$0.44} & \textbf{69.39$\pm$0.34} & \textbf{52.31$\pm$0.43} & \textbf{58.32$\pm$0.64} & \textbf{63.69$\pm$0.64} & \textbf{45.87$\pm$0.32} & \textbf{54.63$\pm$0.79} & 58.48$\pm$0.42 & \textbf{40.41$\pm$0.72} & \textbf{52.04$\pm$1.59} & 54.34$\pm$0.70 & \textbf{37.10$\pm$1.09} & \textbf{46.45$\pm$0.21} & \textbf{46.67$\pm$0.49} & \textbf{29.37$\pm$0.35} \\
\hline

 \multirow{7}{*}{\rotatebox{90}{ModelNet40}} 
 & APADC~\cite{xu2023adaptive}  
 & 39.28$\pm$3.00 & 52.29$\pm$1.15 & 31.34$\pm$2.43 
 & 39.61$\pm$1.10 & 51.79$\pm$0.74 & 31.25$\pm$2.03 
 & 39.27$\pm$0.92 & 51.34$\pm$0.42 & 30.21$\pm$1.28 
 & 33.20$\pm$0.45 & 45.09$\pm$0.76 & 23.17$\pm$1.90 
 & 26.98$\pm$0.65 & 38.32$\pm$0.21 & 18.93$\pm$0.38 \\

  &ProImp~\cite{li2023incomplete}  &42.20$\pm$0.48 & 58.64$\pm$0.65 & 36.88$\pm$0.93 & 40.26$\pm$0.96 & 55.77$\pm$0.58 & 35.45$\pm$0.99 & 34.32$\pm$2.35 & 51.11$\pm$1.94 & 29.46$\pm$2.86 & 33.83$\pm$0.56 & 51.67$\pm$0.63 & 29.73$\pm$0.95 &N/A  &N/A  &N/A \\
 & ICMVC~\cite{chao2024incomplete} 
 & 47.26$\pm$1.27 & \textbf{61.55$\pm$0.57} & 40.90$\pm$2.45 
 & 45.81$\pm$1.58 & \textbf{60.73$\pm$0.76} & 39.07$\pm$1.40 
 & 41.42$\pm$1.07 & \textbf{58.19$\pm$0.57} & 34.40$\pm$1.14 
 & 30.63$\pm$1.07 & 50.39$\pm$1.17 & 26.39$\pm$1.88 
 & ~7.42$\pm$0.08 & ~9.26$\pm$0.22 & ~1.33$\pm$0.05 \\

 & DCG~\cite{zhang2025incomplete}  &42.01$\pm$1.02 &53.69$\pm$0.83 &34.55$\pm$0.73 &42.90$\pm$1.02 &53.89$\pm$0.75 &35.02$\pm$0.95 &41.74$\pm$1.41 &51.74$\pm$1.38 &33.19$\pm$0.95 &40.38$\pm$2.27 &50.32$\pm$0.80 &31.82$\pm$1.32 &14.05$\pm$0.98 &13.12$\pm$0.80 &~3.29$\pm$0.81 \\
 & GHICMC~\cite{chao2025global}
 & OOM&OOM &OOM &OOM&OOM &OOM &OOM &OOM  &OOM &OOM &OOM &OOM &OOM &OOM &OOM \\
 & FreeCSL~\cite{dai2025imputation} 
 & 43.44$\pm$1.11 & 58.21$\pm$0.47 & 36.41$\pm$1.02 & 42.93$\pm$0.41 & 57.56$\pm$0.20 & 36.00$\pm$0.78 & 42.45$\pm$0.83 & 56.43$\pm$0.94 & 35.86$\pm$1.65 & 41.09$\pm$1.23 & \textbf{55.18$\pm$0.70} & \textbf{33.65$\pm$1.45} & 25.90$\pm$0.50 & 37.73$\pm$0.66 & 17.89$\pm$1.04 \\
& \textbf{TreeEIC [ours]} 
& \textbf{48.05$\pm$1.97} & 59.23$\pm$1.17 & \textbf{42.81$\pm$2.34}
& \textbf{48.25$\pm$0.99} & 57.26$\pm$0.66 & \textbf{42.90$\pm$1.78}
& \textbf{45.37$\pm$1.71} & 55.28$\pm$1.00 & \textbf{37.74$\pm$2.46}
& \textbf{42.60$\pm$2.22} & 51.31$\pm$0.93 & 32.29$\pm$3.39
& \textbf{39.46$\pm$1.40} & \textbf{53.79$\pm$0.57} & \textbf{32.97$\pm$2.29} \\

\hline
\multirow{7}{*}{\rotatebox{90}{MVP}} 
 & APADC~\cite{xu2023adaptive}  
 & 83.10$\pm$0.48 & 72.47$\pm$0.74 & 65.83$\pm$0.89 
 & 84.20$\pm$2.71 & 74.97$\pm$3.74 & 68.58$\pm$5.40 
 & 86.30$\pm$1.86 & 76.78$\pm$2.61 & 72.12$\pm$3.53 
 & 82.38$\pm$3.51 & 72.72$\pm$1.64 & 67.20$\pm$3.08 
 & 49.51$\pm$3.50 & 44.63$\pm$5.33 & 29.87$\pm$5.84 \\

 & ProImp~\cite{li2023incomplete}  &83.71$\pm$3.39 &74.32$\pm$3.02 &67.91$\pm$5.75 &83.20$\pm$2.28 &72.88$\pm$2.21 &67.53$\pm$3.31 &74.64$\pm$7.36 &68.32$\pm$3.94 &59.28$\pm$6.42 &77.37$\pm$6.25 &67.26$\pm$4.02 &59.64$\pm$6.82 &N/A &N/A &N/A \\
 & ICMVC~\cite{chao2024incomplete} 
 & 88.66$\pm$3.94 & 81.35$\pm$2.81 & 78.57$\pm$4.62 
 & 88.17$\pm$3.38 & 80.89$\pm$2.07 & 77.63$\pm$3.67 
 & 80.26$\pm$4.58 & 75.38$\pm$3.27 & 69.43$\pm$5.13 
 & 85.66$\pm$3.77 & 77.42$\pm$3.07 & 73.00$\pm$5.24 
 & 22.78$\pm$0.66 & 9.47$\pm$0.99 & 4.59$\pm$0.55 \\

 & DCG~\cite{zhang2025incomplete}  & 84.25$\pm$5.38 & 76.22$\pm$4.81 & 71.29$\pm$6.92 & 83.41$\pm$4.91 & 76.83$\pm$3.71 & 71.31$\pm$5.96 & 84.34$\pm$5.37 & 76.45$\pm$4.71 & 71.26$\pm$7.31 & 82.26$\pm$5.41 & 73.50$\pm$2.98 & 68.78$\pm$4.94 & 33.75$\pm$5.61 & 22.16$\pm$3.32 & 11.34$\pm$5.00 \\
 & GHICMC~\cite{chao2025global}
 &86.09$\pm$4.31 &79.07$\pm$2.96 &74.60$\pm$5.67 &84.31$\pm$6.49 &79.78$\pm$3.68 &74.24$\pm$7.02 &85.62$\pm$3.74 &79.14$\pm$2.03 &74.24$\pm$3.57 &89.27$\pm$2.82 &81.17$\pm$2.32 &78.70$\pm$3.89 &88.04$\pm$5.56 &81.95$\pm$1.69 & 79.19$\pm$3.67 \\
& FreeCSL~\cite{dai2025imputation}
& 90.86$\pm$0.60 & \textbf{83.10$\pm$0.76} & 80.76$\pm$1.10 & 87.36$\pm$4.79 & 80.75$\pm$2.80 & 76.83$\pm$4.94 & 89.20$\pm$1.83 & 81.31$\pm$1.99 & 78.21$\pm$2.55 & 89.82$\pm$2.02 & 81.85$\pm$1.76 & 79.35$\pm$2.77 & 89.03$\pm$1.44 & 80.33$\pm$1.43 & 77.62$\pm$2.03 \\
& \textbf{TreeEIC [ours]} & \textbf{91.31$\pm$0.50} & 82.87$\pm$0.65 & \textbf{81.55$\pm$1.00} & \textbf{91.18$\pm$0.16} & \textbf{82.46$\pm$0.17} & \textbf{81.26$\pm$0.27} & \textbf{90.80$\pm$0.70} & \textbf{81.77$\pm$1.33} & \textbf{80.54$\pm$1.38} & \textbf{90.80$\pm$0.91} & \textbf{82.00$\pm$1.30} & \textbf{80.53$\pm$1.82} & \textbf{91.07$\pm$0.59} & \textbf{82.13$\pm$1.02} & \textbf{81.10$\pm$1.22} \\

\bottomrule
\end{tabular}
\end{threeparttable}
}
\end{table*}

\begin{table*}[!ht]
\caption{\textbf{Ablation Study on Model Variants.} MPT denotes the missing pattern tree for multi-view decision and MDE is the uncertainty-based weighing for multi-view decision ensemble.}
\small
\centering
\resizebox{\linewidth}{!}{
\begin{threeparttable}
    \begin{tabular}{c|ccc|ccc|ccc|ccc}
    \toprule
      \multicolumn{1}{c|}{} &\multicolumn{3}{c|}{HandWritten} &\multicolumn{3}{c|}{Caltech101-7} &\multicolumn{3}{c|}{OutdoorScene} 
      &\multicolumn{3}{c}{AWA-7} \cr
    \hline
   Variant & ACC & NMI & ARI & ACC & NMI & ARI & ACC & NMI & ARI& ACC & NMI & ARI \\
    \hline
    without MPT     
    &83.24$\pm$3.32 &73.63$\pm$3.05 &68.44$\pm$4.74
    &84.91$\pm$4.85 &74.31$\pm$5.50 &71.70$\pm$6.56
    &63.88$\pm$4.24 &50.05$\pm$2.07 &41.68$\pm$2.85
    &10.06$\pm$0.73 &12.89$\pm$0.88 &~2.59$\pm$0.44
    \\
    without MDE 
    &92.07$\pm$0.39 &84.57$\pm$0.48 &83.61$\pm$0.39
    &87.34$\pm$2.78 &77.14$\pm$3.65 &74.78$\pm$4.64
    &67.68$\pm$3.72 &54.07$\pm$2.41 &45.55$\pm$2.77
    &40.58$\pm$1.52 &42.64$\pm$0.87 &24.73$\pm$0.90
    \\
    MPT+MDE          
    & \textbf{93.77$\pm$1.04} & \textbf{87.09$\pm$1.87} & \textbf{86.76$\pm$2.13}
    &\textbf{88.27$\pm$1.18} &\textbf{78.70$\pm$1.58} &\textbf{76.62$\pm$1.99}
    & \textbf{68.27$\pm$1.23} & \textbf{54.63$\pm$1.18} & \textbf{46.17$\pm$1.18}
    & \textbf{46.45$\pm$0.21} & \textbf{46.67$\pm$0.49} & \textbf{29.37$\pm$0.35}
    \\
    \bottomrule
    \end{tabular}
\end{threeparttable}
}
\end{table*}
\begin{figure*}[t]
    \centering

    \begin{subfigure}[b]{0.32\textwidth}
        \centering
        \includegraphics[width=\linewidth]{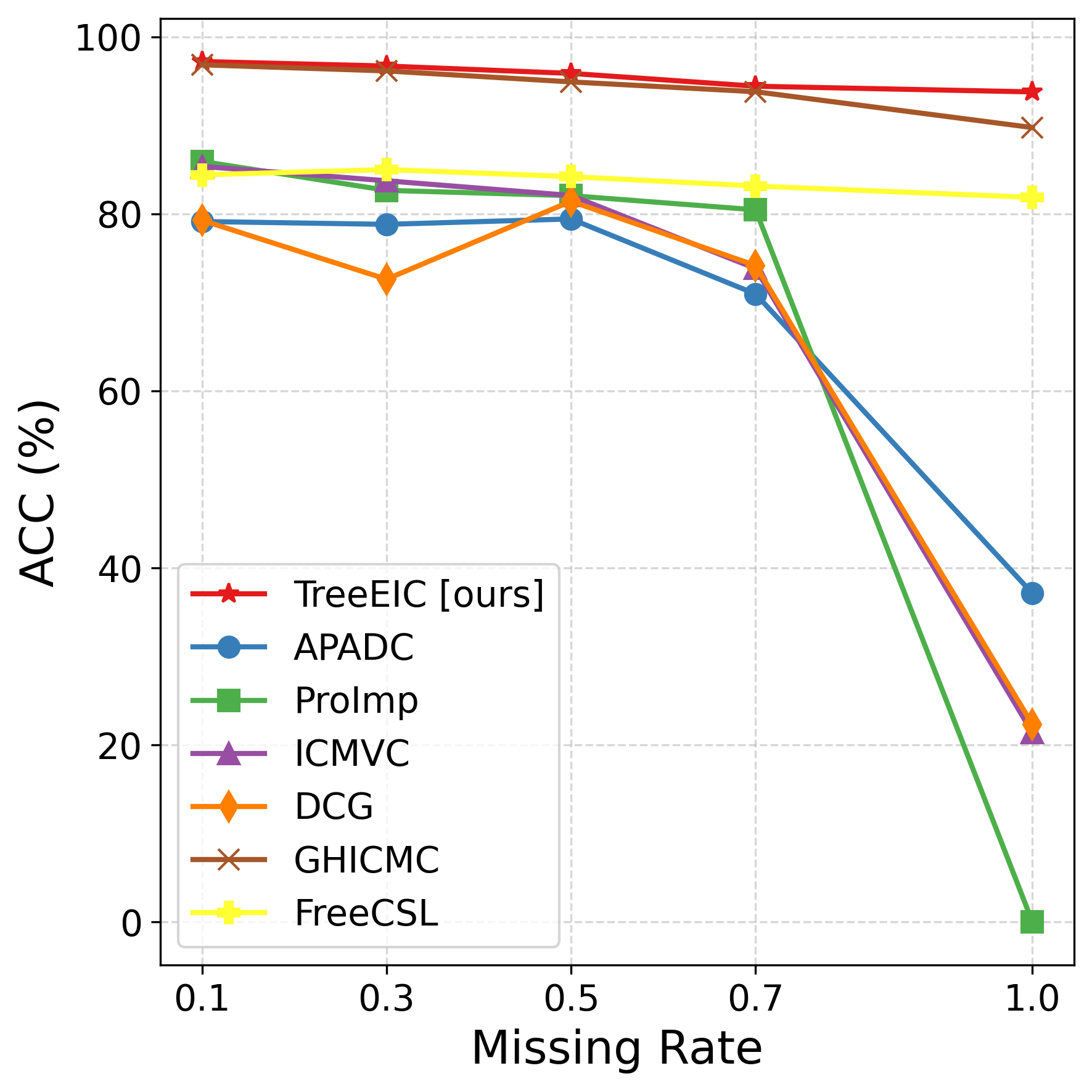}
        \caption{HandWritten}
        \label{fig:handwritten_acc}
    \end{subfigure}
    \hfill
    \begin{subfigure}[b]{0.32\textwidth}
        \centering
        \includegraphics[width=\linewidth]{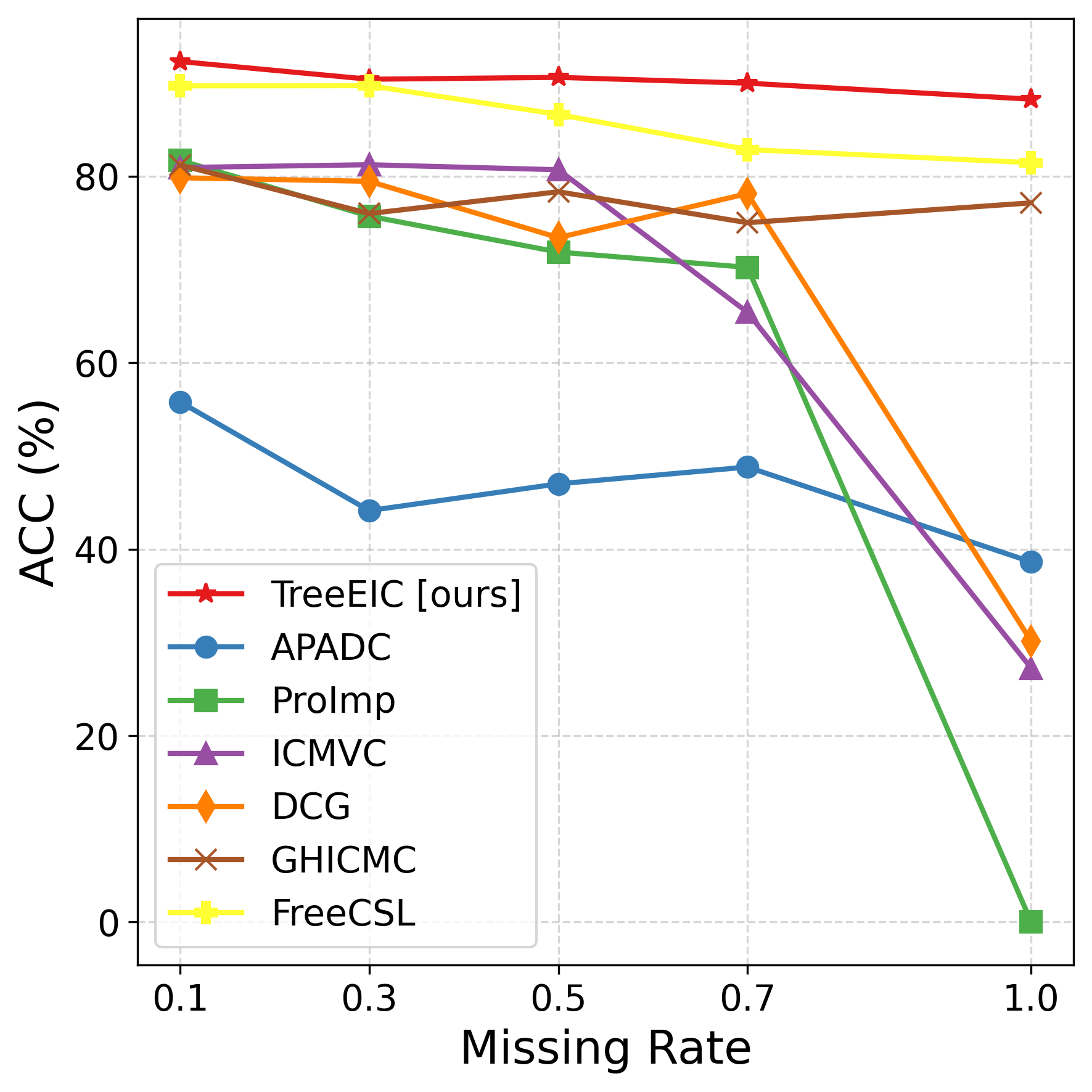}
        \caption{Caltech101-7}
        
    \end{subfigure}
    \hfill
    \begin{subfigure}[b]{0.32\textwidth}
        \centering
        \includegraphics[width=\linewidth]{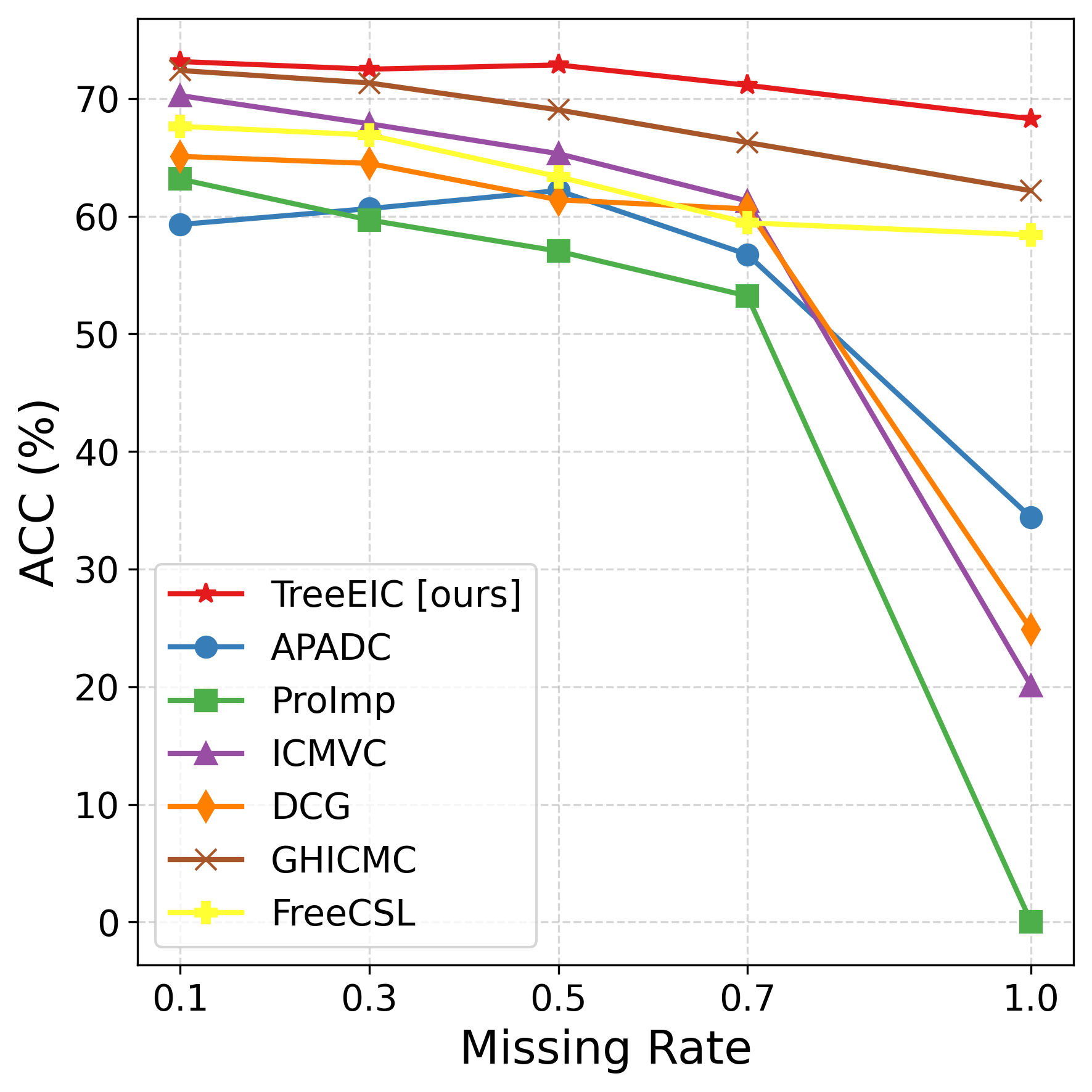}
        \caption{OutdoorScene}
        
    \end{subfigure}

    \vspace{0.3cm} 

    \begin{subfigure}[b]{0.32\textwidth}
        \centering
        \includegraphics[width=\linewidth]{figures/Dataset4_ACCTrend.png}
        \caption{AWA-7}
    \end{subfigure}
    \hfill
    \begin{subfigure}[b]{0.32\textwidth}
        \centering
        \includegraphics[width=\linewidth]{figures/Dataset5_ACCTrend.png}
        \caption{ModelNet40}
        
    \end{subfigure}
    \hfill
    \begin{subfigure}[b]{0.32\textwidth}
        \centering
        \includegraphics[width=\linewidth]{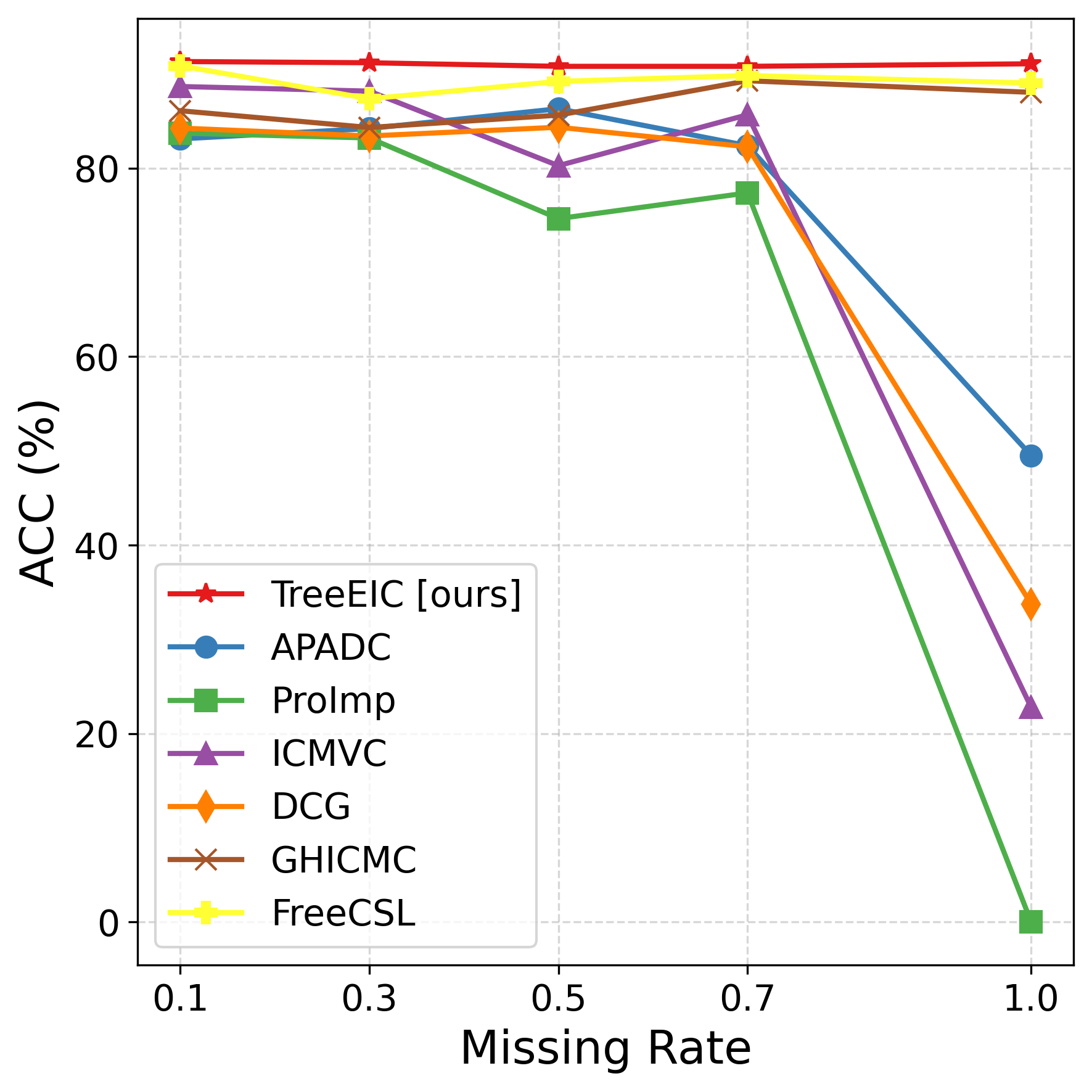}
        \caption{MVP}
        \label{fig:MVP8V_acctrend}
    \end{subfigure}

    \caption{ACC $vs.$ Missing Rate across six datasets. When the missing rate $\tau =1.0$, i.e., the highly inconsistent missing patterns, we can observe that most IMVC methods have heavy performance degradation while our method TreeEIC is still robust.}
\end{figure*}

\begin{table*}[!ht]
\caption{\textbf{Ablation Study on Loss Components.} $\mathcal{L}_{cons}$ is cross-view consistency loss and $\mathcal{L}_{disc}$ is inter-cluster discrimination loss.}
\small
\centering
\resizebox{\linewidth}{!}{
\begin{threeparttable}
    \begin{tabular}{cc|ccc|ccc|ccc|ccc}
    \toprule
    \multicolumn{2}{c|}{Components} &\multicolumn{3}{c|}{HandWritten} &\multicolumn{3}{c|}{Caltech101-7} &\multicolumn{3}{c|}{OutdoorScene} &\multicolumn{3}{c}{AWA-7} \cr
    \hline
    $\mathcal{L}_{cons}$ &$\mathcal{L}_{disc}$  & ACC & NMI & ARI & ACC & NMI & ARI& ACC & NMI & ARI& ACC & NMI & ARI \\
    \hline
    \ding{55} &\ding{55}  
    &35.37$\pm$3.48 &28.24$\pm$2.96 &15.98$\pm$2.41
    &37.11$\pm$3.09 &19.16$\pm$1.18 &12.93$\pm$1.89
    &26.38$\pm$1.08 &12.06$\pm$2.61 &~6.76$\pm$1.89
    &11.10$\pm$0.38 &10.86$\pm$0.28 &~1.96$\pm$0.16
    \\
    \ding{51} &\ding{55}  
    &91.04$\pm$0.73 &83.16$\pm$0.83 &81.73$\pm$1.04
    &52.19$\pm$2.96 &35.33$\pm$3.25 &26.87$\pm$2.98
    &51.34$\pm$3.78 &35.83$\pm$2.69 &27.12$\pm$2.87
    &42.95$\pm$0.49 &42.29$\pm$0.41 &26.04$\pm$0.61
    
    \\
    \ding{55} &\ding{51}          
    &92.49$\pm$2.77 &86.69$\pm$1.93 &85.36$\pm$3.34
    &87.53$\pm$2.63 &77.43$\pm$3.49 &75.16$\pm$4.47
    &67.99$\pm$3.66 &54.48$\pm$2.22 &46.15$\pm$2.46
    &41.25$\pm$1.26 &43.26$\pm$0.77 &25.05$\pm$0.69
 \\
    \ding{51} &\ding{51}          
    & \textbf{93.77$\pm$1.04} & \textbf{87.09$\pm$1.87} & \textbf{86.76$\pm$2.13}
    &\textbf{88.27$\pm$1.18} &\textbf{78.70$\pm$1.58} &\textbf{76.62$\pm$1.99}
    & \textbf{68.27$\pm$1.23} & \textbf{54.63$\pm$1.18} & \textbf{46.17$\pm$1.18}
    & \textbf{46.45$\pm$0.21} & \textbf{46.67$\pm$0.49} & \textbf{29.37$\pm$0.35}
    \\
    
    \bottomrule
    \end{tabular}
\end{threeparttable}
}
\end{table*}

\begin{table*}[!ht]\caption{\textbf{Analysis on Missing-Pattern Tree.} Subscript $_{\mathcal{U}}$ denotes the clustering results of the sample union set $\mathcal{U}$, which will impact the performance of all $N$ samples marked with $_{N}$.}
\tiny
\centering
\resizebox{\linewidth}{!}{
\begin{threeparttable}
    \begin{tabular}{l|c|c|c|c|c|c}
    \toprule
    \multicolumn{1}{c|}{$\tau$}&$|\mathcal{C}|$&$|\mathcal{S}_j|$ &$|\mathcal{U}|$ & $\text{ACC}_{\mathcal{U}} \rightarrow \text{ACC}_N$ & $\text{NMI}_{\mathcal{U}} \rightarrow \text{NMI}_N$ & $\text{ARI}_{\mathcal{U}} \rightarrow \text{ARI}_N$ \cr
    
    \hline
    1      
    &6 &1189  &2000 
    &81.40 $\rightarrow$ 82.45  &70.94 $\rightarrow$ 72.98 &65.51 $\rightarrow$ 67.56
    \\
    2     
    &15 &687 &1957
    &86.20 $\rightarrow$ 89.40 &76.89 $\rightarrow$ 81.36 &73.53 $\rightarrow$ 78.82
    \\
    3     
    &20 &382 &1658
    &87.82 $\rightarrow$ 91.85 &78.93 $\rightarrow$ 83,79 &76.13 $\rightarrow$ 82.94
    \\
    4     
    &15 &196 &1045
    &84.78 $\rightarrow$ 88.70 &77.02 $\rightarrow$81.01 &72.47 $\rightarrow$ 77.79
    \\
    \bottomrule
    \end{tabular}
\end{threeparttable}
}
\end{table*}


\fi

\end{document}